\documentclass[conference]{IEEEtran}
\IEEEoverridecommandlockouts

\usepackage{amsmath}
\usepackage{amsthm}

\usepackage{amsfonts}
\usepackage{amssymb}
\usepackage[noadjust]{cite} 
\usepackage{hyperref}
\usepackage{float}
\usepackage{algorithm}
\usepackage{algorithmic}
\usepackage[tight,footnotesize]{subfigure}
\usepackage{multirow}
\usepackage{adjustbox} 
\usepackage{booktabs}
\usepackage{graphicx}
\usepackage{textcomp}
\usepackage{xcolor}
\def\BibTeX{{\rm B\kern-.05em{\sc i\kern-.025em b}\kern-.08em
    T\kern-.1667em\lower.7ex\hbox{E}\kern-.125emX}}
\usepackage{placeins}
\usepackage{clipboard}

\usepackage{amsthm}
\newtheorem{question}{Q}

\usepackage{color, colortbl}
\definecolor{Gray}{gray}{0.9}

\usepackage{xcolor}

\newcommand{\figref}[1]{Fig.~\ref{#1}}
\newcommand{\tabref}[1]{Table~\ref{#1}}
\newcommand{\qref}[1]{\textbf{Q~\ref{#1}}}
\newcommand{\secref}[1]{Section~\ref{#1}}
\renewcommand{\eqref}[1]{Eq.~\ref{#1}}

\newcommand{\obs}[0]{\boldsymbol{o}}

\newcommand{\state}[0]{\boldsymbol{s}}

\newcommand{\action}[0]{\boldsymbol{u}}

\newcommand{\control}[0]{u}
\newcommand{\controlmax}[0]{\control_{\rm max}}

\newcommand{\reward}[0]{r}
\newcommand{\policy}[0]{\pi_{\phi}}
\newcommand{\radius}[0]{d}
\newcommand{\azimuth}[0]{\theta_{a}}
\newcommand{\elevationangle}[0]{\theta_{e}}
\newcommand{\sunangle}[0]{\theta_{\rm Sun}}
\newcommand{\numpoints}[0]{n_p}

\newcommand{\totalvel}[0]{\Vert \boldsymbol{v} \Vert}
\newcommand{\maxvel}[0]{\totalvel_{\rm max}}
\newcommand{\position}[0]{\Vert \boldsymbol{p} \Vert}
\newcommand{\deltav}[0]{\Delta V}
\newcommand{\deltat}[0]{\Delta t}

\begin{document}

\title{Investigating the Impact of Choice on Deep Reinforcement Learning for Space Controls\\
% \thanks{This work is supported by the Department of Defense (DoD) through the Air Force Research Laboratory Innovation Fund Program. The views expressed are those of the authors and do not reflect the official guidance or position of the United States Government, the Department of Defense or of the United States Air Force. This work has been approved for public release: distribution unlimited. Case Number AFRL-XXXX-XXXXX.}
}

\author{
\IEEEauthorblockN{Nathaniel Hamilton\textsuperscript{\textsection}}
\IEEEauthorblockA{\textit{Parallax Advanced Research} \\
Beavercreek, OH, USA \\
nathaniel.hamilton@parallaxresearch.org}
\and
\IEEEauthorblockN{Kyle Dunlap\textsuperscript{\textsection}}
\IEEEauthorblockA{\textit{Parallax Advanced Research} \\
Beavercreek, OH, USA \\
kyle.dunlap@parallaxresearch.org}
\and
\IEEEauthorblockN{Kerianne L. Hobbs}
\IEEEauthorblockA{\textit{Autonomy Capability Team (ACT3)} \\
\textit{Air Force Research Laboratory}\\
Wright-Patterson Air Force Base, USA \\
kerianne.hobbs@us.af.mil}
}

\maketitle
\begingroup\renewcommand\thefootnote{\textsection}
\footnotetext{These authors contributed equally to this work.}
\endgroup

%% REMOVE THIIIIIS %%% 
% \todo{Remove page numbers}
% \thispagestyle{plain}
% \pagestyle{plain}

\begin{abstract}
% \todo{write this}
% Reinforcement Learning (RL) has become increasingly popular after demonstrating promising performance and success across many complex tasks. For space applications, traditional control methods are often used during operation. As the number of space assets continues to grow, RL can enable rapid development of control methods for different space related tasks. 
For many space applications, traditional control methods are often used during operation. However, as the number of space assets continues to grow, autonomous operation can enable rapid development of control methods for different space related tasks. One method of developing autonomous control is Reinforcement Learning (RL), which has become increasingly popular after demonstrating promising performance and success across many complex tasks.
While it is common for RL agents to learn bounded continuous control values, this may not be realistic or practical for many space tasks that traditionally prefer an on/off approach for control. This paper analyzes using discrete action spaces, where the agent must choose from a predefined list of actions. The experiments explore how the number of choices provided to the agents affects their measured performance during and after training. This analysis is conducted for an inspection task, where the agent must circumnavigate an object to inspect points on its surface, and a docking task, where the agent must move into proximity of another spacecraft and ``dock'' with a low relative speed. A common objective of both tasks, and most space tasks in general, is to minimize fuel usage, which motivates the agent to regularly choose an action that uses no fuel. Our results show that a limited number of discrete choices leads to optimal performance for the inspection task, while continuous control leads to optimal performance for the docking task.
\end{abstract}


%%%%%%%%%%%%%%%%%%%%%%%%%%%%%%%%%%%%%%%%%%%%%%%%%%%%%%%%%%%%%%%%%%%%%%
% Introduction
%%%%%%%%%%%%%%%%%%%%%%%%%%%%%%%%%%%%%%%%%%%%%%%%%%%%%%%%%%%%%%%%%%%%%%
\section{Introduction}

Autonomous spacecraft operation is critical as the number of space assets grows and operations become more complex. For \textit{On-orbit Servicing, Assembly, and Manufacturing} (OSAM) missions, tasks such as inspection and docking enable the ability to assess, plan for, and execute different objectives. While these tasks are traditionally executed using classical control methods, this requires constant monitoring and adjustment by human operators, which becomes challenging or even impossible as the complexity of the task increases. To this end, the importance of developing high-performing autonomy is growing.

\textit{Reinforcement Learning} (RL) is a fast-growing field for developing high-performing autonomy with growing impact, spurred by success in agents that learn to beat human experts in games like Go \cite{silver2016mastering} and Starcraft \cite{starcraft2019}. RL is a promising application to spacecraft operations due to its ability to react in real-time to changing mission objectives and environment uncertainty \cite{ravaioli2022safe, hamilton2022zero}. Previous works demonstrate using RL to develop waypoints for an inspection mission \cite{LeiGNC22, AurandGNC23},
% \kh{, continuous control for inspection of an illuminated object \cite{vanWijkAAS_23}},
and inspecting an uncooperative space object \cite{Brandonisio2021}. Additionally, RL has also been used for similar docking problems, including a six degree-of-freedom docking task \cite{oestreich2021autonomous}, avoiding collisions during docking \cite{broida2019spacecraft}, and guidance for docking \cite{hovell2021deep}.
% \kh{Additionally, research has investigating developing methods to bound and intervene when necessary to assure safety of RL-based spacecraft control \cite{dunlap2023run, dunlap2021comparing}}.
Despite these successes, in order for RL solutions to be used in the real world, they must control the spacecraft in a way that is acceptable to human operators. 

Spacecraft control designers and operators typically prefer the spacecraft to choose from a set of discrete actions, where the thrusters are either fully on or off. In general, this follows Pontryagin's maximum principle \cite{kopp1962pontryagin}, which minimizes a cost function to find an optimal trajectory from one state to another. In this case, the cost is fuel use. In contrast, it is common for RL agents to operate in a continuous control space at a specified frequency, where control values can be any value within a certain range. Transitioning from a continuous control space to a discrete one can result in choppy control outputs with poor performance when the discretization is coarse, or an oversized policy that takes too long to train when the discretization is fine \cite{doya2000reinforcement}.

In this paper, we compare RL agents trained using continuous control and this classical control principle to determine their advantages and identify special cases. Our experiments focus on two spacecraft tasks: \textit{inspection} (viewing the surface of another vehicle) and \textit{docking} (approaching and joining with another vehicle). This paper builds on previous work done using RL to solve the inspection task with illumination \cite{vanWijkAAS_23} and the docking task \cite{ravaioli2022safe}. For the same docking task, the effect of Run Time Assurance during RL training was analyzed \cite{dunlap2021Safe, hamilton2023ablation}. For a 2D version of the docking task, LQR control was compared to a bang-bang controller \cite{dunlap2022hybrid}, similar to the agent with three discrete choices that will be analyzed in this paper.

The main contributions of this work include answering the following questions.

\begin{question}
\label{question:no-op}
    \Copy{q_text:no-op}{
    Will increasing the likelihood of choosing ``no thrust'' improve fuel efficiency?}
\end{question}
Fuel efficiency is so important in space missions because fuel is a limited resource that needs to exist beyond any single task. The most effective method to minimize fuel use is for the agent to choose “no thrust”. To this end, we explore two different ways of increasing the likelihood of choosing “no thrust”: (1) transitioning from a continuous to a discrete action space, and (2) decreasing the action space magnitude, so that the continuous range of values is smaller. The results are found in \secref{sec:no-op}.

\begin{question}
\label{question:granularity}
    \Copy{q_text:granularity}{
    % How does the operating range impact the need for “no thrust” against “near zero thrust”?
    % How does the operating range impact the need for action choices with smaller magnitudes vs. finer granularity? 
    % \kh{
    Does a smaller action magnitude or finer granularity matter more at different operating ranges?
    % }
    }
\end{question}
The inspection and docking tasks require different operating ranges. For inspection, agents need to circumnavigate the chief and being further away ensures better coverage for inspection. In contrast, agents need to move closer to the chief in order to complete the docking task.
% To this end, we explore how the operating range impacts the need for an explicit “no thrust” provided by a discrete action space against the many “near zero thrust” control values provided by a continuous control space.
To this end, we explore how the operating range impacts the need for either smaller action magnitudes to choose from or finer granularity of choices.
The results are found in \secref{sec:granularity}.

\begin{question}
\label{question:num_choices}
    \Copy{q_text:num_choices}{
    Is there an optimal balance between discrete and continuous actions?}
\end{question}
% \todo{idk how to explain this question in a similar way at the moment}

% \kh{Is there an optimal balance between action magnitude and number of discrete choices?}
% The number of choices will be varied over many different experiments, where continuous control gives the agent infinite choices. We will determine how many choices becomes too many for the agent, where performance starts to decrease.
While RL agents often perform better with continuous actions, they would likely be more accepted for use in the real world with discrete actions. 
To this end, we explore if a balance can be found between discrete and continuous control to provide an optimal solution that is suitable for RL training and real world operation.
The results are found in \secref{sec:q-num_choices}.

%%%%%%%%%%%%%%%%%%%%%%%%%%%%%%%%%%%%%%%%%%%%%%%%%%%%%%%%%%%%%%%%%%%%%%
% Background
%%%%%%%%%%%%%%%%%%%%%%%%%%%%%%%%%%%%%%%%%%%%%%%%%%%%%%%%%%%%%%%%%%%%%%
\section{Deep Reinforcement Learning}

\textit{Reinforcement Learning} (RL) is a form of machine learning in which an agent acts in an environment, learns through experience, and increases its performance based on rewarded behavior. \textit{Deep Reinforcement Learning} (DRL) is a newer branch of RL in which a neural network is used to approximate the behavior function, i.e. policy $\policy$.
% The basic construction of the DRL approach is shown in \figref{fig:rta_off}. 
% The agent consists of the \textit{Neural Network Controller} (NNC) and RL algorithm, and the environment consists of a plant and observer model.
The agent uses a \textit{Neural Network Controller} (NNC) trained by the RL algorithm to take actions in the environment, which can be comprised of any dynamical system, from Atari simulations (\cite{hamilton2020sonic, alshiekh2018safe}) to complex robotics scenarios (\cite{brockman2016gym, fisac2018general, henderson2018deep, mania2018simple, jang2019simulation, bernini2021few}). 

% \begin{figure}[t]
%     \centering
%     \includegraphics[width=.8\columnwidth]{figures/RL_basic_1.pdf}
%     \caption{DRL training loop.}
%     \label{fig:rta_off}
% \end{figure}

Reinforcement learning is based on the \textit{reward hypothesis} that all goals can be described by the maximization of expected return, i.e. the cumulative reward \cite{silver2015}. During training, the agent chooses an action, $\action_{NN}$, based on the input observation, $\obs$. The action is then executed in the environment, updating the internal state, $\state$, according to the plant dynamics. The updated state, $\state$, is then assigned a scalar reward, $\reward$, and transformed into the next observation vector. The process of executing an action and receiving a reward and next observation is referred to as a \textit{timestep}. Relevant values, like the input observation, action, and reward are collected as a data tuple, i.e. \textit{sample}, by the RL algorithm to update the current NNC policy, $\policy$, to an improved policy, $\policy^*$. How often these updates are done is dependent on the RL algorithm.

In this work we focus solely on the \textit{Proximal Policy Optimization} (PPO) algorithms as our DRL algorithm of choice. PPO has demonstrated success in the space domain for multiple tasks and excels in finding optimal policies across many other domains \cite{schulman2017proximal, hamilton2023ablation, ravaioli2022safe, dunlap2021Safe, vanWijkAAS_23}. Additionally, PPO works for both discrete and continuous action spaces, allowing us to test both types of action spaces without switching algorithms. For RL, the action space is typically defined as either discrete or continuous. For a discrete action space, the agent has a finite set of choices for the action. For a continuous action space, the agent can choose any value for the action within a given range. Therefore, it can be thought of as having infinite choices.
In general, discrete action spaces tend to be used for simple tasks while continuous action spaces tend to be used for more complex tasks.

% \kh{COMMENT: Is the following true or does it just depend on application? e.g. if you just have a thruster that turns on and off vs if you have a thruser with the ability to fine-tune magnitude? or a board game with limited actions vs how hard you hit the accelerator in a car...-- In general, discrete action spaces tend to be used for simple problems while continuous action spaces tend to be used for more complex problems.}

% \todo{need to address why we consider increasing the likelihood of selecting ``no thrust'' because RL learns from experience and does a lot of random actions at the start, so if we provide the best, most fuel efficient action with greater likelihood, it should do better at using it}

Due to the stochastic nature of RL, the agent typically selects random action values at the beginning of the training process. It then learns from this experience, and selects the actions that maximize the reward function. By using discrete actions, it becomes much easier for the agent to randomly select specific discrete actions, and the agent can quickly learn that these actions are useful. This motivates \qref{question:no-op}, where we aim to increase the likelihood of ``no thrust.''

\section{Space Environments}

\begin{figure}[t]
    \centering
    \includegraphics[width=.8\columnwidth]{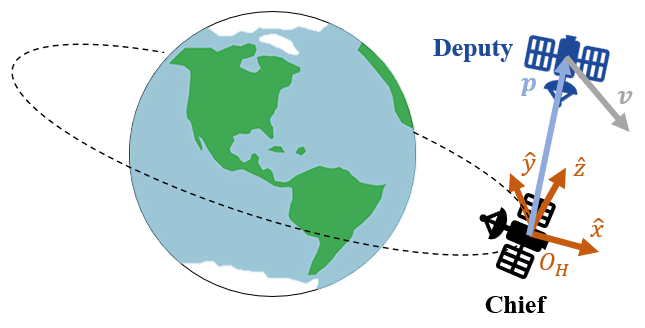}
    \caption{Deputy spacecraft navigating around a chief spacecraft in Hill's Frame.}
    \label{fig:Hills}
\end{figure}

This paper considers two spacecraft tasks: \textit{inspection} and \textit{docking}. Both tasks consider a passive ``chief'' and active ``deputy'' spacecraft, where the agent controls the deputy and the chief is stationary. Both of these tasks are modeled using the Clohessy-Wiltshire equations \cite{clohessy1960terminal} in Hill's frame \cite{hill1878researches}, which is a linearized relative motion reference frame centered around the chief spacecraft, which is in a circular orbit around the Earth. The agent is the deputy spacecraft, which is controlled in relation to the passive chief spacecraft. As shown in \figref{fig:Hills}, the origin of Hill's frame, $\mathcal{O}_H$, is located at the chief's center of mass, the unit vector $\hat{x}$ points away from the center of the Earth, $\hat{y}$ points in the direction of motion of the chief, and $\hat{z}$ is normal to $\hat{x}$ and $\hat{y}$. The relative motion dynamics between the deputy and chief are, 
\begin{equation} \label{eq: system dynamics}
    \dot{\state} = A {\state} + B\action,
\end{equation}
where $\state$ is the state vector $\state=[x,y,z,\dot{x},\dot{y},\dot{z}]^T \in \mathbb{R}^6$, $\action$ is the control vector, i.e. action, $\action= [F_x,F_y,F_z]^T \in [-\controlmax, \controlmax]^3$, and,
\begin{align}
\centering
    A = 
\begin{bmatrix} 
0 & 0 & 0 & 1 & 0 & 0 \\
0 & 0 & 0 & 0 & 1 & 0 \\
0 & 0 & 0 & 0 & 0 & 1 \\
3n^2 & 0 & 0 & 0 & 2n & 0 \\
0 & 0 & 0 & -2n & 0 & 0 \\
0 & 0 & -n^2 & 0 & 0 & 0 \\
\end{bmatrix}, 
    B = 
\begin{bmatrix} 
 0 & 0 & 0 \\
 0 & 0 & 0 \\
 0 & 0 & 0 \\
\frac{1}{m} & 0 & 0 \\
0 & \frac{1}{m} & 0 \\
0 & 0 & \frac{1}{m} \\
\end{bmatrix}.
\end{align}
Here, $n=0.001027$rad/s is the mean motion of the chief's orbit, $m=12$kg is the mass of the deputy, $F$ is the force exerted by the thrusters along each axis, and $\controlmax$ is a constant value varied in the experiments. Both spacecraft are modeled as point masses.

%%%%%%%%%%%%%%%%%%%%%%%%%%%%%%%%%%%%%
% Inspection
%%%%%%%%%%%%%%%%%%%%%%%%%%%%%%%%%%%%%
\subsection{Inspection}

For the inspection task, introduced in \cite{vanWijkAAS_23}, the agent's goal is to navigate the deputy around the chief spacecraft to inspect the entire surface of the chief spacecraft. In this case, the chief is modeled as a sphere of $99$ inspectable points distributed uniformly across the surface. The attitude of the deputy is not modeled, because it is assumed that the deputy is always pointed towards the chief. In order for a point to be inspected, it must be within the field of view of the deputy (not obstructed by the near side of the sphere) and illuminated by the Sun. Illumination is determined using a binary ray tracing technique, where the Sun rotates in the $\hat{x}-\hat{y}$ plane in Hill's frame at the same rate as mean motion of the chief's orbit, $n$.

While the main objective of the task is to inspect all points, a secondary objective is to minimize fuel use. This is considered in terms of $\deltav$, where,
\begin{equation}
    \deltav = \frac{|F_{x}| + |F_{y}| + |F_{z}|}{m} \deltat.
\end{equation}
For this task, $\deltat = 10$ seconds.

\subsubsection{Initial and Terminal Conditions}

Each episode is randomly initialized given the following parameters. First, the Sun is initialized at a random angle with respect to the $\hat{x}$ axis so that $\sunangle \in [0, 2\pi]$rad. Next, the deputy's position is sampled from a uniform distribution for the parameters: radius $\radius \in [50, 100]$m, azimuth angle $\azimuth \in [0, 2\pi]$rad, and elevation angle $\elevationangle \in [-\pi/2, \pi/2]$rad. The position is then computed as,
\begin{equation} \label{eq:init}
    \begin{gathered}
        x = \radius \cos(\azimuth) \cos(\elevationangle), \\
        y = \radius \sin(\azimuth) \cos(\elevationangle), \\
        z = \radius \sin(\elevationangle). \\
    \end{gathered}
\end{equation}
If the deputy's initialized position results in pointing within 30 degrees of the Sun, the position is negated such that the deputy is then pointing away from the Sun and towards illuminated points. This prevents unsafe and unrealistic initialization, as sensors can burnout when pointed directly at the Sun. Finally, the deputy's velocity is similarly sampled from a velocity magnitude $\totalvel \in [0, 0.3]$m/s, azimuth angle $\azimuth \in [0, 2\pi]$rad, and elevation angle $\elevationangle \in [-\pi/2, \pi/2]$rad, and the velocity is computed using the same technique as \eqref{eq:init}.

An episode is terminated under the following conditions: 
\begin{itemize}
    \item the deputy inspects all $99$ points, 
    \item the deputy crashes into the chief (enters within a minimum relative distance of $15$m, where the chief and deputy have radii $10$ and $5$m respectively), 
    \item the deputy exceeds a maximum relative distance from the chief of $800$m, and/or 
    \item the simulation exceeds $1223$ timesteps (the time for the Sun to appear to orbit the chief twice, or $3.4$hrs).
\end{itemize}

\subsubsection{Observations}

The environment is partially observable, using sensors to condense full state information into manageable components of the observation space. At each timestep, the agent receives an observation comprised of the following components. The first component is the deputy's current position in Hill's frame, where each element is divided by a value of $100$ to ensure most values fall in the range $[-1, 1]$. The second component is the deputy's current velocity in Hill's frame, where each element is multiplied by a value of $2$ to ensure most values fall in the range $[-1, 1]$. The third component is the angle describing the Sun's position with respect to the $\hat{x}$ axis, $\sunangle$. The fourth component is the total number of points that have been inspected so far during the episode, $\numpoints$, divided by a value of $100$. The final component is a unit vector pointing towards the nearest cluster of uninspected points, where clusters are determined using k-means clustering. The resulting observation is thus $\obs = [x, y, z, \dot{x}, \dot{y}, \dot{z}, \numpoints, \sunangle, x_{UPS}, y_{UPS}, z_{UPS}]$.

\subsubsection{Reward Function}

The reward function consists of the following three elements\footnote{The reward function was defined in \cite{vanWijkAAS_23}, and the authors determined that the specified configuration produces the desired behavior. An exploration of the reward function is outside the scope of this work.}. First, a reward of $+0.1$ is given for every new point that the deputy inspects at each timestep. Second, a negative reward is given that is proportional to the $\deltav$ used at each timestep. This is given as $-w*\deltav$, where $w$ is a scalar multiplier that changes during training to help the agent first learn to inspect all points and then minimize fuel usage. At the beginning of training, $w=0.001$. If the mean percentage of inspected points for the previous training iteration exceeds 90\%, $w$ is increased by $0.00005$, and if this percentage drops below 80\% for the previous iteration, $w$ is decreased by the same amount. $w$ is enforced to always be in the range $[0.001, 0.1]$. Finally, a reward of $-1$ is given if the deputy collides with the chief and ends the episode. This is the only sparse reward given to the agent.
For evaluation, a constant value of $w=0.1$ is used, while all other rewards remain the same.

%%%%%%%%%%%%%%%%%%%%%%%%%%%%%%%%%%%%%
% Docking
%%%%%%%%%%%%%%%%%%%%%%%%%%%%%%%%%%%%%
\subsection{Docking}

For the docking task, the agent's goal is to navigate the deputy spacecraft to within a docking radius, $\radius_d = 10$m of the chief at a relative speed below a maximum docking speed, $\nu_0 = 0.2$m/s. Secondary objectives for the task are to minimize fuel use and adhere to a distance-dependent speed limit defined as,
\begin{equation}\label{eq:vel_lim}
    \totalvel \leq \nu_0 + \nu_1 (\position  - \radius_d),
\end{equation}
where $\position$ and $\totalvel$ are the deputy's position and velocity, and $\nu_1=2n$rad/s is the slope of the speed limit. The distance-dependent speed limit requires the deputy to slow down as it approaches the chief to dock safely, and values were chosen based on their relation to elliptical natural motion trajectories \cite{Dunlap2023}.

\subsubsection{Initial and Terminal Conditions}

Each episode is randomly initialized given the following parameters. First, the deputy's position is sampled from a radius $\radius \in [100, 150]$m, azimuth angle $\azimuth \in [0, 2\pi]$rad, and elevation angle $\elevationangle \in [-\pi/2, \pi/2]$rad, where position is computed according to \eqref{eq:init}. Second, velocity is sampled from a maximum velocity magnitude $\totalvel \in [0, 0.8]*\totalvel_{\rm max}$ (where $\totalvel_{\rm max}$ is determined by \eqref{eq:vel_lim} given the current position), azimuth angle $\azimuth \in [0, 2\pi]$rad, and elevation angle $\elevationangle \in [-\pi/2, \pi/2]$rad, where velocity is computed according to \eqref{eq:init}.

An episode is terminated under the following conditions: 
\begin{itemize}
    \item the deputy successfully docks with the chief ($\position  \leq \radius_d, \totalvel \leq \nu_0$), 
    \item the deputy crashes into the chief ($\position  \leq \radius_d, \totalvel > \nu_0$), 
    \item the deputy exceeds a maximum relative distance from the chief of $800$m, and/or 
    \item the simulation exceeds $2000$ timesteps ($\deltat$ = 1 second).
\end{itemize}

\subsubsection{Observations}

% Similar to the inspection environment, the docking environment is partially observable and broken up into components.
Similar to the inspection environment, the docking environment's observation is broken up into components. The first and second components are the deputy's position and velocity, divided by $100$ and $0.5$ respectively. The third component is the deputy's current velocity magnitude $\totalvel$, and the fourth component is the maximum velocity given the current position according to \eqref{eq:vel_lim}, $\maxvel$. Thus, the observation is $\obs = [x, y, z, \dot{x}, \dot{y}, \dot{z}, \totalvel, \maxvel]$.

\subsubsection{Reward Function}

The reward function consists of the following six elements\footnote{The reward function was defined in \cite{ravaioli2022safe}, and the authors determined that the specified configuration produces the desired behavior. An exploration of the reward function is outside the scope of this work.}. First, a distance change reward is used to encourage the deputy to move towards the chief at each timestep. This reward is given  by,
\begin{equation}
    \reward = 2 * (e^{-a \position_0} - e^{-a \position_{-1}}),
\end{equation}
where $a = \log(2)/100$, $\position_0$ is the deputy's current position, and $\position_{-1}$ is the deputy's position at the last timestep. Second, a negative reward of $-0.01$ is multiplied by the $\deltav$ used by the deputy at each timestep. Unlike the inspection task, this reward remains constant throughout training. Third, if the deputy violated the distance dependent speed limit at the current timestep, a negative reward of $-0.01$ is multiplied by the magnitude of the violation (that is, $-0.01 (\totalvel - \maxvel)$). Fourth, a negative reward of $-0.01$ is given at each timestep to encourage the agent to complete the task as quickly as possible. Fifth, a sparse reward of $+1$ is given if the agent successfully completes the task. Finally, a sparse reward of $-1$ is given if the agent crashes into the chief.

%%%%%%%%%%%%%%%%%%%%%%%%%%%%%%%%%%%%%%%%%%%%%%%%%%%%%%%%%%%%%%%%%%%%%%
% Experiments
%%%%%%%%%%%%%%%%%%%%%%%%%%%%%%%%%%%%%%%%%%%%%%%%%%%%%%%%%%%%%%%%%%%%%%
\section{Experiments}

A common objective of both of these tasks (and most space tasks in general) is to minimize the use of $\deltav$. If the agent always chooses a value of zero for all controls, it will use zero m/s of $\deltav$. However in this case, it is unlikely that the agent is able to successfully complete the task, and therefore a balance must be found between maximizing task completion and minimizing $\deltav$. With continuous actions, it is very difficult for an agent to choose an exact value of zero for control, and therefore it is often using a small amount of $\deltav$ at every timestep, as seen in \cite{vanWijkAAS_23}. On the other hand, discrete actions allow an agent to easily choose zero.% This also aligns with how spacecraft are often operated in the real world, where thrusters fire with full force or are turned off \todo{do we have a cite for this?}.

Several experiments are run for both the inspection and docking environments to determine how choice affects the learning process. First, a baseline configuration is trained with continuous actions, where the agent can choose any value for $\action \in [-\controlmax, \controlmax]$. Next, several configurations are trained with discrete actions where the number of choices is varied. In each case, the action values are evenly spaced over the interval $[-\controlmax, \controlmax]$. For example, 3 choices for the agent are $[-\controlmax, 0, \controlmax]$ and 5 choices for the agent are $[-\controlmax, -\controlmax/2, 0, \controlmax/2, \controlmax]$. Experiments are run for 3, 5, 7, 9, 11, 21, 31, 41, 51, and 101 choices. The number of choices is always odd such that zero is an option. This set of experiments is repeated for values of $\controlmax = 1.0$N and $\controlmax = 0.1$N, to determine if the magnitude of the action choices affect the results.

For the docking environment, two additional configurations are trained: 5 discrete choices $[-1.0, -0.1, 0.0, 0.1, 1.0]$ (referred to as $1.0/0.1$), and 9 discrete choices $[-1.0, -0.1, -0.01, -0.001, 0.0, 0.001, 0.01, 0.1, 1.0]$ (referred to as $1.0/../0.001$). These configurations are designed to give the agent finer control at small magnitudes, and rationale will be discussed further in \secref{sec:q-num_choices}.

For each configuration, 10 different agents are trained for different random seeds (which are held constant for all the training configurations). Each agent is trained over 5 million timesteps. The policies are periodically evaluated during training, approximately every 500,000 timesteps, to record their performance according to several metrics\footnote{The training curves are not shown in the results, but are included in the Appendix.}. The common metrics for both the inspection and docking environments are: average $\deltav$ used per episode, average percentage of successful episodes, average total reward per episode, and average episode length. For the inspection environment, the average number of inspected points is also considered, and for the docking environment, the average number of timesteps where the speed limit is violated and the average final speed are both considered. 

Each of the 10 policies is evaluated over a set of 10 random test cases, where the same test cases are used every time the policy is evaluated. We record and present the \textit{InterQuartile Mean}\footnote{IQM sorts and discards the bottom and top 25\% of the recorded metric data and calculates the mean score on the remaining middle 50\% of data. IQM interpolates between mean and median across runs, for a more robust measure of performance \cite{agarwal2021deep}.} (IQM) for each metric. The IQM is used as it is a better representation of what we can expect to see with future studies as it is not unduly affected by outliers and has a smaller uncertainty even with a handful of runs \cite{agarwal2021deep}. At the conclusion of training, the final trained policies are again evaluated deterministically for 100 random test cases to better understand the behavior of the trained agents.

% Inspection and Docking deltaV comparisons
\begin{figure*}[ht]
\centering
\subfigure[$\deltav$ use in the inspection environment with $\controlmax=1.0$N.]{\includegraphics[width=0.45\linewidth]{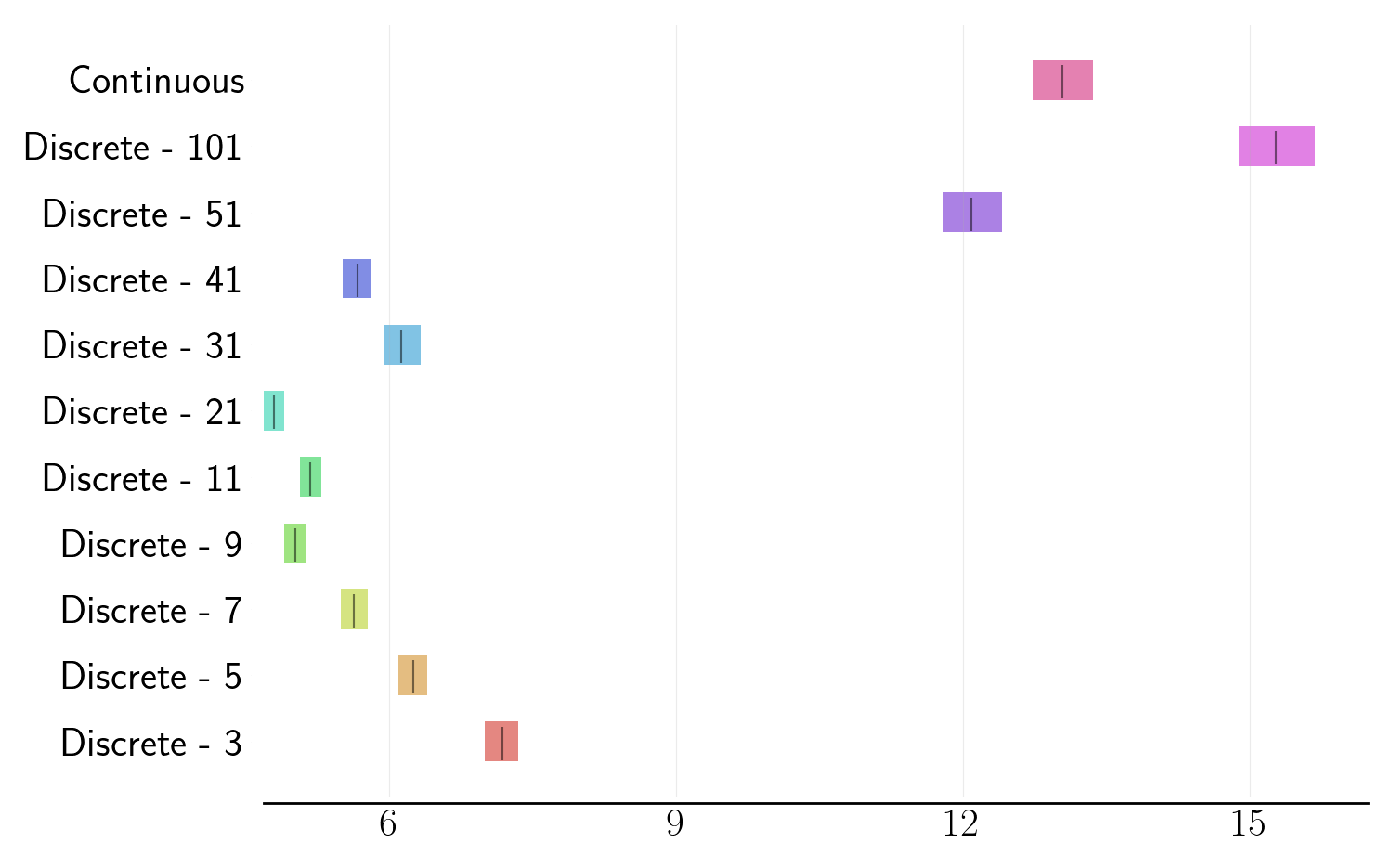}\label{subfig:delta-v-1.0-insp}}
% \subfigure[Inspection environment, $\controlmax=0.1$N.]{\includegraphics[width=0.45\linewidth]{figures/inspection/0_1/DeltaV_int_est.png}\label{subfig:delta-v-0.1-insp}}
\subfigure[$\deltav$ use in the inspection environment with $\controlmax=0.1$N.]{\includegraphics[width=0.45\linewidth]{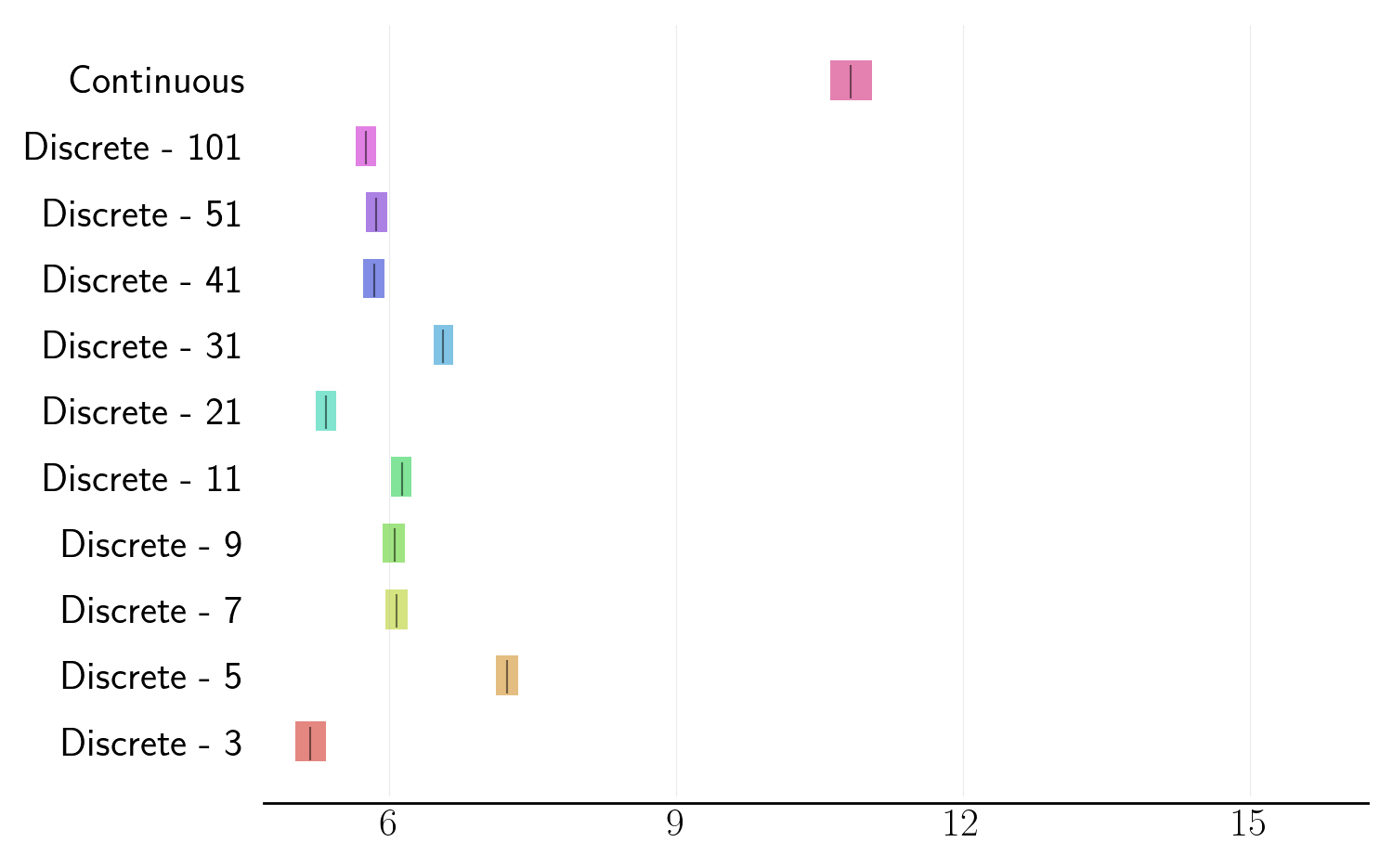}\label{subfig:delta-v-0.1-insp}}
\subfigure[$\deltav$ use in the docking environment with $\controlmax=1.0$N.]{\includegraphics[width=0.45\linewidth]{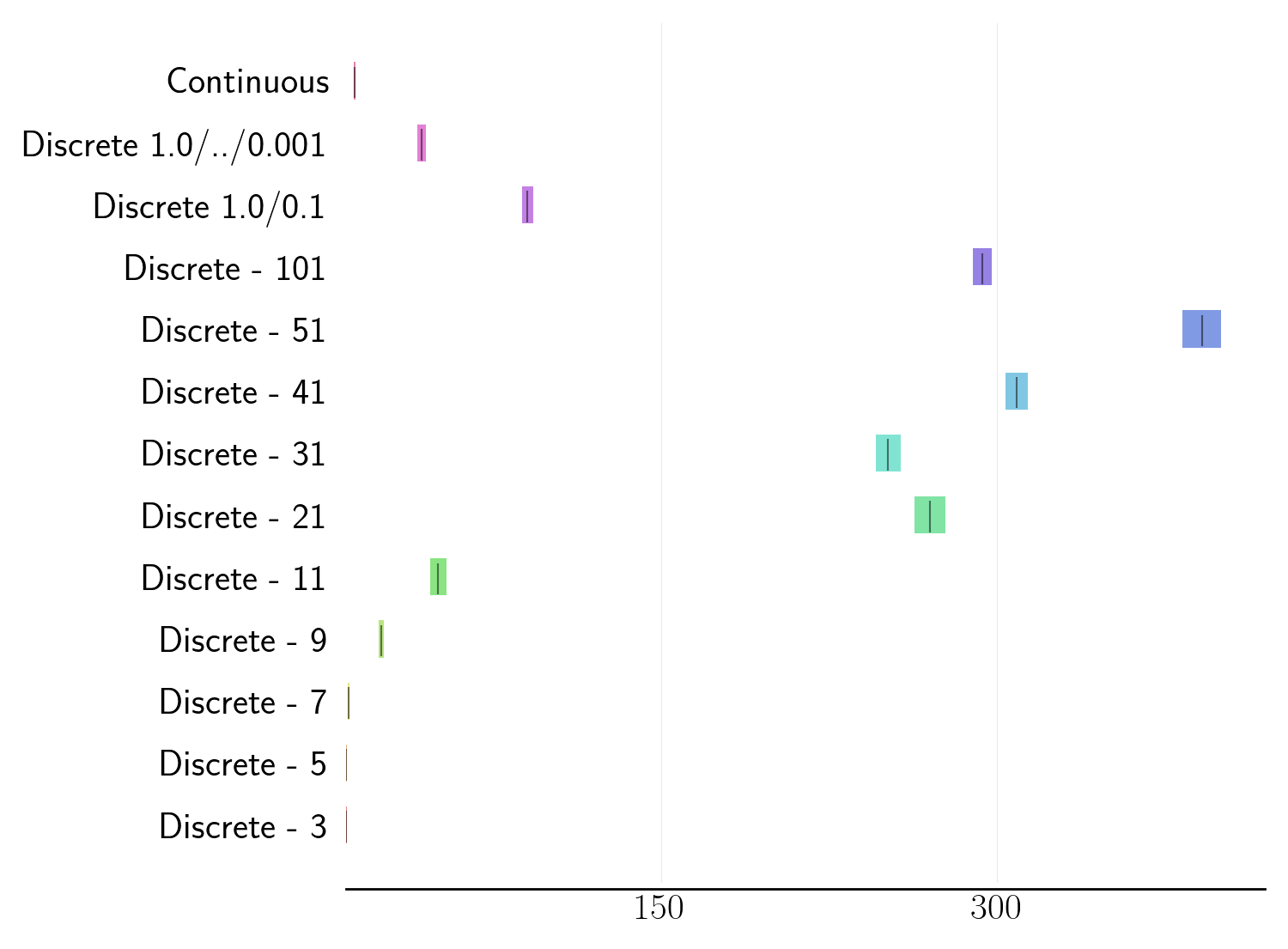}\label{subfig:delta-v-1.0-dock}}
\subfigure[$\deltav$ use in the docking environment with $\controlmax=0.1$N.]{\includegraphics[width=0.45\linewidth]{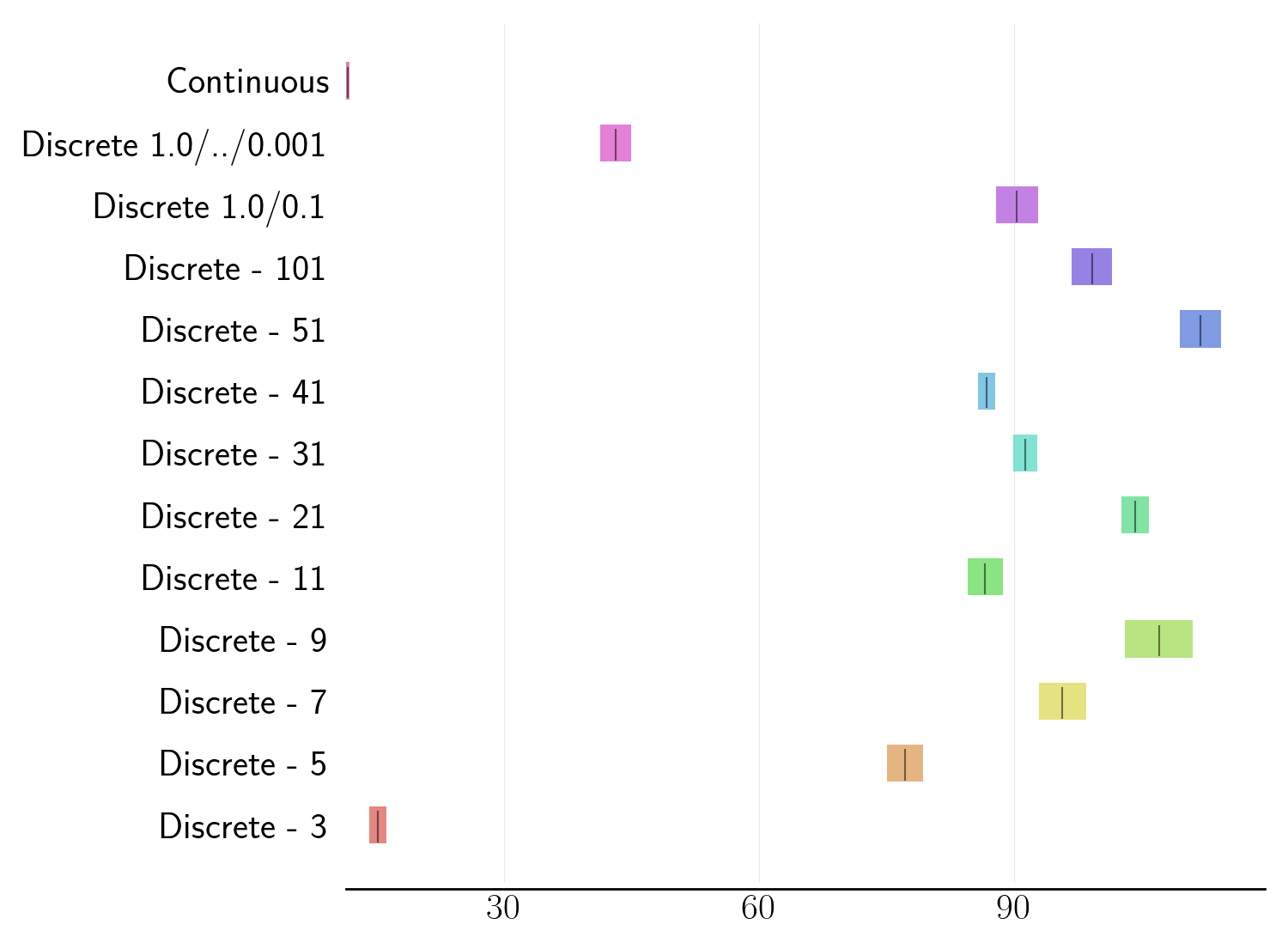}\label{subfig:delta-v-0.1-dock}}
\caption{Comparison of $\deltav$ (m/s) for all final policies. Each marker represents the IQM of 100 trials, and the shaded region is the $95\%$ confidence interval about the IQM.}
\label{fig:delta-v-int-est}
\end{figure*}

%%%%%%%%%%%%%%%%%%%%%%%%%%%%%%%%%%%%%%%%%%%%%%%%%%%%%%%%%%%%%%%%%%%%%%
% Results
%%%%%%%%%%%%%%%%%%%%%%%%%%%%%%%%%%%%%%%%%%%%%%%%%%%%%%%%%%%%%%%%%%%%%%
\section{Results and Discussion} \label{sec:results}
% \todo{re-org section to focus on question and answer format}

In this section, we answer the questions posed in the introduction by analyzing the overarching trends found in our experiments. In the interest of being concise but detailed, we include here selected results that highlight the trends we found and provide all the results in the Appendix.

%%%%%%%%%%%%%%%%%%%%%%%%%%%%%%%%%%%%%%%%%%%%%%%%%%%%%%%%%%%%%%%%%%%%%%%%%%%%%%%%%%
% Easy No-Op -> less deltaV?
%%%%%%%%%%%%%%%%%%%%%%%%%%%%%%%%%%%%%%%%%%%%%%%%%%%%%%%%%%%%%%%%%%%%%%%%%%%%%%%%%%
\subsection{\Paste{q_text:no-op}}
\label{sec:no-op}
\noindent \textbf{Answer:} Yes. Increasing the likelihood of selecting ``no thrust" as an action greatly reduces $\deltav$ use.

\begin{table*}[htb]
\centering
\caption{Inspection With Continuous Actions, IQM $\pm$ STD}
\begin{tabular}{llllll}
\toprule
$\controlmax$ & Total Reward & Inspected Points & Success Rate & $\deltav$ (m/s) & Episode Length (steps) \\
\midrule
$1.0$N & 7.8198 $\pm$ 0.5292 & 95.81 $\pm$ 4.6808 & 0.448 $\pm$ 0.4973 & 13.0222 $\pm$ 2.0929 & 323.98 $\pm$ 36.636 \\
$0.1$N & 8.7324 $\pm$ 0.199 & 99.0 $\pm$ 0.0 & 1.0 $\pm$ 0.0 & 10.8143 $\pm$ 1.7896 & 333.496 $\pm$ 13.6857 \\
\bottomrule
\end{tabular}
\label{tab:con_insp}
\end{table*}

% \kh{do we have values for discrete actions to compare to the values for discrete actions in tables I and II? - okay found it in the appendix...maybe we move all to the appendix and just refer to plots?}

\begin{table*}[htb]
\centering
\caption{Docking With Continuous Actions, IQM $\pm$ STD}
\begin{tabular}{lllllll}
\toprule
$\controlmax$ & Total Reward & Success Rate & $\deltav$ (m/s) & Violation (\%) & Final Speed (m/s) & Episode Length (steps) \\
\midrule
$1.0$N & 1.4105 $\pm$ 0.5279 & 0.57 $\pm$ 0.4951 & 13.2319 $\pm$ 1.7153 & 0.0 $\pm$ 0.0 & 0.0141 $\pm$ 0.0043 & 1780.944 $\pm$ 237.6564 \\
$0.1$N & 1.8289 $\pm$ 0.5193 & 0.842 $\pm$ 0.3647 & 11.6234 $\pm$ 1.3619 & 0.0 $\pm$ 0.0 & 0.0131 $\pm$ 0.0074 & 1497.154 $\pm$ 343.938 \\
\bottomrule
\end{tabular}
\label{tab:con_dock}
\end{table*}

% \todo{restructure for 1) discrete vs continuous, 2) more discrete options, and 3) reducing umax}

To answer this question, we employed two methods for increasing the likelihood of selecting ``no thrust'' (i.e. $\control = [0.0, 0.0, 0.0]$N): (1) transitioning from a continuous to a discrete action space, and (2) decreasing the action space magnitude so the continuous range is smaller.  

\subsubsection{Continuous to Discrete Action Space}

Transitioning from a continuous action space to a discrete one increases the likelihood of selecting “no thrust” by making it an explicit choice. With a continuous action space, the directional thrusts can be any value between $\pm \controlmax$ so the likelihood of all directional thrust randomly being exactly 0.0 is very low. There are many combinations when all thrust values are near 0.0. 

With a discrete action space, it is straightforward for the agent to select exactly 0.0.
% \todo{We test further how reducing choices in the discrete action space... idk how to words it} 
However, depending on the number of choices available to the agent, it can become more difficult for the agent to choose zero thrust. For the agent with three discrete choices, there is a $1$ in $3^3$ chance that the agent does not thrust at all (due to the three control inputs), while for the agent with $101$ discrete choices, there is a $1$ in $101^3$ chance that the agent does not thrust at all. Therefore, it is easier for agents with fewer discrete actions to explicitly choose “no thrust.”
% \kh{dumb question...it isn't obvious to me while its 1 in  $101^3$ chance it choses no thrust at all. and why not just 1 in 3 or 1 in 101.} \todo{because of the 3 directional components. There are 101 choices for each direction (x, y, z).}
% However, although it is harder for the agent to choose zero thrust with more choices, it enables the agent to choose actions with thrust closer to zero.

In \figref{fig:delta-v-int-est} we compare the $\deltav$ used by the final policies trained in continuous and discrete action spaces. For the inspection environment, our results show that transitioning to a discrete action space generally reduces $\deltav$ use. 
% Interestingly, we see some \todo{trends in the number of discrete choices, with larger range (bigger umax) adding choices improves fuel efficiency until more than 21 choices, showing a peak. With the smaller range, fuel efficiency platues and fewer appears to be better for this application.}
Interestingly, we see in \figref{subfig:delta-v-1.0-insp} that adding more discrete choices reduces $\deltav$ use until the agent has 31 choices or more. This shows that while it is harder for the agent to choose zero thrust with more choices, it also enables the agent to choose actions with thrust closer to zero, reducing the over-corrections caused by a coarse discretization of the action space.

For the docking environment, our results show that transitioning to a discrete action space generally results in a large increase in $\deltav$ use. The blocks representing the continuous configurations $\deltav$ use in \figref{fig:delta-v-int-est}(c)\&(d) are centered around $13.23$m/s and $11.62$m/s respectively. Increasing the granularity of choices did not help, instead trending towards larger use of $\deltav$ instead. However, reducing the number of discrete choices clearly reduces $\deltav$ use as it is easier to choose “no thrust.”

% For the docking environment with $\controlmax=1.0$N, it is clear that more choices leads to higher $\deltav$ use. This is also generally true for the inspection environment with $\controlmax=1.0$N, but we see that 9, 11, and 21 choices lead to slightly lower $\deltav$ use. This occurs because the agents are able to choose actions that are close to zero, but not exactly zero, which helps the agents complete the task more efficiently.
% By reducing $\controlmax$, this makes it easier for the agent to choose actions that are closer to zero. For both environments with $\controlmax=0.1$N, the agents trained with three discrete actions result in the lowest $\deltav$ use, but all other configurations result in similar $\deltav$ use.

% \begin{figure}[ht]
% \centering
% \subfigure[Inspection environment, $\controlmax=0.1$N.]{\includegraphics[width=\linewidth]{figures/inspection/0_1/DeltaV_int_est.png}\label{subfig:delta-v-0.1-insp}}\\
% \subfigure[Docking environment, $\controlmax=0.1$N.]{\includegraphics[width=\linewidth]{figures/docking/0_1/DeltaV_int_est.png}\label{subfig:delta-v-0.1-dock}}
% \caption{Comparison of $\deltav$ (m/s) for the final policies with $\controlmax=0.1$N. Each marker represents the IQM of 100 trials, and the shaded region is the $95\%$ confidence interval about the IQM.}
% \label{fig:delta-v-0.1-int-est}
% \end{figure}

\subsubsection{Decreasing the Action Space Magnitude}

% \todo{how does umax decrease increase likelihood?}
As mentioned earlier, it is difficult to select ``no thrust" with a continuous action space, but there are many combinations of ``near zero" available. By reducing the magnitude of the action space (i.e. decreasing $\controlmax$) we increase the likelihood of choosing those ``near zero" actions for both our continuous and discrete configurations. Additionally, by reducing $\controlmax$, we decrease the maximum fuel use for any given timestep, which should also result in a reduction in fuel use.

Our results in \figref{fig:delta-v-int-est} show that reducing $\controlmax$ from $1$N to $0.1$N generally reduces the amount of $\deltav$ used, in some cases reducing by more than $300$m/s in the docking environment (discrete 41, 51, and 101). Therefore, to reduce $\deltav$ use, our results show it is best to reduce $\controlmax$.

To better highlight how reducing $\controlmax$ impacts agents with continuous actions, \tabref{tab:con_insp} and \tabref{tab:con_dock} show the performance of the final policies across all metrics for the inspection and docking environments respectively. \tabref{tab:con_insp} shows that as $\controlmax$ is decreased from $1.0$N to $0.1$N for the inspection environment, the total reward, inspected points, and success rate all increase while $\deltav$ decreases. \tabref{tab:con_dock} similarly shows that as $\controlmax$ is decreased in the docking environment, the total reward and success rate increase while $\deltav$ decreases. Both cases show that decreasing the action space magnitude enables better performance.

% For the agents with discrete actions, \figref{fig:delta-v-int-est} shows that $\deltav$ use generally decreases as the action magnitude is decreased. For the inspection environment, the agents trained with 3, 51, and 101 discrete actions see a great reduction in $\deltav$ use as the action magnitude is decreased. For the docking environment, the $\deltav$ use for agents trained with 21 to 101 discrete actions is significantly reduced by over $200$m/s in some cases.

%%%%%%%%%%%%%%%%%%%%%%%%%%%%%%%%%%%%%%%%%%%%%%%%%%%%%%%%%%%%%%%%%%%%%%%%%%%%%%%%%%
% More Granular = Better Performance?
%%%%%%%%%%%%%%%%%%%%%%%%%%%%%%%%%%%%%%%%%%%%%%%%%%%%%%%%%%%%%%%%%%%%%%%%%%%%%%%%%%
\subsection{\Paste{q_text:granularity}}
\label{sec:granularity}
\noindent \textbf{Answer:} It depends on the task. For the inspection task, smaller action magnitude is more important. For the docking task, finer granularity is more important.

% Histogram
\begin{figure*}[ht]
\centering
\subfigure[Inspection environment with $\controlmax=1.0$N.]{\includegraphics[width=.32\linewidth]{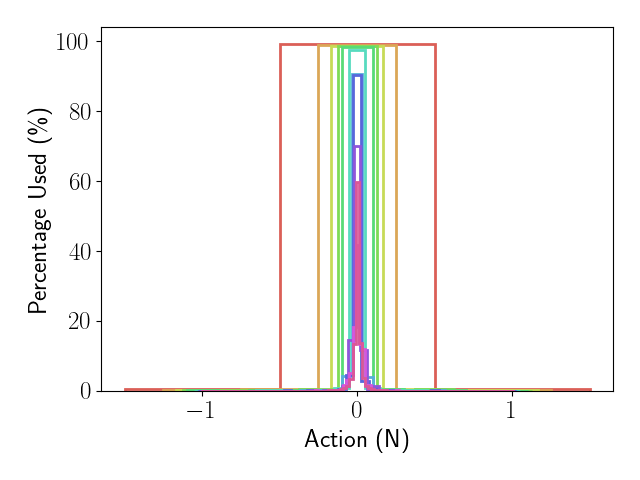}}
\subfigure[Inspection environment with $\controlmax=0.1$N.]{\includegraphics[width=.32\linewidth]{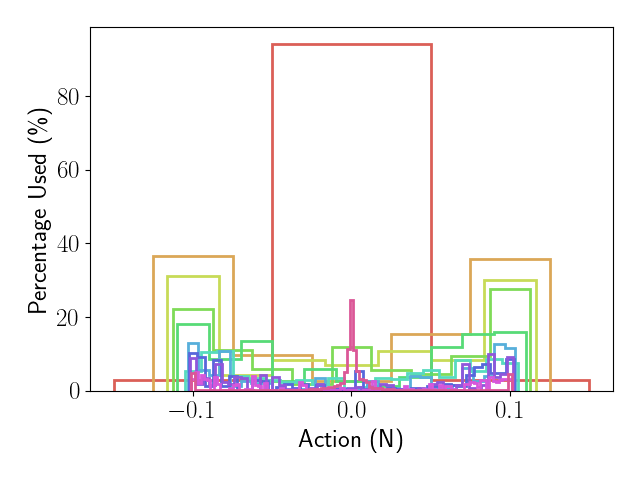}\label{subfig:histogram-w}}
\subfigure[Docking environment with $\controlmax=1.0$N.]{\includegraphics[width=.32\linewidth]{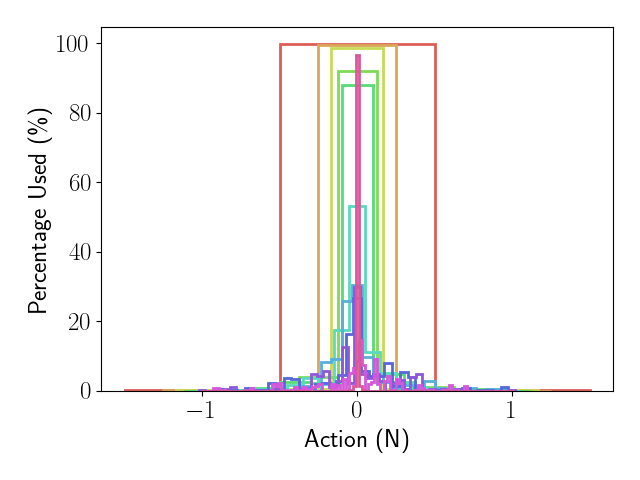}}
\subfigure[Docking environment with $\controlmax=0.1$N.]{\includegraphics[width=.32\linewidth]{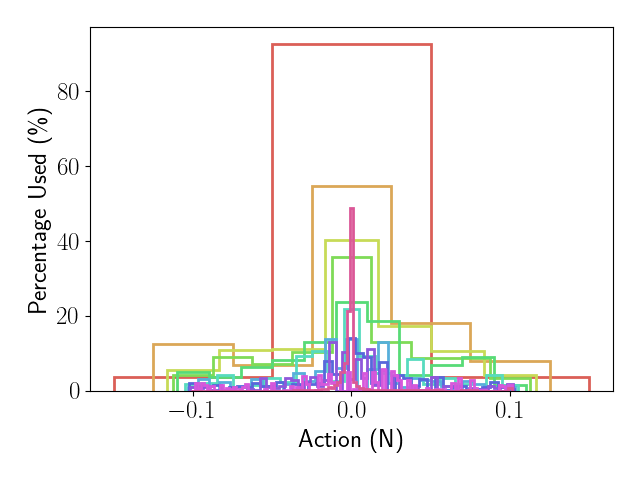}\label{subfig:histogram-bell}}
\subfigure{\includegraphics[width=.32\linewidth]{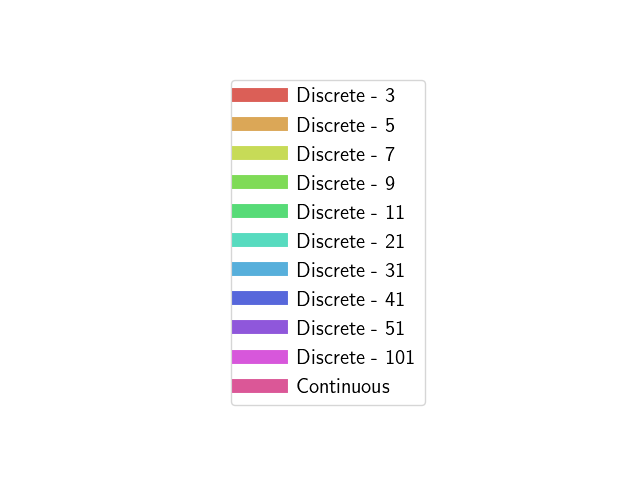}}
\caption{Comparison of actions taken in each environment. Each histogram shows the percentage of actions used the experiments, with the policies trained with continuous actions divided into 101 discrete intervals for comparison.}
\label{fig:histogram}
\end{figure*}

% Inspection reward comparison
\begin{figure*}[ht]
\centering
\subfigure[Total reward in the inspection environment with $\controlmax=1.0$N.]{\includegraphics[width=.48\linewidth]{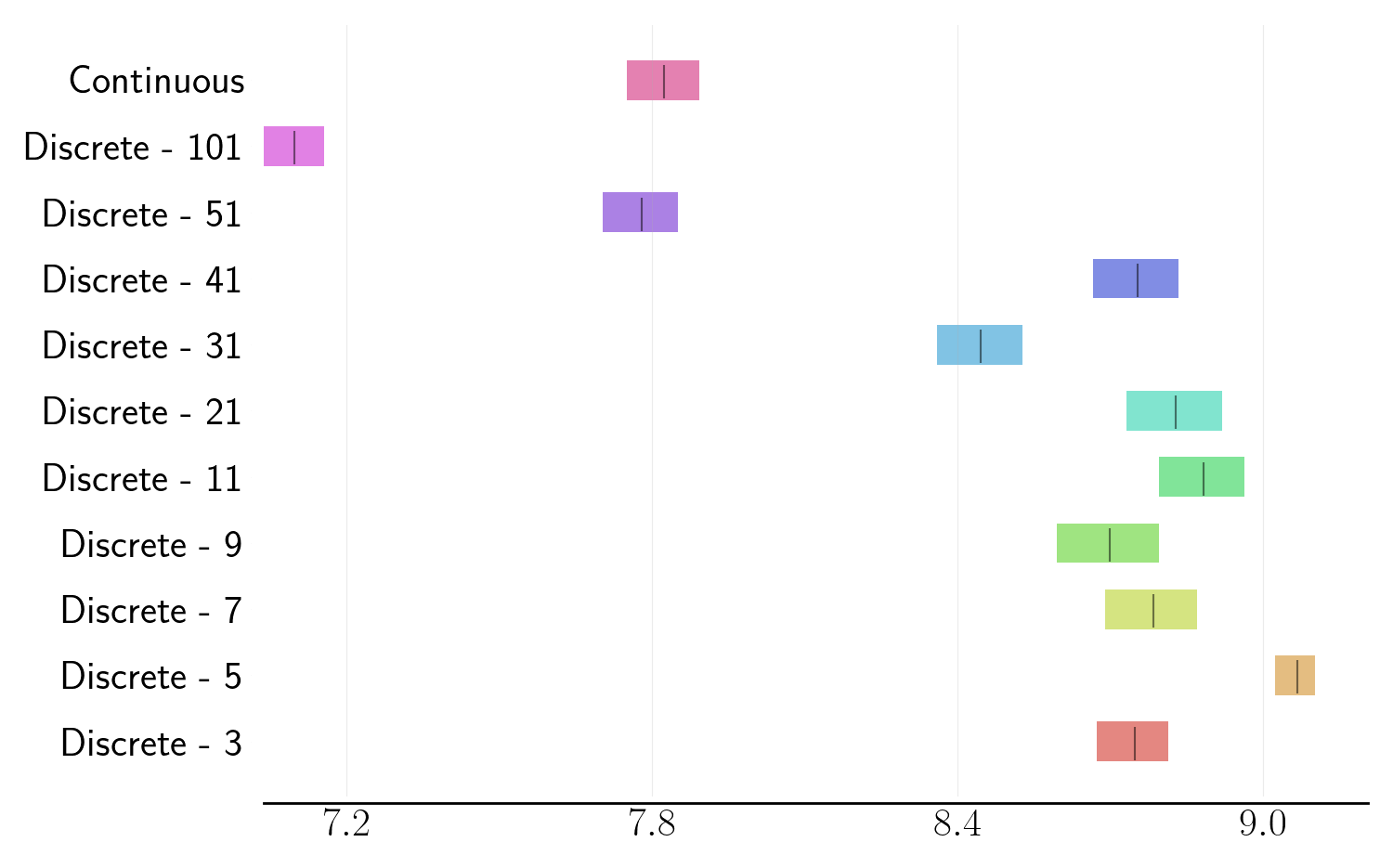}}
% \subfigure[Total reward with $\controlmax=0.1$N]{\includegraphics[width=.48\linewidth]{figures/inspection/0_1/TotalReward_int_est.png}}
\subfigure[Total reward in the inspection environment with $\controlmax=0.1$N.]{\includegraphics[width=.48\linewidth]{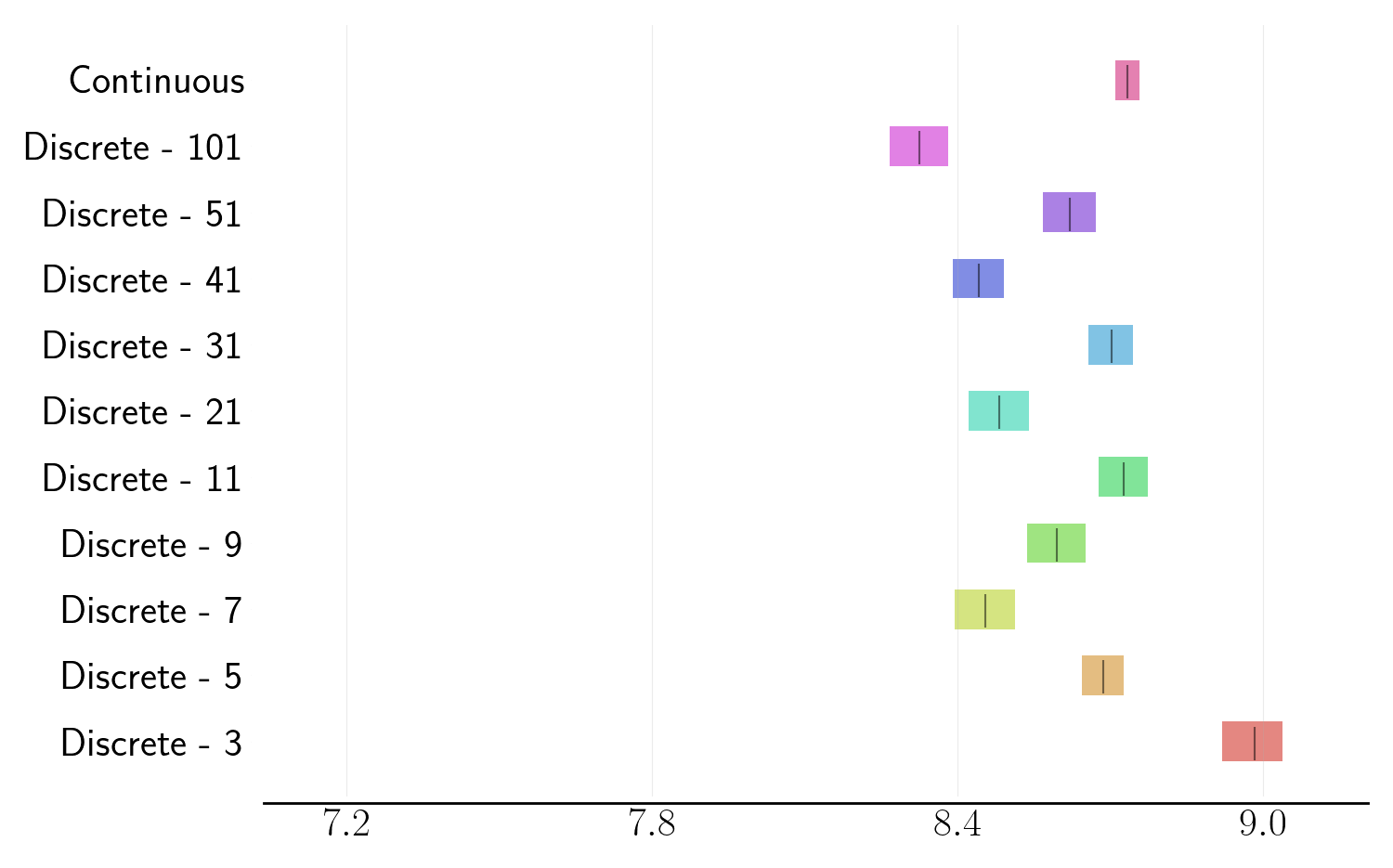}}
% \subfigure[Inspected points with $\controlmax=1.0$N]{\includegraphics[width=.48\linewidth]{figures/inspection/1_0/InspectedPoints_int_est.png}}
% \subfigure[Inspected points with $\controlmax=0.1$N]{\includegraphics[width=.48\linewidth]{figures/inspection/0_1/InspectedPoints_int_est.png}}
\caption{Comparison of the total reward for the final policies in the inspection environment. Each marker represents the IQM of 100 trials, and the shaded region is the $95\%$ confidence interval about the IQM.}
\label{fig:insp-reward-int-est}
\end{figure*}

% Docking success comparison
\begin{figure*}[ht]
\centering
% \subfigure[Total reward with $\controlmax=1.0N$]{\includegraphics[width=.48\linewidth]{figures/docking/1_0/TotalReward_int_est.png}}
% \subfigure[Total reward with $\controlmax=0.1N$]{\includegraphics[width=.48\linewidth]{figures/docking/0_1/TotalReward_int_est.png}}
\subfigure[Success rate in the docking environment with $\controlmax=1.0$N.]{\includegraphics[width=.48\linewidth]{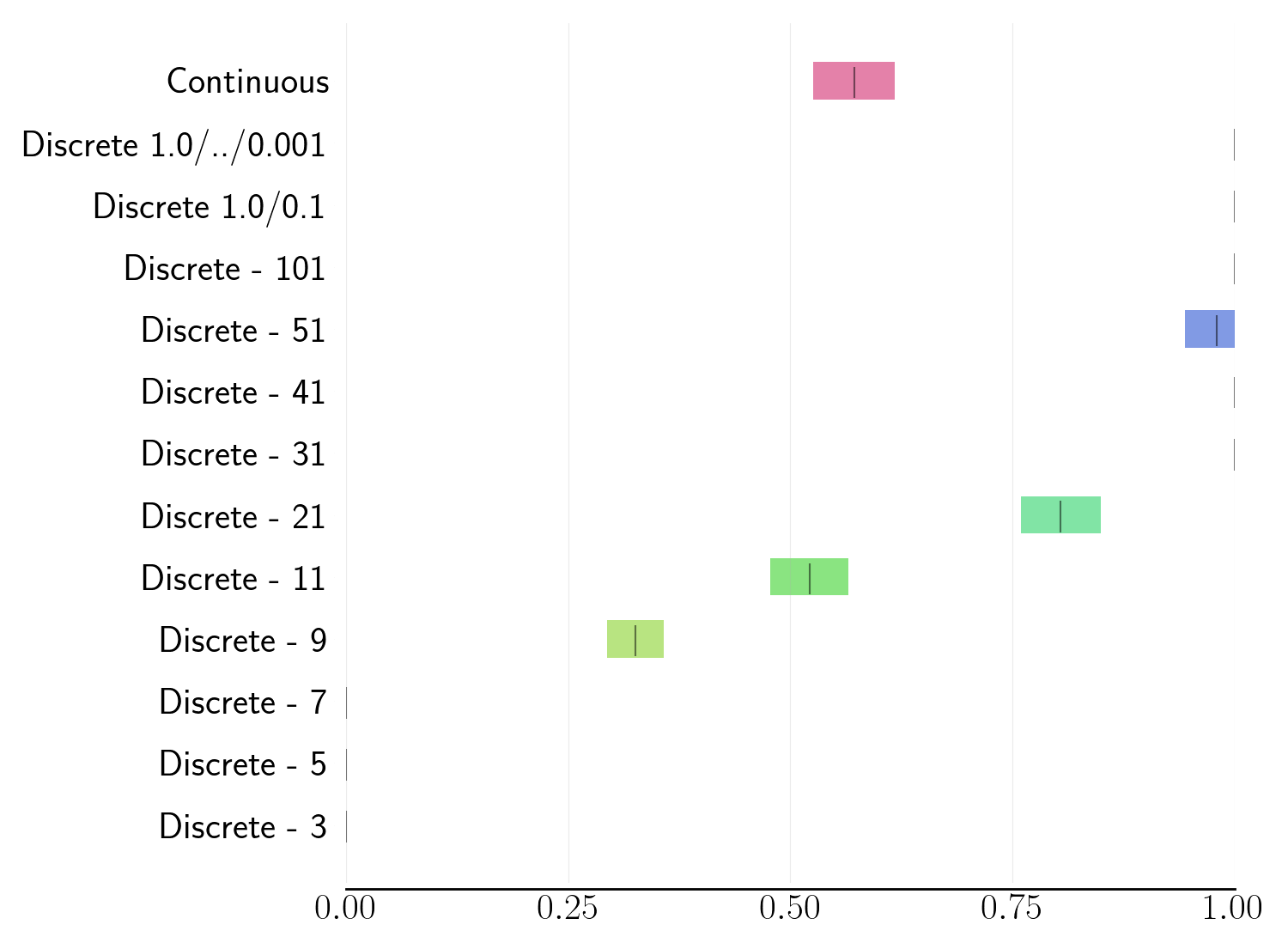}}
\subfigure[Success rate in the docking environment with $\controlmax=0.1$N.]{\includegraphics[width=.48\linewidth]{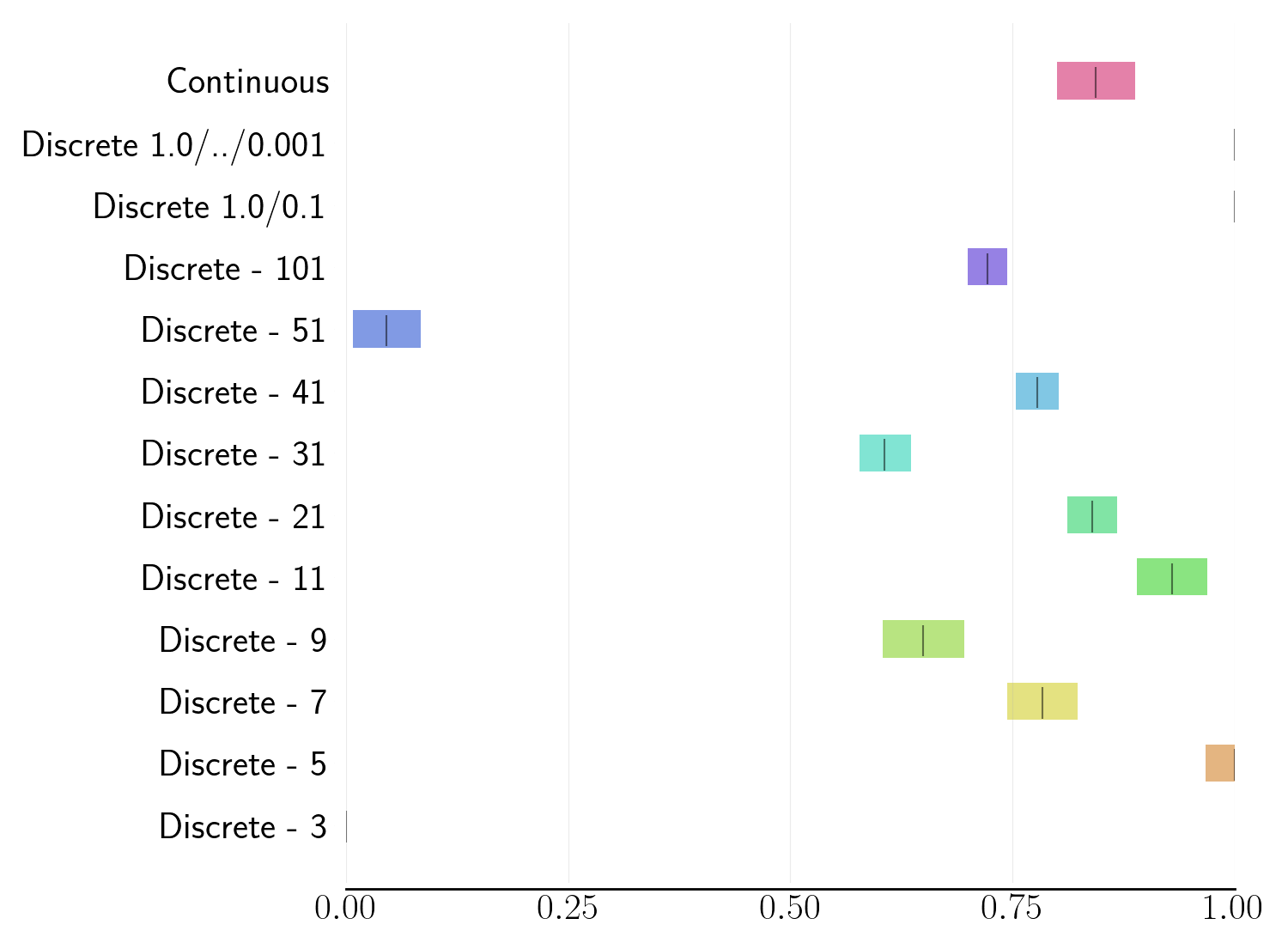}}
\caption{Comparison of the success rate for the final policies in the docking environment. Each marker represents the IQM of 100 trials, and the shaded region is the $95\%$ confidence interval about the IQM.}
\label{fig:dock-reward-int-est}
\end{figure*}

To answer this question, we first analyze what actions the agents take most often. \figref{fig:histogram} shows the percentage that each action is taken for the different experiments. For the inspection environment with $\controlmax=1.0$N, close to 100\% of all actions taken are zero or very close to zero. However when $\controlmax=0.1$N, the agents often choose actions that are either zero or $\pm \controlmax$. For the docking environment with $\controlmax=1.0$N, while many actions taken are close to zero, there is a clear bell curve that appears centered on zero control. When $\controlmax=0.1$N, this bell curve becomes more apparent, but the agents still tend to choose actions that are close to zero. These results suggest that action magnitude is more important in solving the inspection task, while granularity is more important in solving the docking task.

\subsubsection{Inspection}

For the inspection task, the total reward for the final policies for each configuration is shown in \figref{fig:insp-reward-int-est}. This represents the agent's ability to balance the objectives of inspecting all points and reducing $\deltav$ use. For the agents trained with $\controlmax=1.0$N, it can be seen that all agents trained with 41 or less choices result in similar final reward. Recall that in \figref{subfig:delta-v-1.0-insp}, it can be seen that the lowest $\deltav$ for the inspection environment with $\controlmax=1.0$N occurs for the agent trained with 21 discrete actions. Notably, this is the configuration with the least number of choices where the agent can choose $u=0.1$N.

For the agents trained with $\controlmax=0.1$N, it can be seen that reward tends to decrease slightly as the number of choices increases. The agent trained with three choices results in the highest reward, and \figref{subfig:delta-v-0.1-insp} shows that this configuration also results in the lowest $\deltav$ use. These results show that using a smaller action magnitude is much more important than increasing the granularity of choices for the inspection task. This result is intuitive as the agent does not need to make precise adjustments to its trajectory to complete the task, and can rely on following a general path to orbit the chief and inspect all points.

\subsubsection{Docking}

For the docking task, the success rate for the final policies for each configuration is shown in \figref{fig:dock-reward-int-est}. For the agents trained with $\controlmax=1.0$N, the agents with more choices have the highest success rates. For the agents trained with $\controlmax=0.1$N, outside of the agents with 3 and 51 choices, all other configurations result in similar success rates. Along with \figref{fig:delta-v-int-est}(c)\&(d), this shows that more choice and finer granularity leads to higher $\deltav$ use, but it is necessary to successfully complete the task.

However, there is one notable exception to this trend: the agents trained with continuous actions (highest granularity). It can be seen that this configuration uses by far the least $\deltav$ out of all agents that achieved at least a 50\% success rate. In particular, the agent trained with continuous actions and $\controlmax=0.1$N achieves the best balance of high success with low $\deltav$. These results show that increasing the granularity of choices is much more important than using a smaller action magnitude for the docking task. This result is intuitive because the agent must be able to make precise adjustments to its trajectory as it approaches and docks with the chief.

%%%%%%%%%%%%%%%%%%%%%%%%%%%%%%%%%%%%%%%%%%%%%%%%%%%%%%%%%%%%%%%%%%%%%%%%%%%%%%%%%%
% Sweet sweet balance
%%%%%%%%%%%%%%%%%%%%%%%%%%%%%%%%%%%%%%%%%%%%%%%%%%%%%%%%%%%%%%%%%%%%%%%%%%%%%%%%%%
\subsection{\Paste{q_text:num_choices}}\label{sec:q-num_choices}
\noindent \textbf{Answer:} No, for these tasks it is better to choose either discrete or continuous actions.

% \todo{tie in value of looking at observed behavior to identify crucial actions and refine decisions on action space: see docking 1-0.1}
% \nph{I love the direction this is going and we should include the newest attempt at blending discrete and continuous. Have the best discrete results, continuous, and our attempts to beat it. That will show how we tried to find an optimal balance and the result is really just to pick one once you've identified what optimal should be like (bell vs W).}

% Inspection epiosodes
\begin{figure*}[ht]
\centering
\subfigure[Continuous actions with $\controlmax=1.0$N.]{\includegraphics[width=0.48\linewidth]{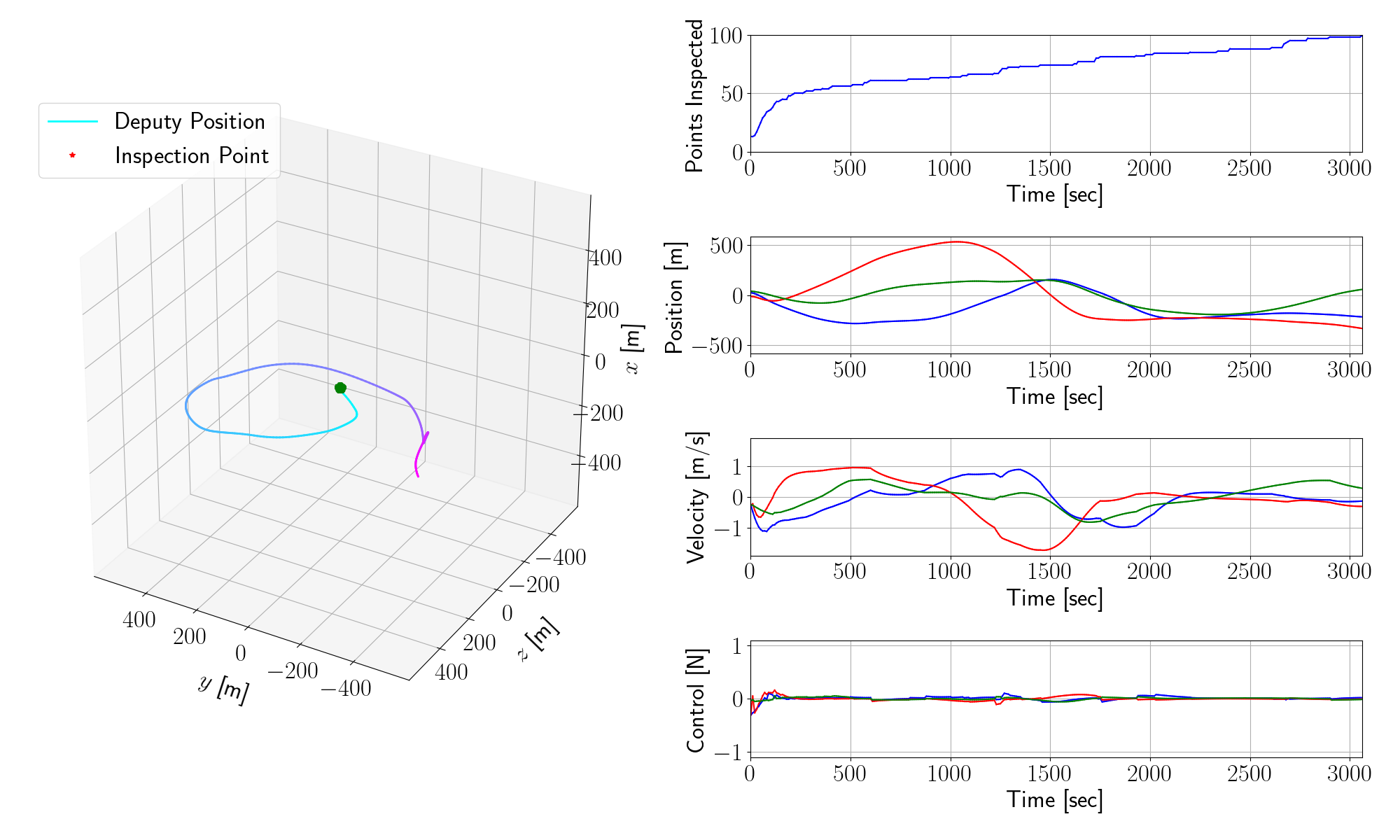}\label{subfig:insp-ep-plot-con_1.0}}\qquad
\subfigure[3 discrete actions with $\controlmax=1.0$N.]{\includegraphics[width=0.48\linewidth]{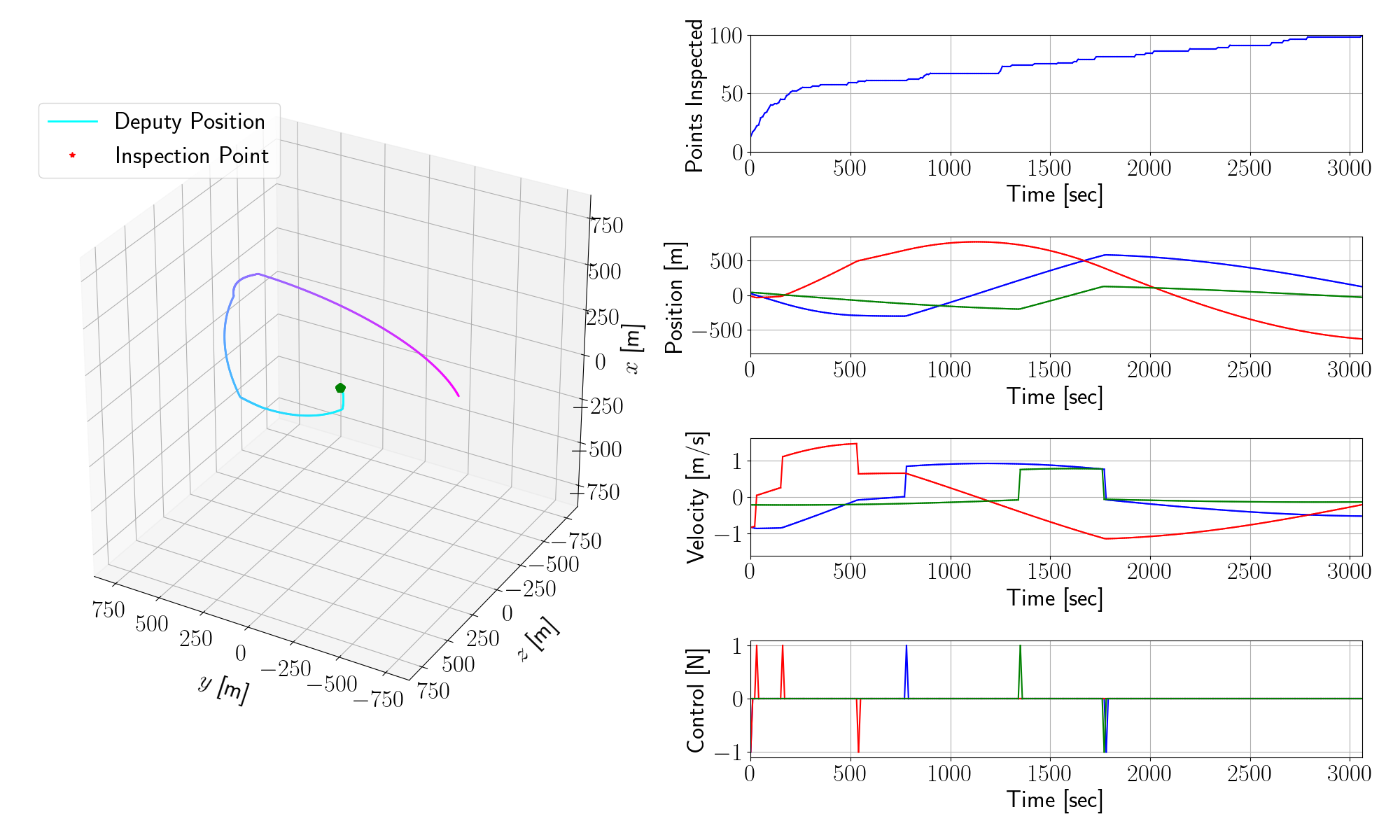}\label{subfig:insp-ep-plot-dis_1.0_3}}\qquad
\subfigure[101 discrete actions with $\controlmax=1.0$N.]{\includegraphics[width=0.48\linewidth]{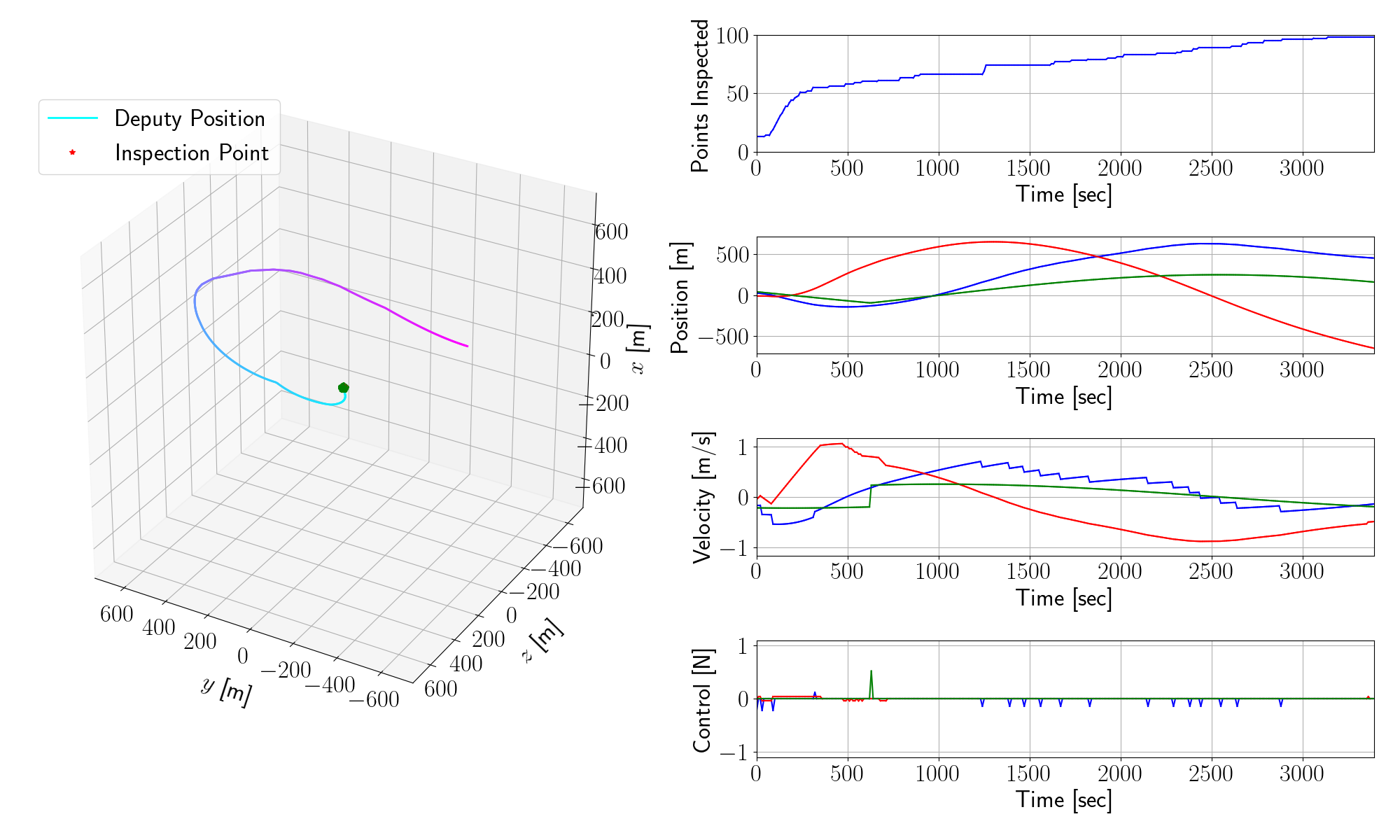}\label{subfig:insp-ep-plot-dis_1.0_101}}
\subfigure[3 discrete actions with $\controlmax=0.1$N.]{\includegraphics[width=0.48\linewidth]{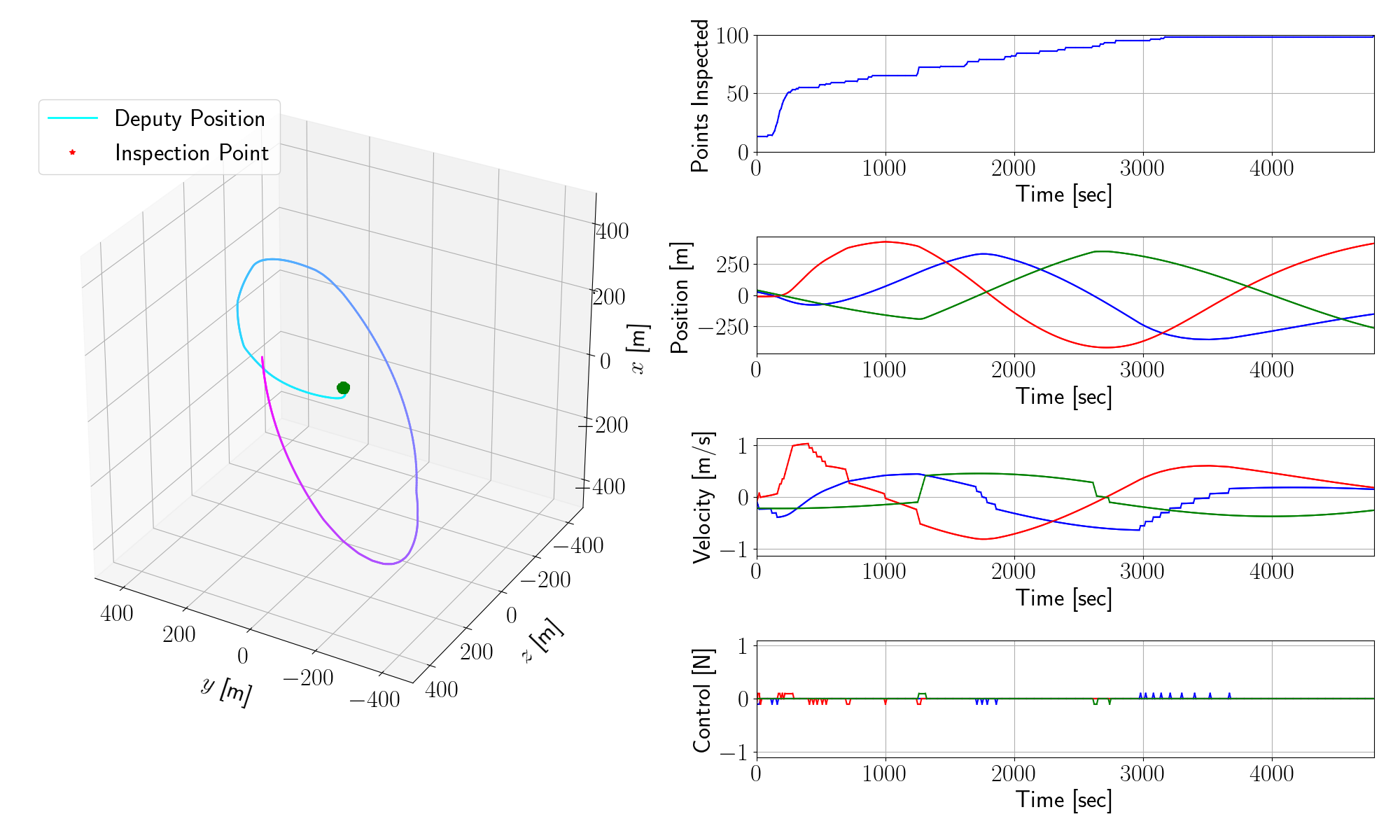}\label{subfig:insp-ep-plot-dis_0.1_3}}
\caption{Example episodes for policies trained in the inspection environment. The trajectory changes from light to dark as time progresses. Blue, red, and green lines represent the $x$, $y$, and $z$ components respectively.}
\label{fig:inspection_ep_plots_1}
\end{figure*}

% Docking episodes
\begin{figure*}[ht]
\centering
\subfigure[Continuous actions with $\controlmax=1.0$N.]{\includegraphics[width=0.48\linewidth]{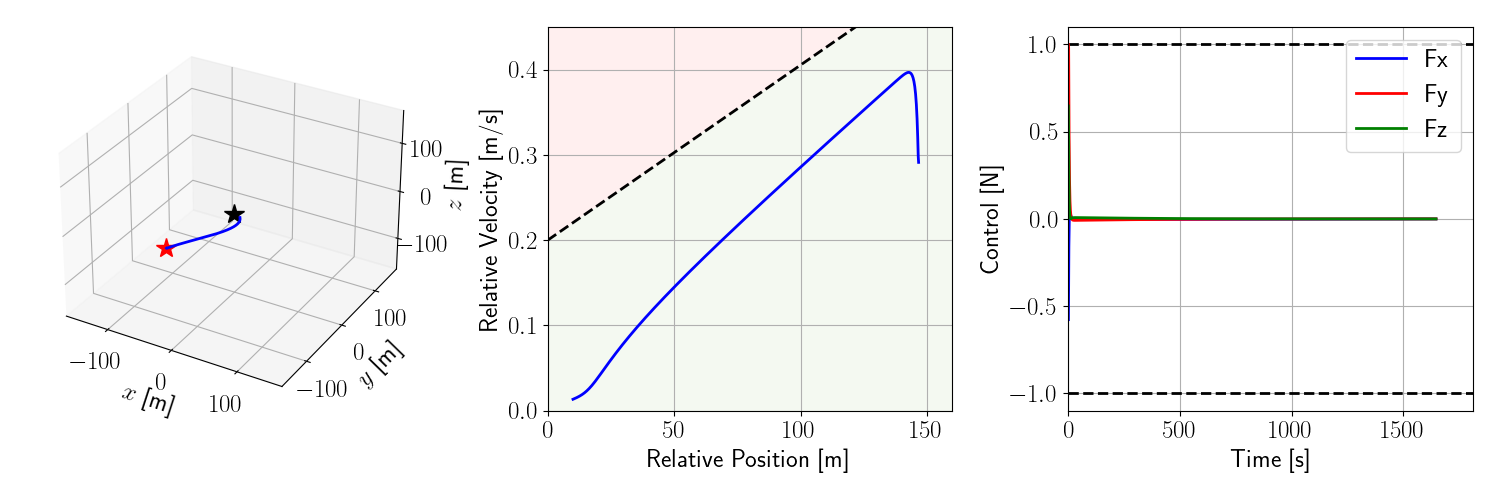}}\qquad
\subfigure[3 discrete actions with $\controlmax=1.0$N.]{\includegraphics[width=0.48\linewidth]{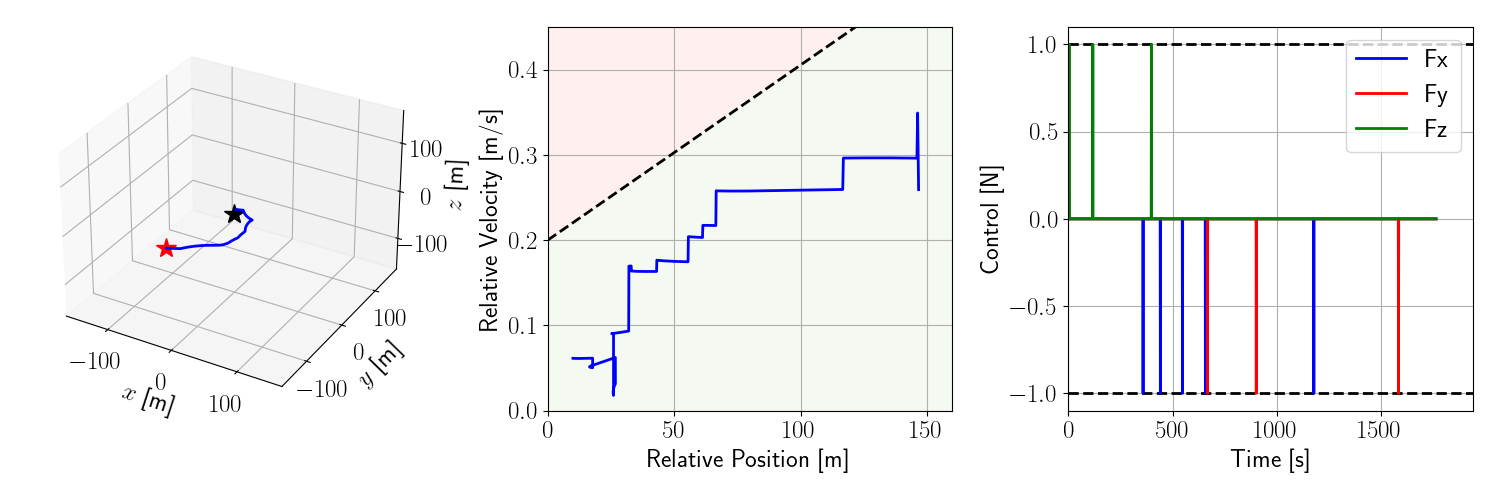}}\qquad
\subfigure[101 discrete actions with $\controlmax=1.0$N.]{\includegraphics[width=0.48\linewidth]{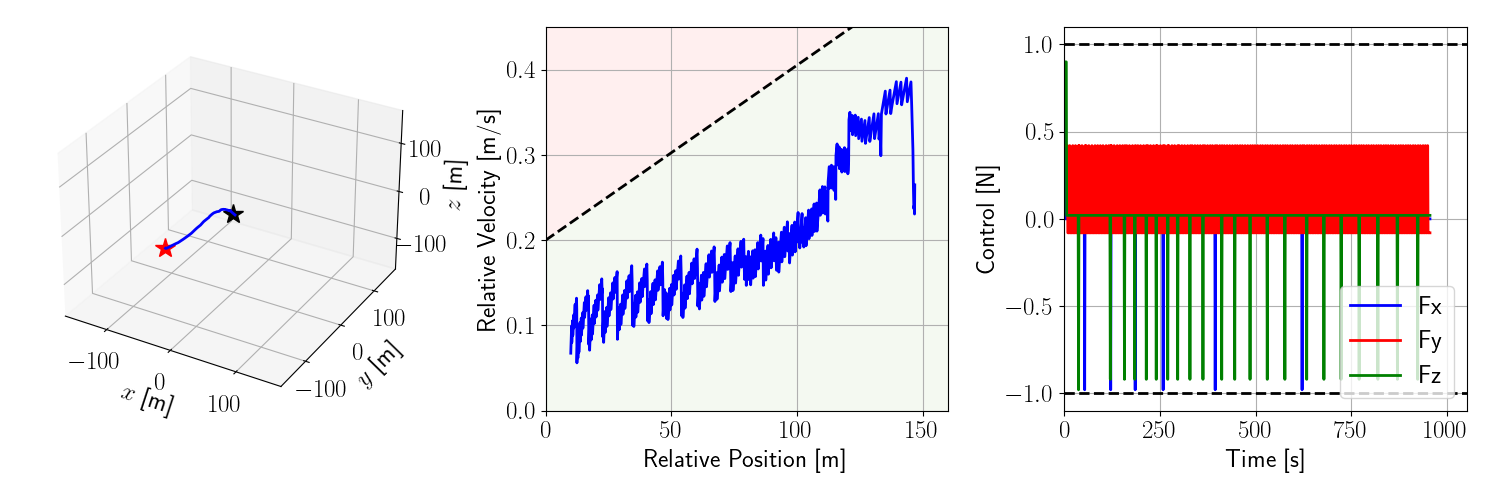}}
\subfigure[3 discrete actions with $\controlmax=0.1$N.]{\includegraphics[width=0.48\linewidth]{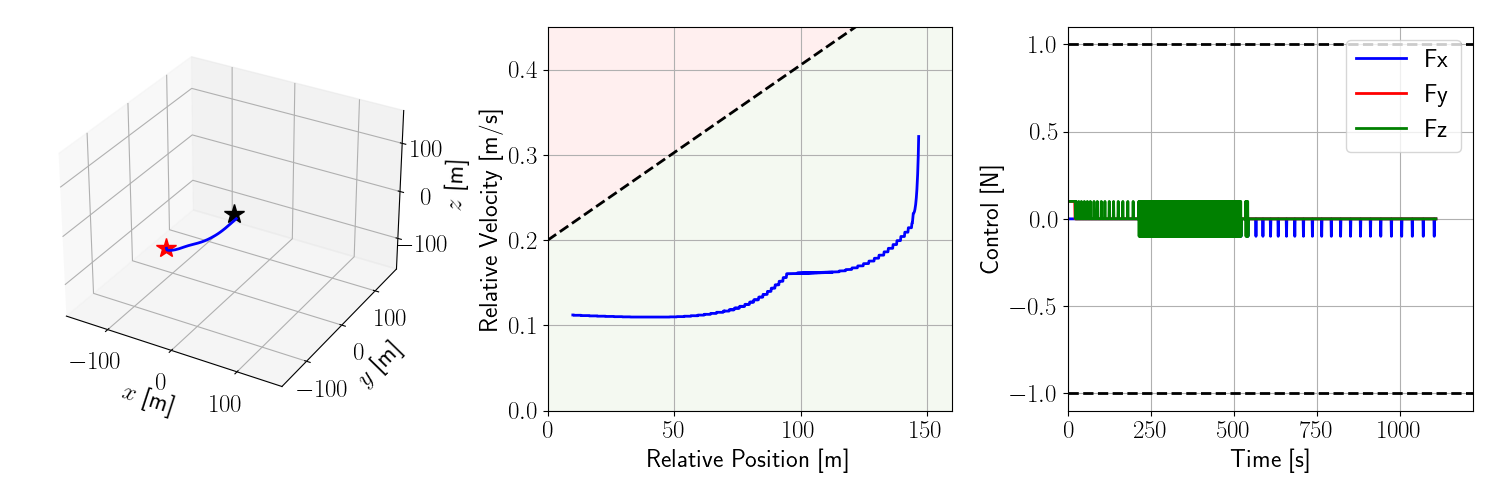}}
\subfigure[5 discrete actions: $1.0/0.1$ configuration.]{\includegraphics[width=0.48\linewidth]{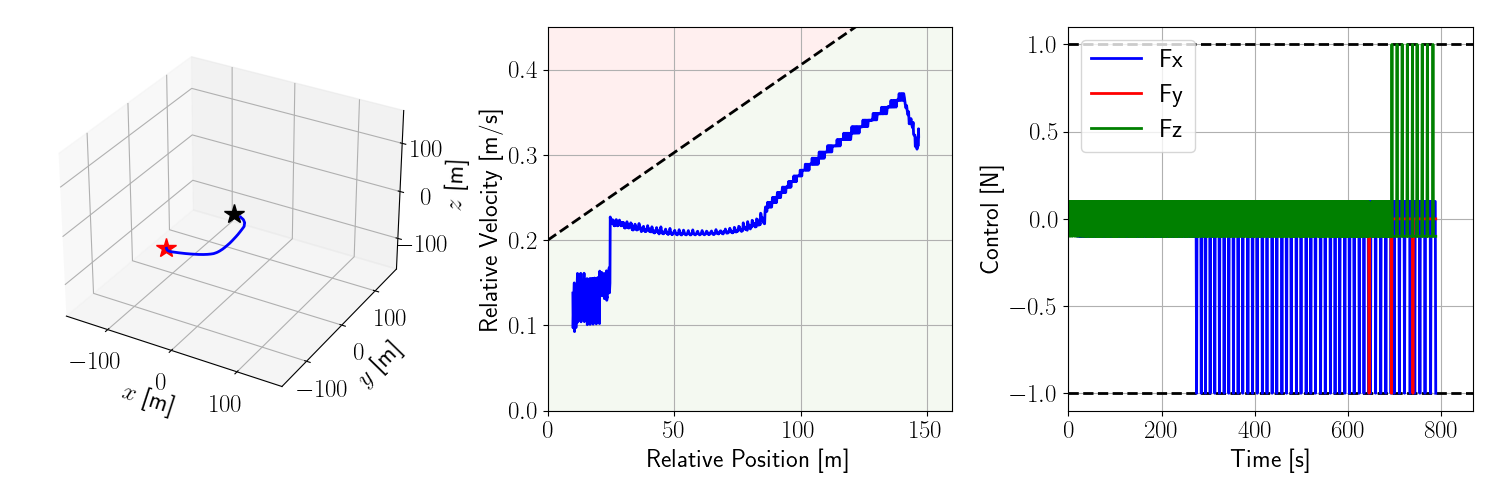}\label{subfig:dock-ep-plot-1.0/0.1}}
\subfigure[9 discrete actions: $1.0/../0.001$ configuration.]{\includegraphics[width=0.48\linewidth]{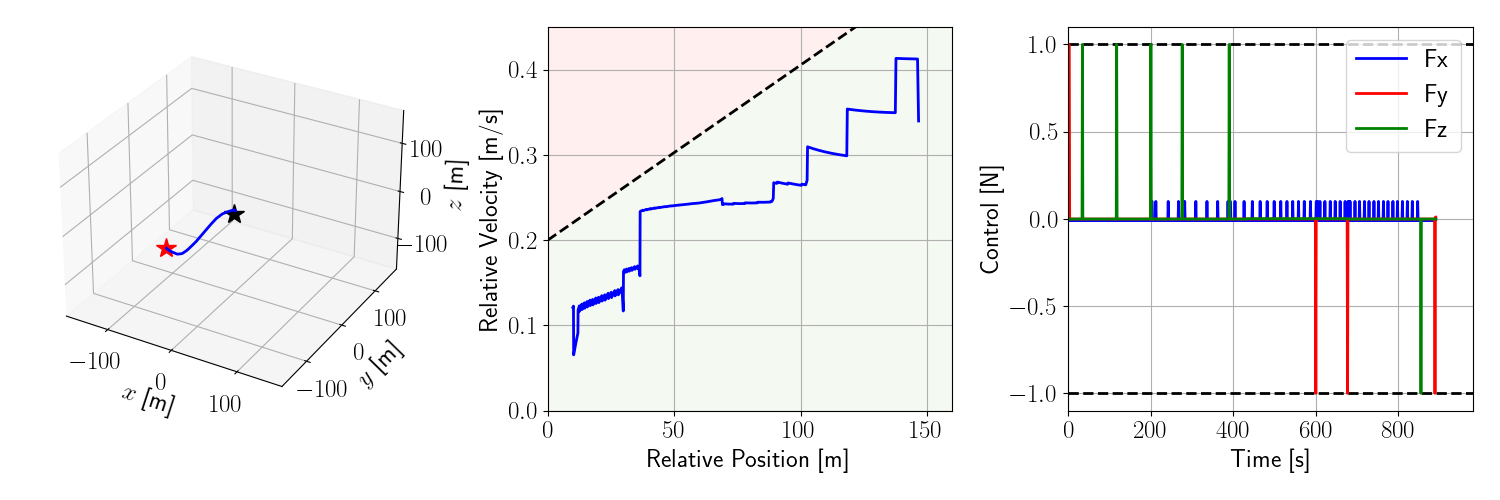}\label{subfig:dock-ep-plot-1.0//0.001}}
\caption{Example episodes for policies trained in the docking environment. For the trajectories, the chief's position is represented by a black star, and the deputy's initial position is represented by a red star. For the distance dependent speed limit, safe regions are shaded green and unsafe regions are shaded red.}
\label{fig:docking_ep_plots_1}
\end{figure*}

To answer this question, we consider the behavior of the trained agents. \figref{fig:inspection_ep_plots_1} shows example trajectories of trained agents in the inspection task. Ideally, these agents will circumnavigate the chief along a smooth trajectory. In \figref{subfig:insp-ep-plot-con_1.0}, it can be seen that this is easily accomplished when the agent uses continuous actions, as it can constantly make small refinements to the trajectory to keep it smooth. On the other hand, when the agent can only use three discrete actions with $\controlmax=1.0$N as shown in \figref{subfig:insp-ep-plot-dis_1.0_3}, the trajectory becomes far less smooth. It can be seen that the agent jerks back and forth as it attempts to adjust its trajectory.

To balance continuous and discrete actions, the number of discrete choices can be increased to allow the agent to make smaller adjustments to its trajectory. As seen in \figref{subfig:insp-ep-plot-dis_1.0_101}, having 101 choices helps the agent's trajectory become much smoother. However, as answered by question \qref{question:granularity}, this comes at a cost of performance. The optimal performance for the inspection task came with three discrete actions with $\controlmax=0.1$N. In \figref{subfig:insp-ep-plot-dis_0.1_3}, it can be seen that this configuration also results in a much smoother trajectory, as the agent can make smaller adjustments to its trajectory due to the smaller $\controlmax$. Therefore, the optimal behavior can also be achieved using discrete actions for the inspection task. This also follows \figref{subfig:histogram-w}, where the actions most commonly used are zero and $\pm \controlmax$.

For the docking task, ideally the agent will slow down and approach the chief along a smooth trajectory. \figref{fig:docking_ep_plots_1} shows example trajectories of trained agents in the docking task, where similar results to the inspection task can be seen. Continuous actions allow for the smoothest trajectory, three discrete actions with $\controlmax=1.0$N is the least smooth trajectory, and 101 discrete actions or three discrete actions with $\controlmax=0.1$N allow for smoother trajectories while using discrete actions. However, there is still a considerable amount of ``chattering" for the control, where the agent frequently switches between multiple control values as it attempts to refine its trajectory.

To best balance continuous and discrete actions, we analyze the behavior shown in \figref{subfig:histogram-bell}, and attempt to provide action choices for the agent that mimic a bell curve. These experiments are the $1.0/0.1$ configuration and the $1.0/../0.001$ configuration. These configurations allow the agent to use actions with high magnitudes when it is far from the chief, but small magnitudes when it gets closer to the chief. From \figref{subfig:delta-v-0.1-dock} and \figref{fig:dock-reward-int-est}, it can be seen that the $1.0/../0.001$ configuration achieves much lower $\deltav$ than the $1.0/0.1$ configuration, with both having 100\% success. These configurations also achieve lower $\deltav$ than most discrete action experiments, but still do not perform as well as the agent trained with continuous actions. As shown in \figref{subfig:dock-ep-plot-1.0/0.1} and \figref{subfig:dock-ep-plot-1.0//0.001}, these configurations also do not have as smooth of trajectories as the agent with continuous actions, and there is still frequent chattering in the control. Therefore, despite attempting to balance discrete and continuous actions, the optimal behavior for the docking task is still achieved using continuous actions.

%%%%%%%%%%%%%%%%%%%%%%%%%%%%%%%%%%%%%%%%%%%%%%%%%%%%%%%%%%%%%%%%%%%%%%
% Conclusions
%%%%%%%%%%%%%%%%%%%%%%%%%%%%%%%%%%%%%%%%%%%%%%%%%%%%%%%%%%%%%%%%%%%%%%
\section{Conclusions and Future Work}

In this paper, we trained 480 unique agents to investigate how choice impacts the learning process for space control systems.
% In conclusion, the results show that (\qref{question:no-op}) it is very important for the agent to easily be able to choose not to thrust.
In conclusion, the results show that (\qref{question:no-op}) increasing the likelihood of selecting ``no thrust" as an action greatly reduces $\deltav$ use.
By either making zero thrust a more likely choice, or reducing the action magnitude to make choices closer to zero, this significantly reduces the $\deltav$ used by the agent. Next, our results indicate that (\qref{question:granularity}) increasing granularity of choices or adjusting action magnitude for optimal performance is highly dependent on the task. For the inspection task, selecting an appropriate action magnitude is more important than increasing granularity of choices. It was found the optimal configuration for completing the inspection task was three discrete actions with $\controlmax=0.1$N. For the docking task, the opposite is true, where the optimal configuration was continuous actions with $\controlmax=0.1$N.
This makes sense considering the operating range of the tasks, where the agent can complete the inspection task by orbiting the chief at a further relative distance, while the agent must complete the docking task by making small adjustments to its trajectory as it approaches the docking region. Finally, our results show that (\qref{question:num_choices}) there is not an optimal balance between discrete and continuous actions, and it is better to choose one or the other. When attempting to balance discrete and continuous actions for the docking environment by providing actions with decreasing magnitude, it was found that this configuration performed better than most discrete action configurations, but it still did not perform as well as agents with continuous actions.

% \todo{Future work}
In future work, we want to consider more complex six degree-of-freedom dynamics, where the agent can also control its orientation. We also want to explore more complex discrete action choices, including adding a time period for the thrust selection to better replicate a scheduled burn. 
% \todo{anything else?}

% (\ref{question:num_choices}) for the inspection task, too many choices led to worse performance (where the agent used more fuel), but for the docking task, more choices were needed for better performance (the agent must make small adjustments to complete the task). Next, it was found that (\ref{question:granularity}) adjusting the action magnitude was more effective than adjusting the granularity. In general, most policies were more successful and used less fuel with smaller action magnitudes, while finer granularity leading to more choices did not always improve performance. Finally, it was found that (\ref{question:no-op}) it is very important for the agent to easily be able to choose not to thrust. For the inspection environment, less choices makes it easier to choose not to thrust, which improved performance. For the docking environment, zero thrust is still an important and frequent choice even when the agent uses continuous actions.
% 
% Overall, it was found that for the inspection environment, 3 discrete actions with $\controlmax=0.1$N provide the best performance, and for the docking environment, continuous actions with $\controlmax=0.1$N provide the best performance.

\section*{Acknowledgements}
This research was sponsored by the Air Force Research Laboratory under the Safe Trusted Autonomy for Responsible Spacecraft (STARS) Seedlings for Disruptive Capabilities Program.
The views expressed are those of the authors and do not reflect the official guidance or position of the United States Government, the Department of Defense, or of the United States Air Force. This work has been approved for public release: distribution unlimited. Case Number AFRL-2024-0298.

%%%%%%%%%%%%%%%%%%%%%%%%%%%%%%%%%%%%%%%%%%%%%%%%%%%%%%%%%%%%%%%%%%%%%%%%%%%
% References
%%%%%%%%%%%%%%%%%%%%%%%%%%%%%%%%%%%%%%%%%%%%%%%%%%%%%%%%%%%%%%%%%%%%%%%%%%%
% \newpage
% \bibliographystyle{acm}
% % argument is your BibTeX string definitions and bibliography database(s)
% \nocite{*}
% \bibliography{references}

\bibliographystyle{ieeetr} 
\bibliography{references}

\newpage
% \appendix
% \appendixpage
\clearpage

\begin{appendices}

\section{Sample Complexity Figures}

Sample complexity is a metric that indicates how ``fast'' an agent trained by periodically checking performance throughout training. If an agent has better sample complexity (i.e. trains ``faster'') then its sample complexity curve will measure better values after fewer timesteps. For a metric like reward, better sample complexity will be closer to the top left of the plot, while a metric like $\deltav$ use will show better sample complexity by being closer to the bottom left corner of the plot.

\begin{figure}[h]
\centering
\subfigure[Total reward.]{\includegraphics[width=0.45\linewidth]{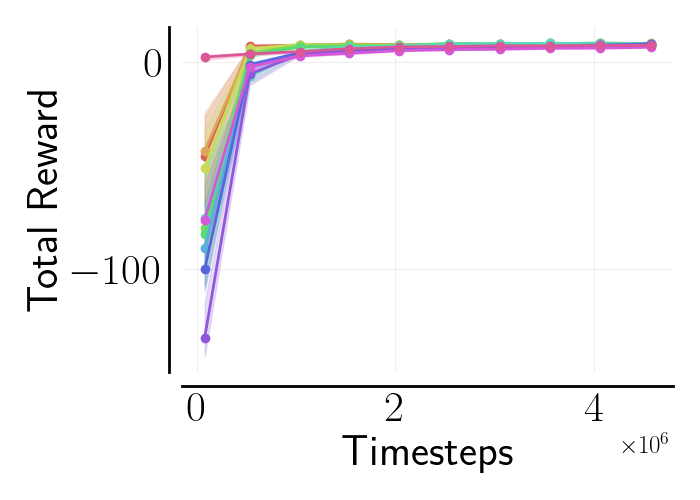}}\qquad
\subfigure[Inspected points.]{\includegraphics[width=0.45\linewidth]{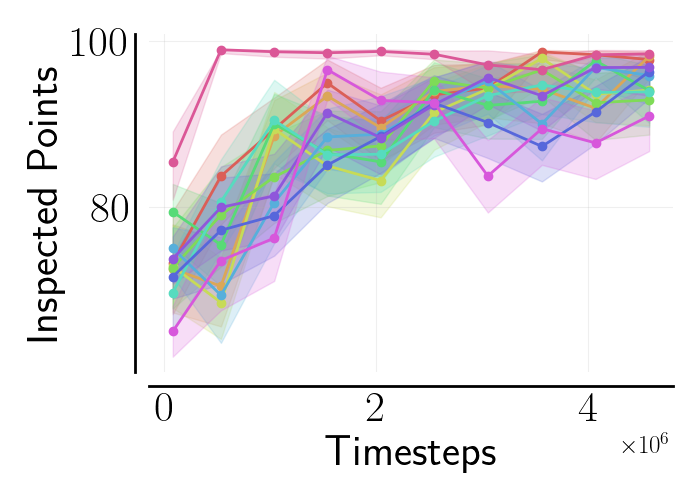}}\\
\subfigure[Success rate.]{\includegraphics[width=0.45\linewidth]{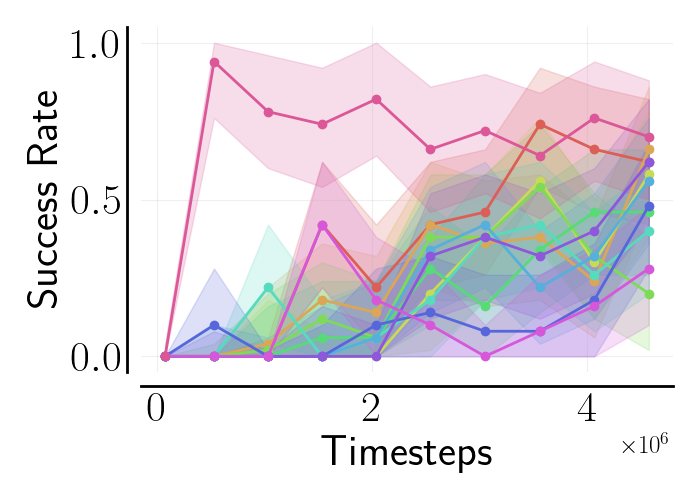}}\qquad
\subfigure[$\deltav$ (m/s).]{\includegraphics[width=0.45\linewidth]{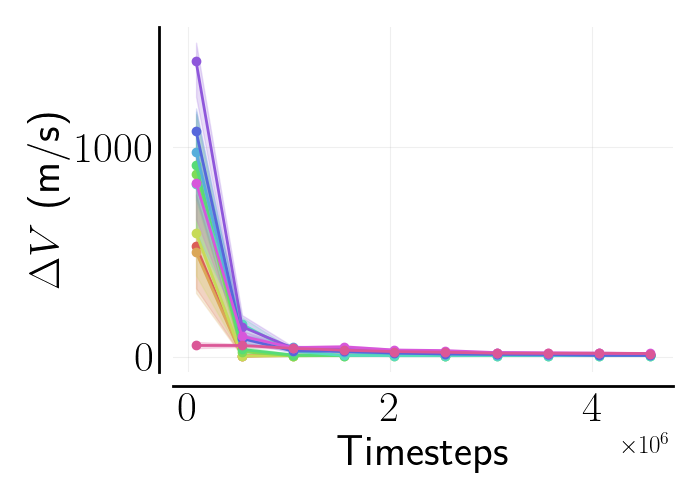}}\\
\subfigure[Episode length (steps).]{\includegraphics[width=0.45\linewidth]{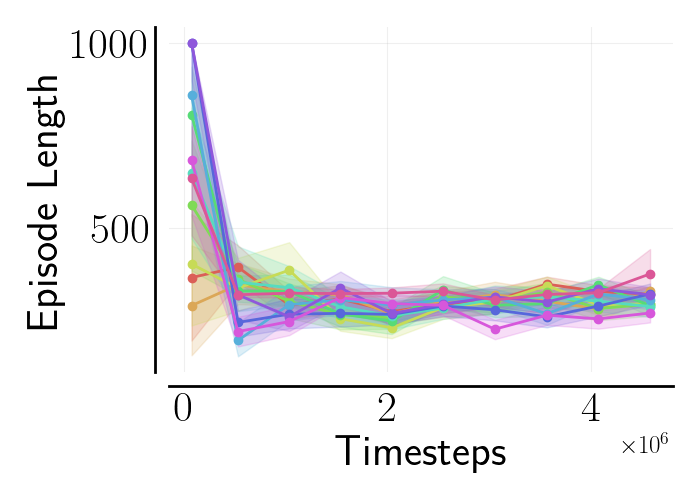}}\qquad
\subfigure{\includegraphics[width=0.45\linewidth]{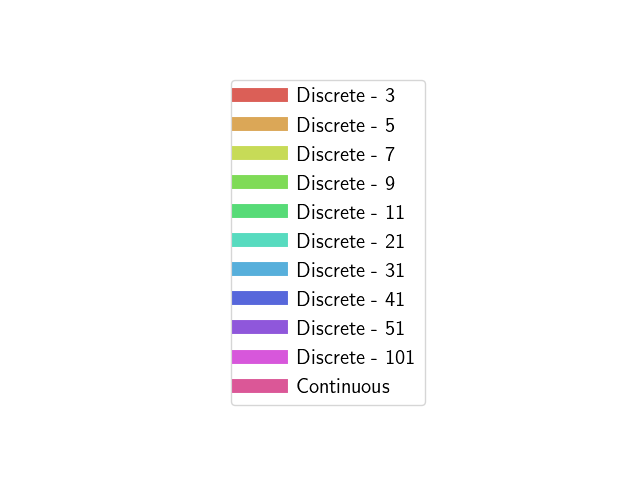}}
\caption{Sample complexity results collected from experiments run in the inspection environment with $\controlmax=1.0$N. Each curve represents the IQM of 10 trials, and the shaded region is the $95\%$ confidence interval about the IQM.}
\label{fig:app_inspection-1.0-sample-efficiency}
\end{figure}

\begin{figure}[h]
\centering
\subfigure[Total reward.]{\includegraphics[width=0.45\linewidth]{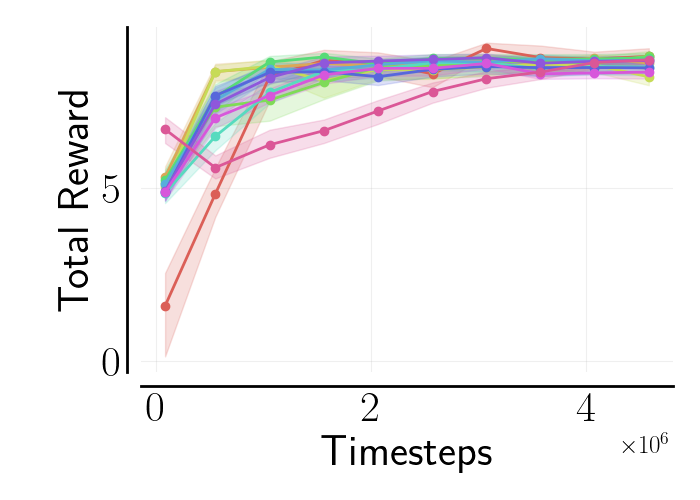}}\qquad
\subfigure[Inspected points.]{\includegraphics[width=0.45\linewidth]{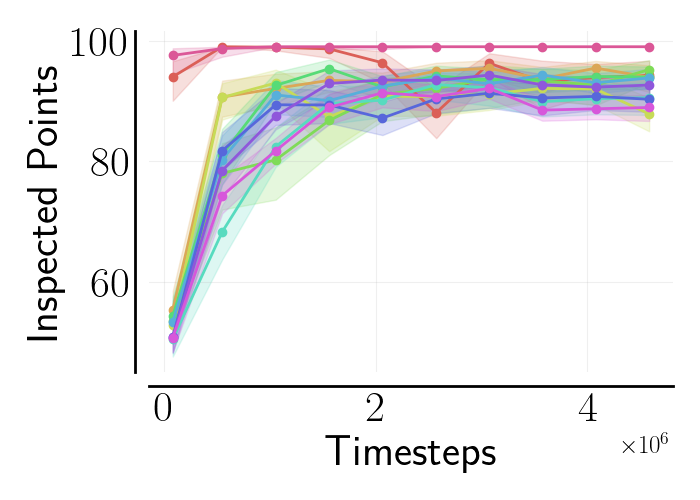}}\\
\subfigure[Success rate.]{\includegraphics[width=0.45\linewidth]{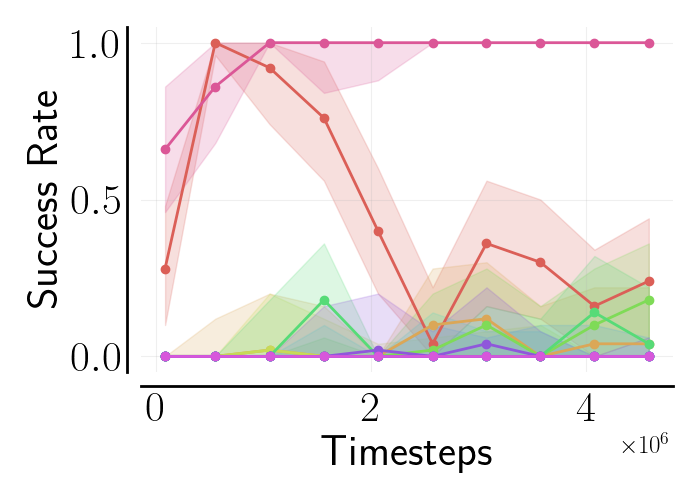}}\qquad
\subfigure[$\deltav$ (m/s).]{\includegraphics[width=0.45\linewidth]{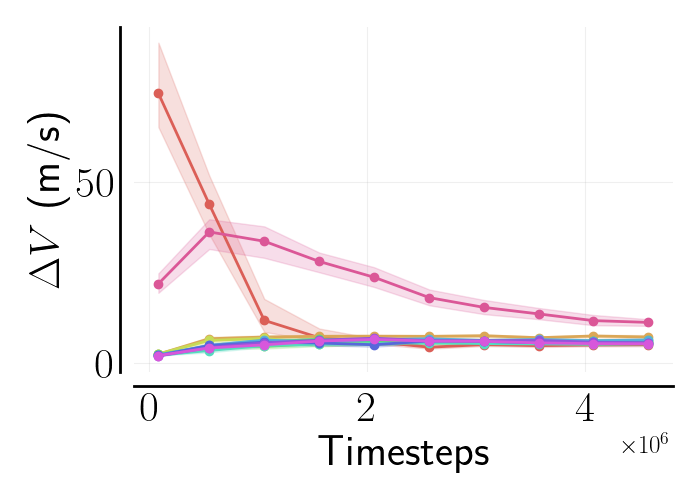}}\\
\subfigure[Episode length (steps).]{\includegraphics[width=0.45\linewidth]{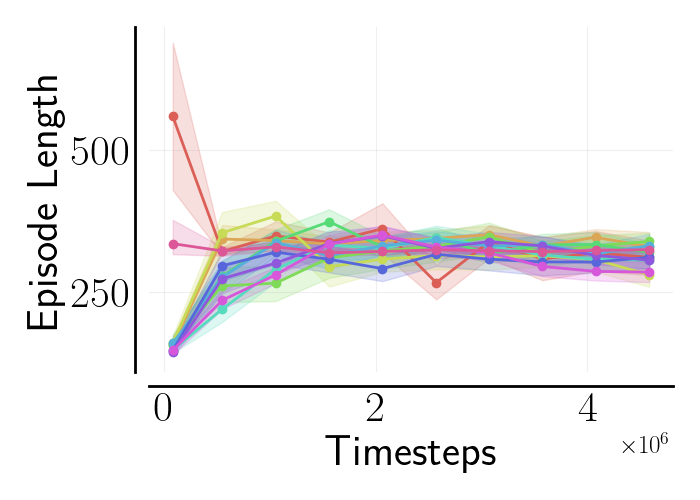}}\qquad
\subfigure{\includegraphics[width=0.45\linewidth]{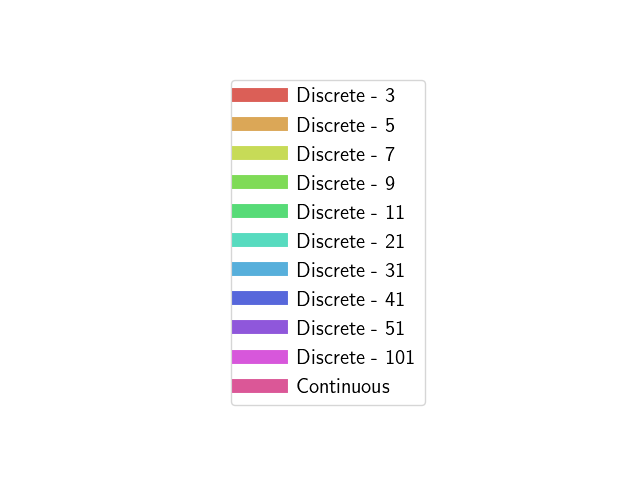}}
\caption{Sample complexity results collected from experiments run in the inspection environment with $\controlmax=0.1$N. Each curve represents the IQM of 10 trials, and the shaded region is the $95\%$ confidence interval about the IQM.}
\label{fig:app_inspection-0.1-sample-efficiency}
\end{figure}

In \figref{fig:app_inspection-1.0-sample-efficiency} and \figref{fig:app_inspection-0.1-sample-efficiency} we show the sample complexity for agents trained in the inspection environment with $\controlmax = 1.0$N and $\controlmax = 0.1$N respectively.

\begin{figure}[h]
\centering
\subfigure[Total reward.]{\includegraphics[width=0.45\linewidth]{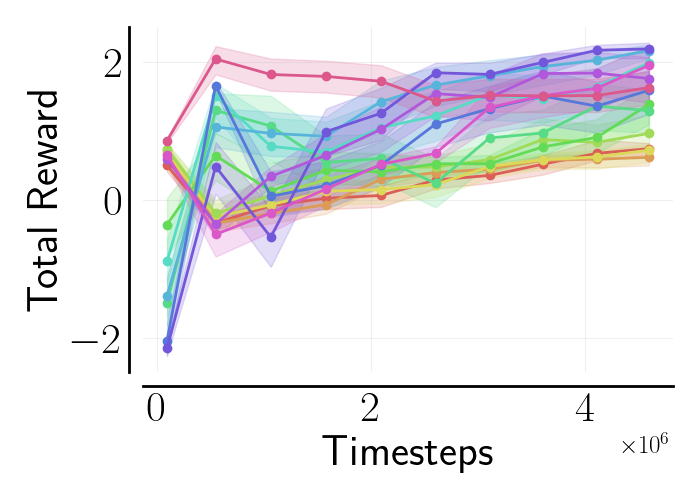}}\qquad
\subfigure[Success rate.]{\includegraphics[width=0.45\linewidth]{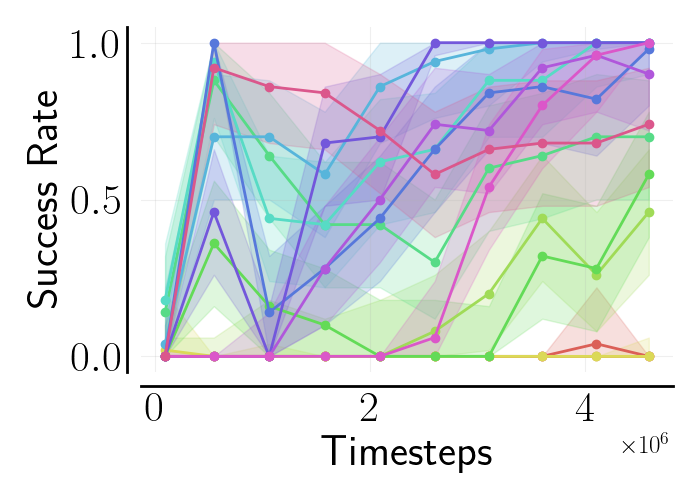}}\\
\subfigure[$\deltav$ (m/s).]{\includegraphics[width=0.45\linewidth]{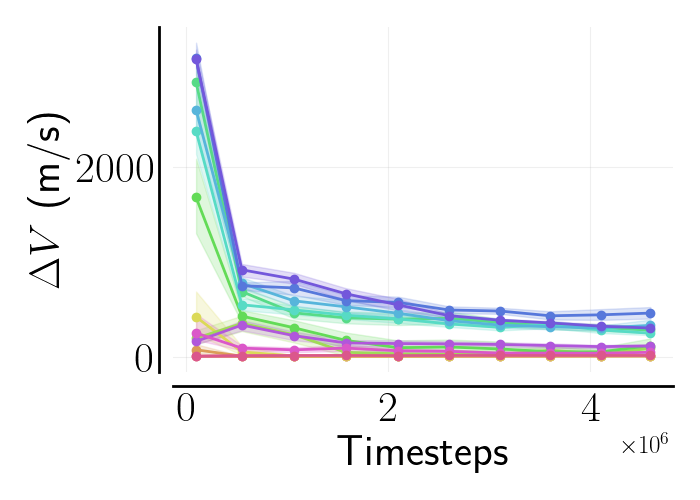}}\qquad
\subfigure[Constraint violation (\%).]{\includegraphics[width=0.45\linewidth]{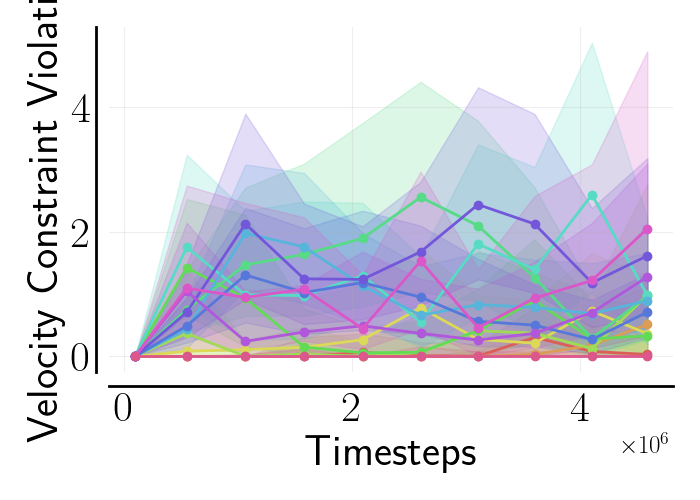}}\\
\subfigure[Final speed (m/s).]{\includegraphics[width=0.45\linewidth]{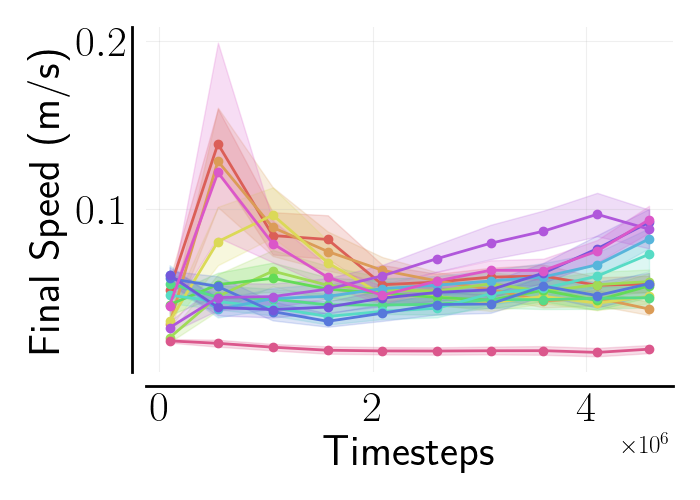}}\qquad
\subfigure[Episode length (steps).]{\includegraphics[width=0.45\linewidth]{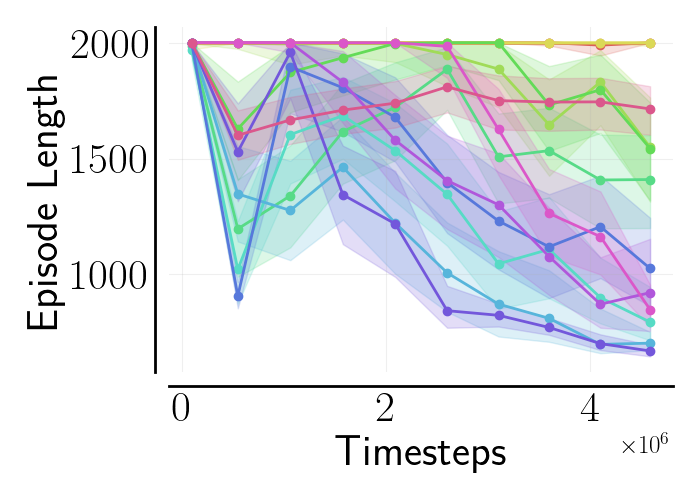}}\\
\subfigure{\includegraphics[width=0.7\linewidth]{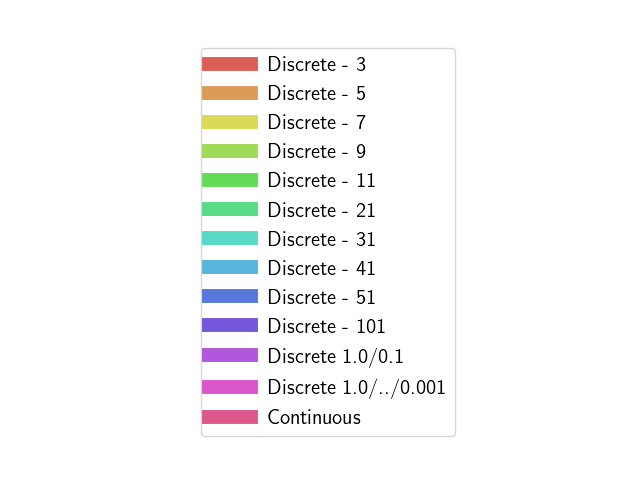}}
\caption{Sample complexity results collected from experiments run in the docking environment with $\controlmax=1.0$N. Each curve represents the IQM of 10 trials, and the shaded region is the $95\%$ confidence interval about the IQM.}
\label{fig:app_docking-1.0-sample-efficiency}
\end{figure}

\begin{figure}[h]
\centering
\subfigure[Total reward.]{\includegraphics[width=0.45\linewidth]{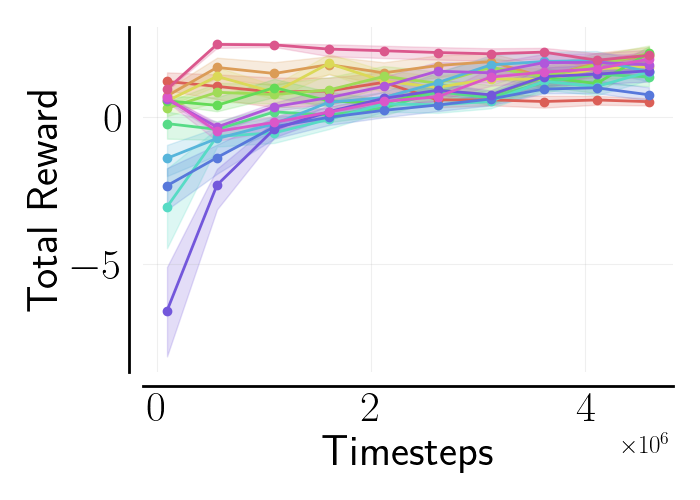}}\qquad
\subfigure[Success rate.]{\includegraphics[width=0.45\linewidth]{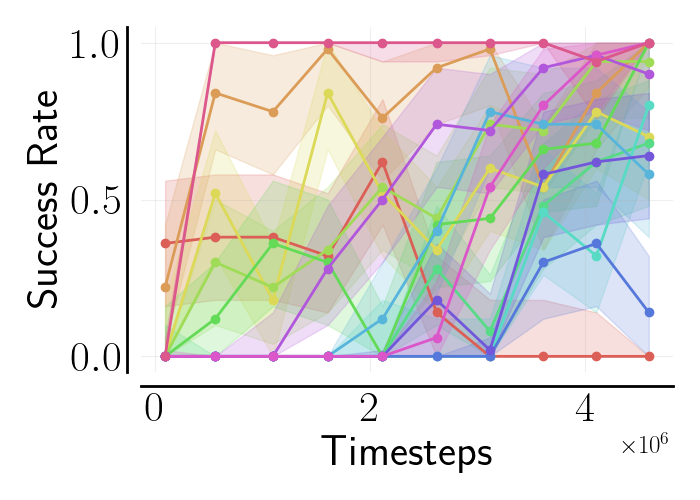}}\\
\subfigure[$\deltav$ (m/s).]{\includegraphics[width=0.45\linewidth]{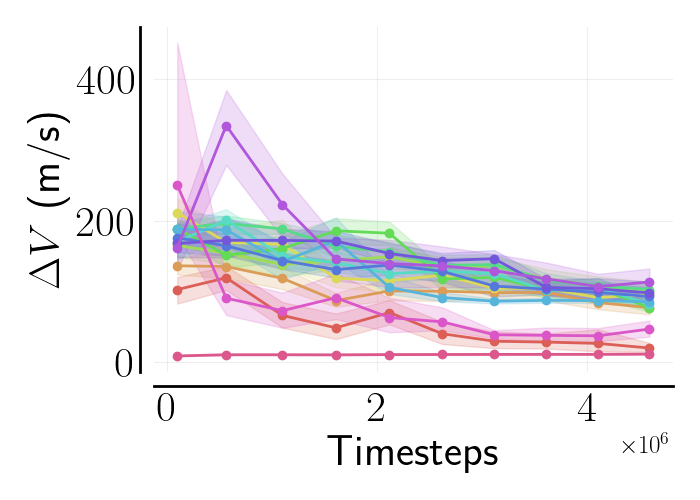}}\qquad
\subfigure[Constraint violation (\%).]{\includegraphics[width=0.45\linewidth]{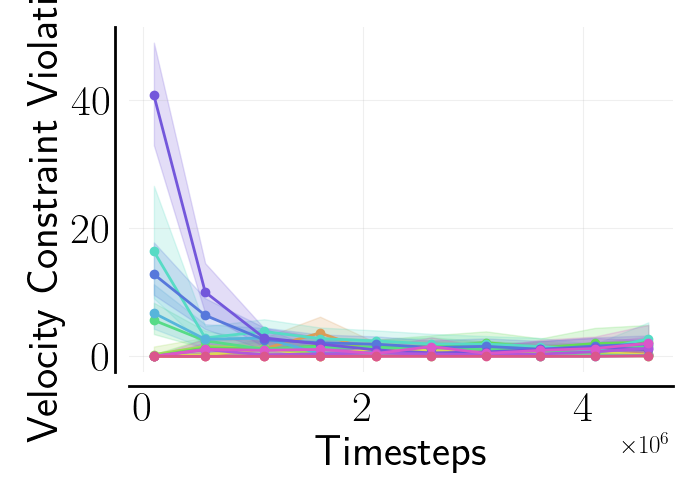}}\\
\subfigure[Final speed (m/s).]{\includegraphics[width=0.45\linewidth]{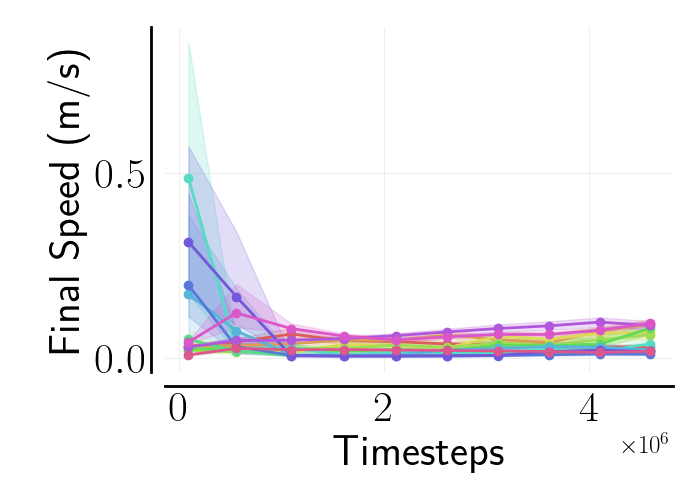}}\qquad
\subfigure[Episode length (steps).]{\includegraphics[width=0.45\linewidth]{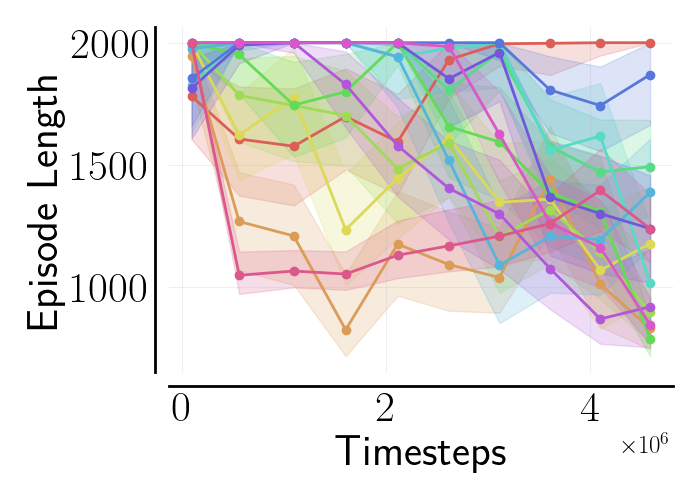}}\\
\subfigure{\includegraphics[width=0.7\linewidth]{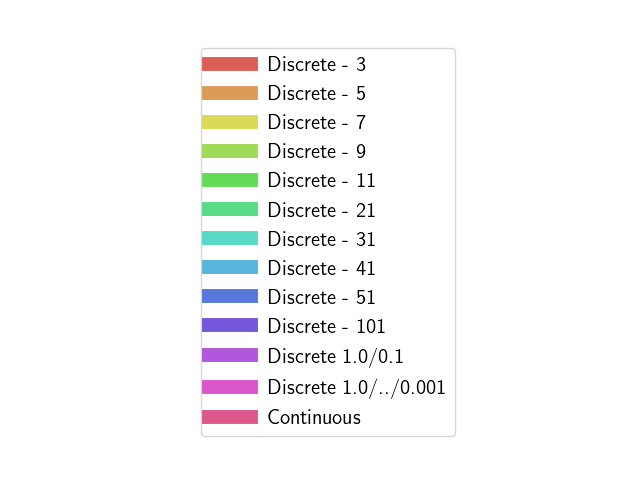}}
\caption{Sample complexity results collected from experiments run in the docking environment with $\controlmax=0.1$N. Each curve represents the IQM of 10 trials, and the shaded region is the $95\%$ confidence interval about the IQM.}
\label{fig:app_docking-0.1-sample-efficiency}
\end{figure}

In \figref{fig:app_docking-1.0-sample-efficiency} and \figref{fig:app_docking-0.1-sample-efficiency} we show the sample complexity for agents trained in the docking environment with $\controlmax = 1.0$N and $\controlmax = 0.1$N respectively.

%%%%%%%%%%%%%%%%%%%%%%%%%%%%%%%%%%%%%%%%%%%%%%%%%%%%%%%%%%%%%%%%%%%%%%%%%%%%%%%%%%%%%%%
% Final Policy Comparison Tables
%%%%%%%%%%%%%%%%%%%%%%%%%%%%%%%%%%%%%%%%%%%%%%%%%%%%%%%%%%%%%%%%%%%%%%%%%%%%%%%%%%%%%%%
\section{Final Policy Comparison Tables}

\begin{table*}
\centering
\caption{IQM Inspection $\controlmax=1.0$N}
\label{tab:app-inspection-1.0}
\begin{tabular}{llllll}
\toprule
Experiment & Total Reward & Inspected Points & Success Rate & $\deltav$ (m/s) & Episode Length (steps) \\
\midrule
Continuous & 7.8198 $\pm$ 0.5292 & 95.81 $\pm$ 4.6808 & 0.448 $\pm$ 0.4973 & 13.0222 $\pm$ 2.0929 & 323.98 $\pm$ 36.636 \\
Discrete - 101 & 7.0945 $\pm$ 0.4483 & 90.176 $\pm$ 7.2419 & 0.176 $\pm$ 0.3808 & 15.2489 $\pm$ 2.8737 & 271.516 $\pm$ 43.964 \\
Discrete - 51 & 7.7759 $\pm$ 0.528 & 93.38 $\pm$ 7.5001 & 0.466 $\pm$ 0.4988 & 12.0699 $\pm$ 2.7292 & 292.3 $\pm$ 42.679 \\
Discrete - 41 & 8.7482 $\pm$ 0.5727 & 96.102 $\pm$ 4.2047 & 0.42 $\pm$ 0.4936 & 5.662 $\pm$ 1.1497 & 325.44 $\pm$ 42.3286 \\
Discrete - 31 & 8.4412 $\pm$ 0.7315 & 94.64 $\pm$ 5.9031 & 0.434 $\pm$ 0.4956 & 6.1217 $\pm$ 1.5649 & 301.928 $\pm$ 42.8778 \\
Discrete - 21 & 8.8244 $\pm$ 0.5655 & 93.968 $\pm$ 6.2552 & 0.382 $\pm$ 0.4859 & 4.7953 $\pm$ 0.6288 & 294.198 $\pm$ 42.7977 \\
Discrete - 11 & 8.8792 $\pm$ 0.4944 & 94.936 $\pm$ 5.5771 & 0.436 $\pm$ 0.4959 & 5.1757 $\pm$ 0.6842 & 300.084 $\pm$ 36.9724 \\
Discrete - 9 & 8.6939 $\pm$ 0.6077 & 92.804 $\pm$ 6.6629 & 0.28 $\pm$ 0.449 & 5.015 $\pm$ 0.6076 & 285.528 $\pm$ 42.9038 \\
Discrete - 7 & 8.7802 $\pm$ 0.4923 & 94.894 $\pm$ 5.8569 & 0.498 $\pm$ 0.5 & 5.6317 $\pm$ 0.8868 & 295.616 $\pm$ 37.309 \\
Discrete - 5 & 9.0643 $\pm$ 0.1962 & 98.306 $\pm$ 1.1766 & 0.63 $\pm$ 0.4828 & 6.2467 $\pm$ 0.872 & 330.052 $\pm$ 25.9433 \\
Discrete - 3 & 8.7449 $\pm$ 0.3727 & 96.388 $\pm$ 4.0853 & 0.466 $\pm$ 0.4988 & 7.1767 $\pm$ 0.9963 & 309.28 $\pm$ 32.9718 \\
\bottomrule
\end{tabular}
\end{table*}

In \tabref{tab:app-inspection-1.0} we show the final policy results for agents trained with $\controlmax = 1.0$N in the inspection environment. The table shows the \textit{InterQuartile Mean} (IQM) and standard deviation for each metric.

\begin{table*}
\centering
\caption{IQM Inspection $\controlmax=0.1$N}
\label{tab:app-inspection-0.1}
\begin{tabular}{llllll}
\toprule
Experiment & Total Reward & Inspected Points & Success Rate & $\deltav$ (m/s) & Episode Length (steps) \\
\midrule
Continuous & 8.7324 $\pm$ 0.199 & 99.0 $\pm$ 0.0 & 1.0 $\pm$ 0.0 & 10.8143 $\pm$ 1.7896 & 333.496 $\pm$ 13.6857 \\
Discrete - 101 & 8.3228 $\pm$ 0.3276 & 89.098 $\pm$ 3.8216 & 0.0 $\pm$ 0.0 & 5.7558 $\pm$ 0.6629 & 312.282 $\pm$ 30.5927 \\
Discrete - 51 & 8.6179 $\pm$ 0.3202 & 92.516 $\pm$ 4.0042 & 0.0 $\pm$ 0.0 & 5.8669 $\pm$ 0.6587 & 322.662 $\pm$ 31.1315 \\
Discrete - 41 & 8.4393 $\pm$ 0.3339 & 90.492 $\pm$ 4.1859 & 0.0 $\pm$ 0.0 & 5.8385 $\pm$ 0.7675 & 319.386 $\pm$ 32.385 \\
Discrete - 31 & 8.6998 $\pm$ 0.2798 & 94.136 $\pm$ 3.4741 & 0.072 $\pm$ 0.2585 & 6.5631 $\pm$ 0.637 & 343.786 $\pm$ 30.6773 \\
Discrete - 21 & 8.4793 $\pm$ 0.424 & 90.526 $\pm$ 5.07 & 0.0 $\pm$ 0.0 & 5.3385 $\pm$ 0.6996 & 324.068 $\pm$ 39.9499 \\
Discrete - 11 & 8.7238 $\pm$ 0.3049 & 94.034 $\pm$ 3.8777 & 0.092 $\pm$ 0.289 & 6.1293 $\pm$ 0.7255 & 335.342 $\pm$ 26.6045 \\
Discrete - 9 & 8.5916 $\pm$ 0.3967 & 92.73 $\pm$ 4.8807 & 0.07 $\pm$ 0.2551 & 6.0518 $\pm$ 0.7598 & 334.318 $\pm$ 34.7523 \\
Discrete - 7 & 8.4511 $\pm$ 0.4318 & 91.23 $\pm$ 5.6037 & 0.04 $\pm$ 0.196 & 6.0729 $\pm$ 0.9819 & 321.638 $\pm$ 40.5397 \\
Discrete - 5 & 8.6838 $\pm$ 0.2979 & 94.988 $\pm$ 3.8595 & 0.226 $\pm$ 0.4182 & 7.2322 $\pm$ 0.741 & 344.036 $\pm$ 35.2377 \\
Discrete - 3 & 8.9796 $\pm$ 0.3325 & 96.698 $\pm$ 3.3898 & 0.458 $\pm$ 0.4982 & 5.1738 $\pm$ 0.946 & 334.572 $\pm$ 42.9696 \\
\bottomrule
\end{tabular}
\end{table*}

In \tabref{tab:app-inspection-0.1} we show the final policy results for agents trained with $\controlmax = 0.1$N in the inspection environment. The table shows the IQM and standard deviation for each metric.

\begin{table*}
\centering
\caption{IQM Docking $\controlmax=1.0$N}
\label{tab:app-docking-1.0}
\resizebox{\linewidth}{!}{
\begin{tabular}{lllllll}
\toprule
Experiment & Total Reward & Success Rate & $\deltav$ (m/s) & Violation (\%) & Final Speed (m/s) & Episode Length (steps) \\
\midrule
Continuous & 1.4105 $\pm$ 0.5279 & 0.57 $\pm$ 0.4951 & 13.2319 $\pm$ 1.7153 & 0.0 $\pm$ 0.0 & 0.0141 $\pm$ 0.0043 & 1780.944 $\pm$ 237.6564 \\
Discrete 1.0/../0.001 & 1.9307 $\pm$ 0.5696 & 1.0 $\pm$ 0.0 & 43.01 $\pm$ 16.9203 & 1.5736 $\pm$ 2.5145 & 0.0886 $\pm$ 0.0159 & 876.132 $\pm$ 184.738 \\
Discrete 1.0/0.1 & 2.0274 $\pm$ 0.5583 & 1.0 $\pm$ 0.0 & 90.128 $\pm$ 27.5016 & 0.8247 $\pm$ 1.4835 & 0.1027 $\pm$ 0.0177 & 747.38 $\pm$ 117.5253 \\
Discrete - 101 & 2.1488 $\pm$ 0.2312 & 1.0 $\pm$ 0.0 & 293.3862 $\pm$ 29.6104 & 1.6089 $\pm$ 2.4567 & 0.0861 $\pm$ 0.0164 & 671.596 $\pm$ 54.1947 \\
Discrete - 51 & 1.7891 $\pm$ 0.5505 & 0.978 $\pm$ 0.1467 & 391.2823 $\pm$ 75.8659 & 0.3771 $\pm$ 0.6783 & 0.0613 $\pm$ 0.0152 & 927.566 $\pm$ 278.6582 \\
Discrete - 41 & 2.107 $\pm$ 0.2374 & 1.0 $\pm$ 0.0 & 308.5834 $\pm$ 49.6744 & 0.4602 $\pm$ 0.751 & 0.0753 $\pm$ 0.014 & 728.69 $\pm$ 104.8544 \\
Discrete - 31 & 2.1907 $\pm$ 0.2392 & 1.0 $\pm$ 0.0 & 251.3246 $\pm$ 44.1849 & 0.1856 $\pm$ 0.4052 & 0.0733 $\pm$ 0.0164 & 774.518 $\pm$ 154.2326 \\
Discrete - 21 & 1.5185 $\pm$ 0.6643 & 0.802 $\pm$ 0.3985 & 269.6558 $\pm$ 45.6941 & 0.7576 $\pm$ 1.3076 & 0.0527 $\pm$ 0.0149 & 1220.52 $\pm$ 432.1194 \\
Discrete - 11 & 1.2417 $\pm$ 0.5921 & 0.52 $\pm$ 0.4996 & 50.0763 $\pm$ 66.8249 & 0.1531 $\pm$ 0.4143 & 0.0531 $\pm$ 0.0125 & 1610.08 $\pm$ 408.8315 \\
Discrete - 9 & 0.884 $\pm$ 0.3165 & 0.324 $\pm$ 0.468 & 24.6879 $\pm$ 34.5683 & 0.4601 $\pm$ 1.0115 & 0.0535 $\pm$ 0.0132 & 1717.72 $\pm$ 401.7791 \\
Discrete - 7 & 0.7117 $\pm$ 0.1693 & 0.0 $\pm$ 0.0 & 10.5039 $\pm$ 1.4224 & 0.632 $\pm$ 1.1258 & 0.0429 $\pm$ 0.0084 & 2000.0 $\pm$ 0.0 \\
Discrete - 5 & 0.6349 $\pm$ 0.1275 & 0.0 $\pm$ 0.0 & 9.5683 $\pm$ 1.3141 & 0.0419 $\pm$ 0.1765 & 0.0426 $\pm$ 0.0081 & 2000.0 $\pm$ 0.0 \\
Discrete - 3 & 0.6934 $\pm$ 0.1322 & 0.0 $\pm$ 0.0 & 9.5583 $\pm$ 1.3253 & 0.0787 $\pm$ 0.2588 & 0.0574 $\pm$ 0.0099 & 2000.0 $\pm$ 0.0 \\
\bottomrule
\end{tabular}
}
\end{table*}

In \tabref{tab:app-docking-0.1} we show the final policy results for agents trained with $\controlmax = 1.0$N in the docking environment. The table shows the IQM and standard deviation for each metric.

\begin{table*}
\centering
\caption{IQM Docking $\controlmax=0.1$N}
\label{tab:app-docking-0.1}
\begin{tabular}{lllllll}
\toprule
Experiment & Total Reward & Success Rate & $\deltav$ (m/s) & Violation (\%) & Final Speed (m/s) & Episode Length (steps) \\
\midrule
Continuous & 1.8289 $\pm$ 0.5193 & 0.842 $\pm$ 0.3647 & 11.6234 $\pm$ 1.3619 & 0.0 $\pm$ 0.0 & 0.0131 $\pm$ 0.0074 & 1497.154 $\pm$ 343.938 \\
% Discrete 1.0/../0.001 & 1.9307 $\pm$ 0.5696 & 1.0 $\pm$ 0.0 & 43.01 $\pm$ 16.9203 & 1.5736 $\pm$ 2.5145 & 0.0886 $\pm$ 0.0159 & 876.132 $\pm$ 184.738 \\
% Discrete 1.0/0.1 & 2.0274 $\pm$ 0.5583 & 1.0 $\pm$ 0.0 & 90.128 $\pm$ 27.5016 & 0.8247 $\pm$ 1.4835 & 0.1027 $\pm$ 0.0177 & 747.38 $\pm$ 117.5253 \\
Discrete - 101 & 1.6612 $\pm$ 0.689 & 0.72 $\pm$ 0.449 & 98.9843 $\pm$ 23.4398 & 0.6285 $\pm$ 1.3201 & 0.023 $\pm$ 0.0107 & 1212.808 $\pm$ 475.4788 \\
Discrete - 51 & 0.7218 $\pm$ 0.2168 & 0.044 $\pm$ 0.2051 & 111.8182 $\pm$ 21.1845 & 1.3326 $\pm$ 1.6915 & 0.0097 $\pm$ 0.0066 & 1948.042 $\pm$ 157.2884 \\
Discrete - 41 & 1.8428 $\pm$ 0.7146 & 0.776 $\pm$ 0.4169 & 86.7219 $\pm$ 7.7169 & 0.1063 $\pm$ 0.3616 & 0.0316 $\pm$ 0.0157 & 1124.558 $\pm$ 464.4369 \\
Discrete - 31 & 1.4687 $\pm$ 0.6669 & 0.604 $\pm$ 0.4891 & 91.2584 $\pm$ 10.8726 & 0.4691 $\pm$ 1.0746 & 0.0306 $\pm$ 0.0161 & 1321.878 $\pm$ 559.5975 \\
Discrete - 21 & 1.7366 $\pm$ 0.6367 & 0.838 $\pm$ 0.3685 & 104.1883 $\pm$ 12.1358 & 0.3216 $\pm$ 0.8018 & 0.0226 $\pm$ 0.0095 & 1268.57 $\pm$ 397.6955 \\
Discrete - 11 & 1.7699 $\pm$ 0.6552 & 0.928 $\pm$ 0.2585 & 86.4941 $\pm$ 12.5111 & 1.1875 $\pm$ 2.3687 & 0.0628 $\pm$ 0.0254 & 912.868 $\pm$ 214.4565 \\
Discrete - 9 & 1.4149 $\pm$ 0.8378 & 0.648 $\pm$ 0.4776 & 106.863 $\pm$ 25.6372 & 0.2353 $\pm$ 0.6563 & 0.0468 $\pm$ 0.028 & 1242.01 $\pm$ 490.6625 \\
Discrete - 7 & 1.8318 $\pm$ 0.7563 & 0.782 $\pm$ 0.4129 & 95.567 $\pm$ 18.1873 & 0.0 $\pm$ 0.0 & 0.0663 $\pm$ 0.037 & 1128.162 $\pm$ 489.9392 \\
Discrete - 5 & 2.0283 $\pm$ 0.6071 & 1.0 $\pm$ 0.0 & 77.0818 $\pm$ 14.2023 & 0.0508 $\pm$ 0.247 & 0.0923 $\pm$ 0.0247 & 886.076 $\pm$ 215.382 \\
Discrete - 3 & 0.6064 $\pm$ 0.167 & 0.0 $\pm$ 0.0 & 15.0623 $\pm$ 7.5436 & 0.0 $\pm$ 0.0 & 0.0279 $\pm$ 0.0108 & 2000.0 $\pm$ 0.0 \\
\bottomrule
\end{tabular}
\end{table*}

In \tabref{tab:app-docking-0.1} we show the final policy results for agents trained with $\controlmax = 0.1$N in the docking environment. The table shows the IQM and standard deviation for each metric.

%%%%%%%%%%%%%%%%%%%%%%%%%%%%%%%%%%%%%%%%%%%%%%%%%%%%%%%%%%%%%%%%%%%%%%%%%%%%%%%%%%%%%%%
% Additional Final Policy Comparison Figures
%%%%%%%%%%%%%%%%%%%%%%%%%%%%%%%%%%%%%%%%%%%%%%%%%%%%%%%%%%%%%%%%%%%%%%%%%%%%%%%%%%%%%%%
\section{Additional Final Policy Comparison Figures}

\begin{figure*}[ht]
\centering
\subfigure[Total Reward.]{\includegraphics[width=0.45\linewidth]{figures/inspection/1_0/TotalReward_int_est.png}}
\subfigure[Inspected points.]{\includegraphics[width=0.45\linewidth]{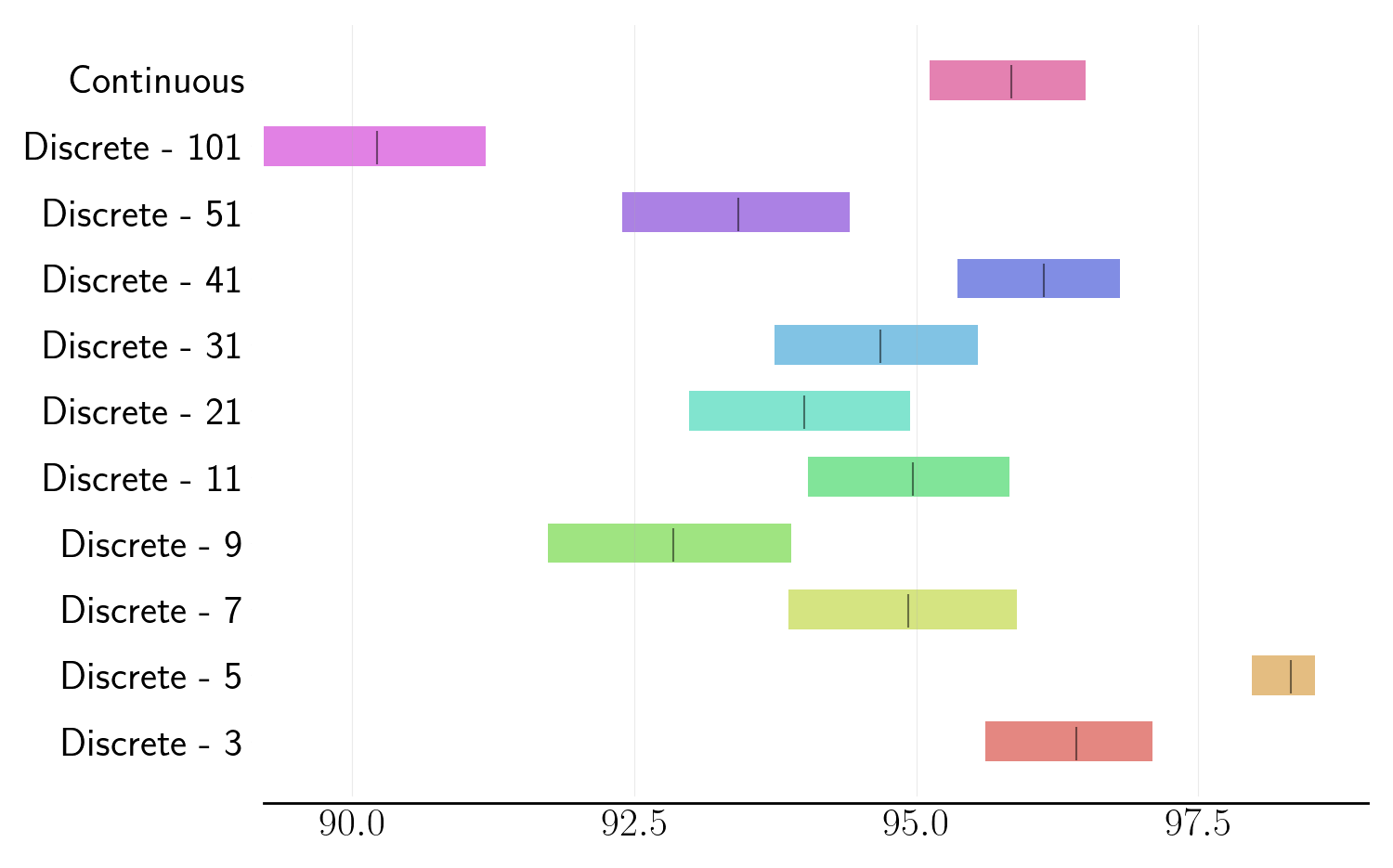}}
\subfigure[$\deltav$ (m/s).]{\includegraphics[width=0.45\linewidth]{figures/inspection/1_0/DeltaV_int_est.png}}
\subfigure[Success rate.]{\includegraphics[width=0.45\linewidth]{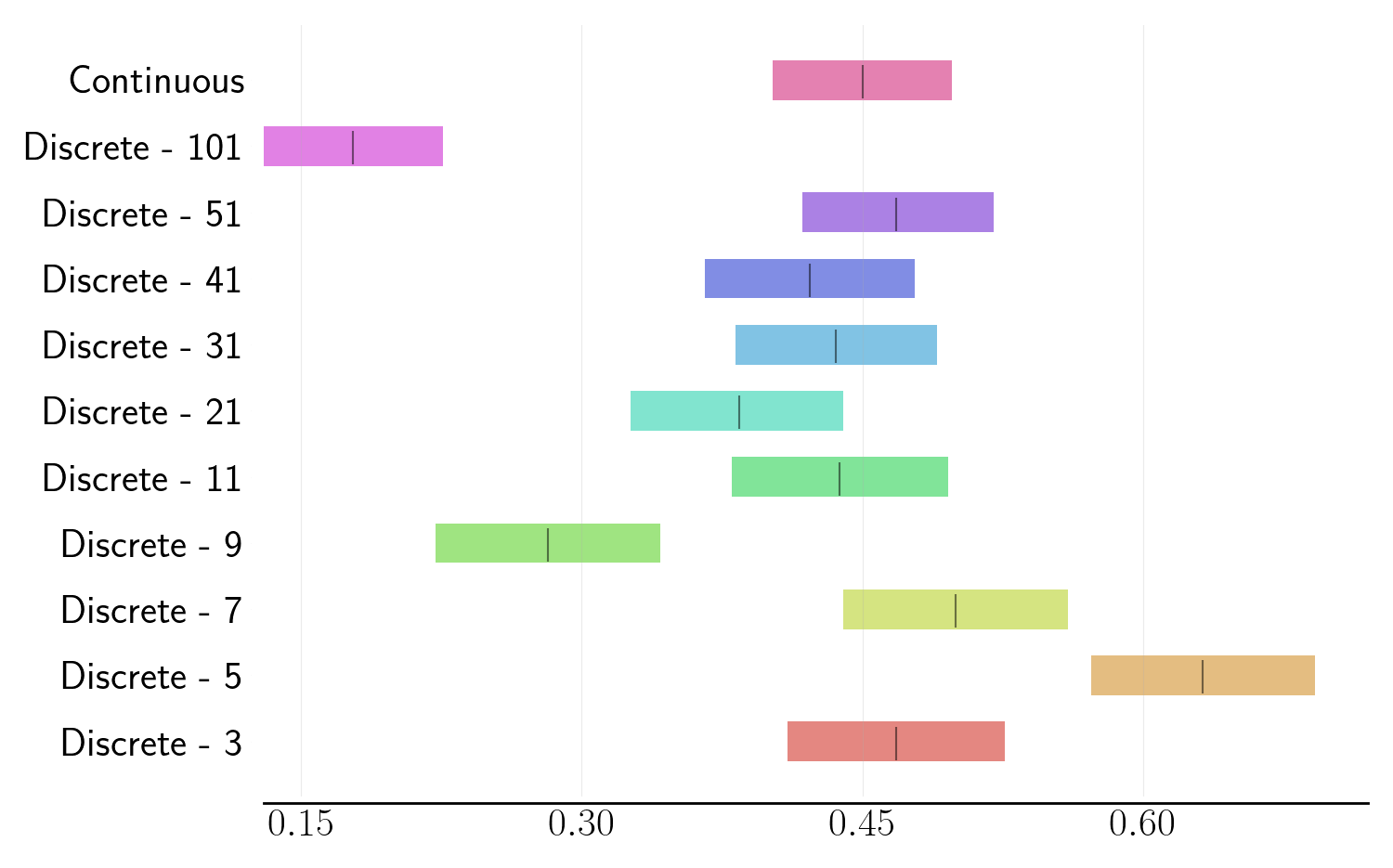}}
\subfigure[Episode length (steps).]{\includegraphics[width=0.45\linewidth]{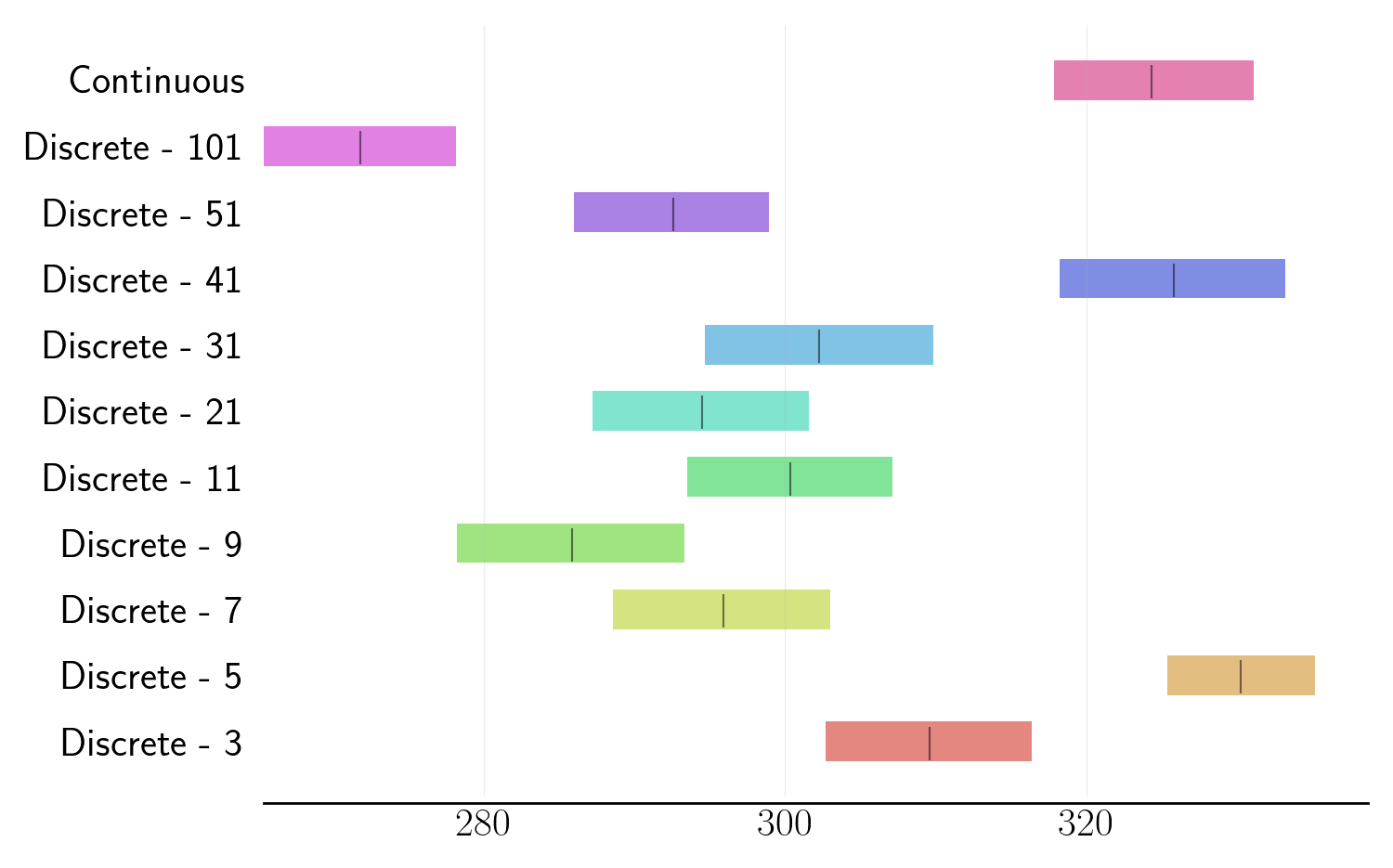}}
\caption{Comparison of final policies trained in the inspection environment with $\controlmax=1.0$N. Each marker represents the IQM of 100 trials, and the shaded region is the $95\%$ confidence interval about the IQM.}
\label{fig:app_inspection-1.0-int-est-extra}
\end{figure*}

\begin{figure*}[ht]
\centering
\subfigure[Total Reward.]{\includegraphics[width=0.45\linewidth]{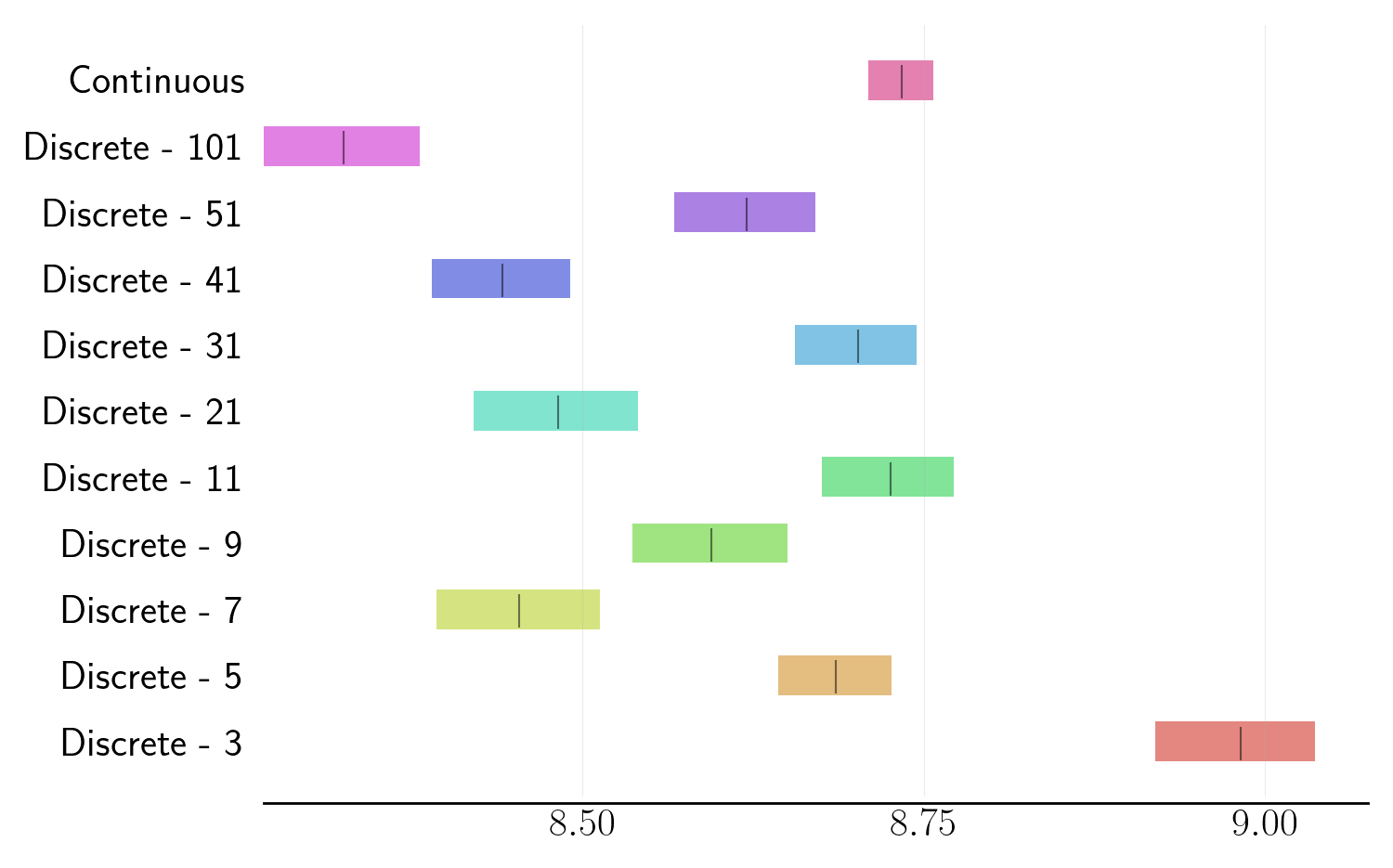}}
\subfigure[Inspected points.]{\includegraphics[width=0.45\linewidth]{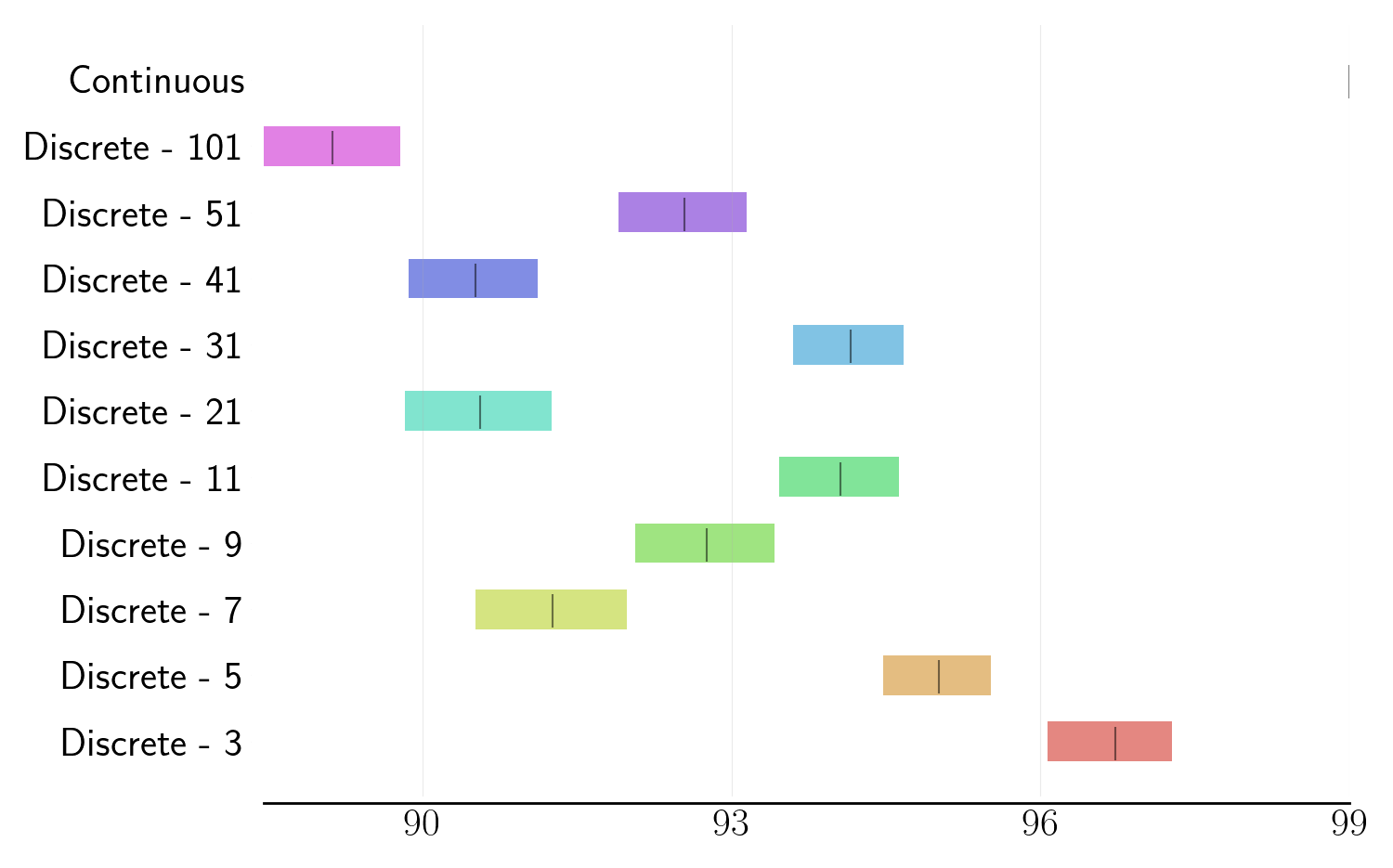}}
\subfigure[$\deltav$ (m/s).]{\includegraphics[width=0.45\linewidth]{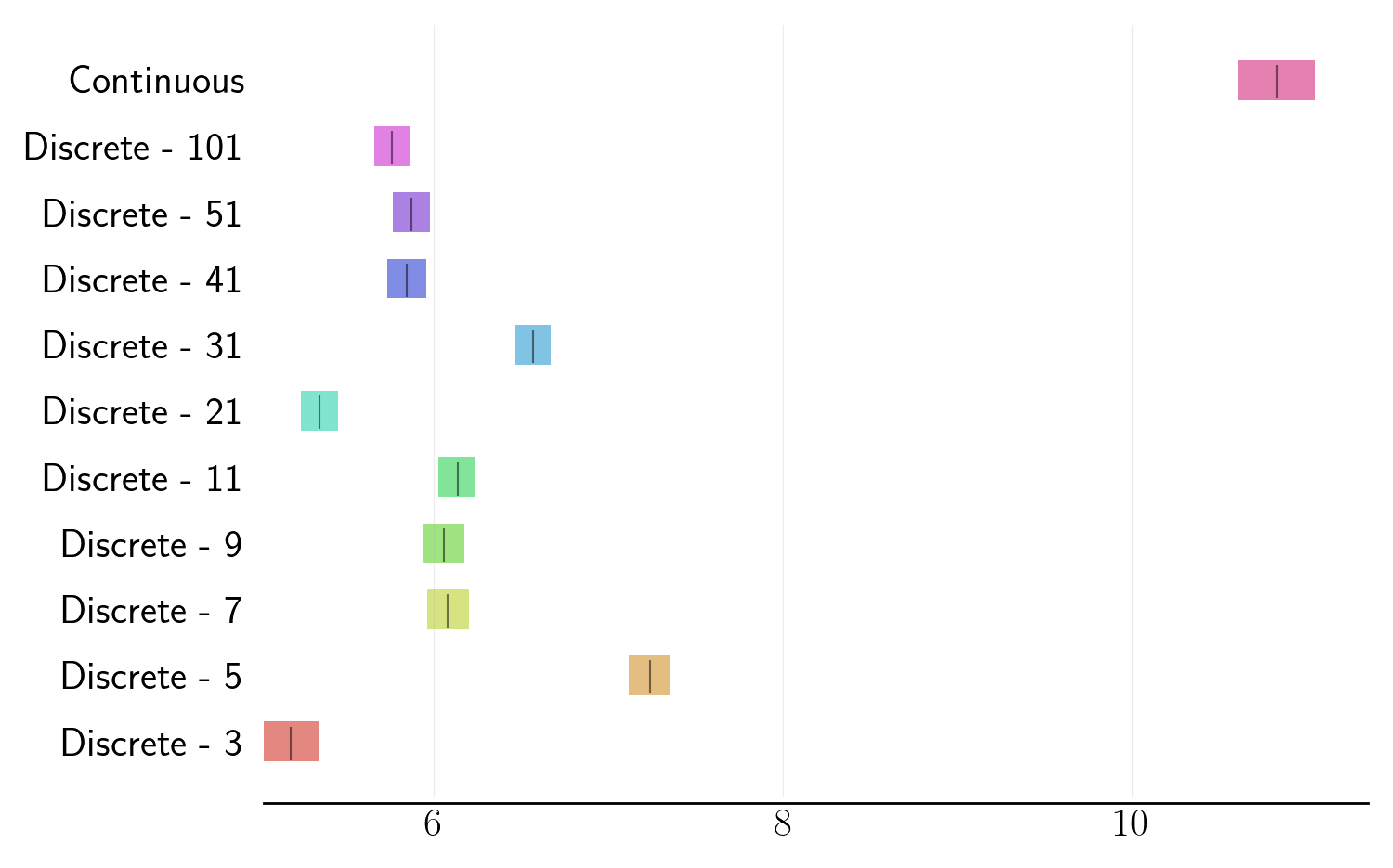}}
\subfigure[Success rate.]{\includegraphics[width=0.45\linewidth]{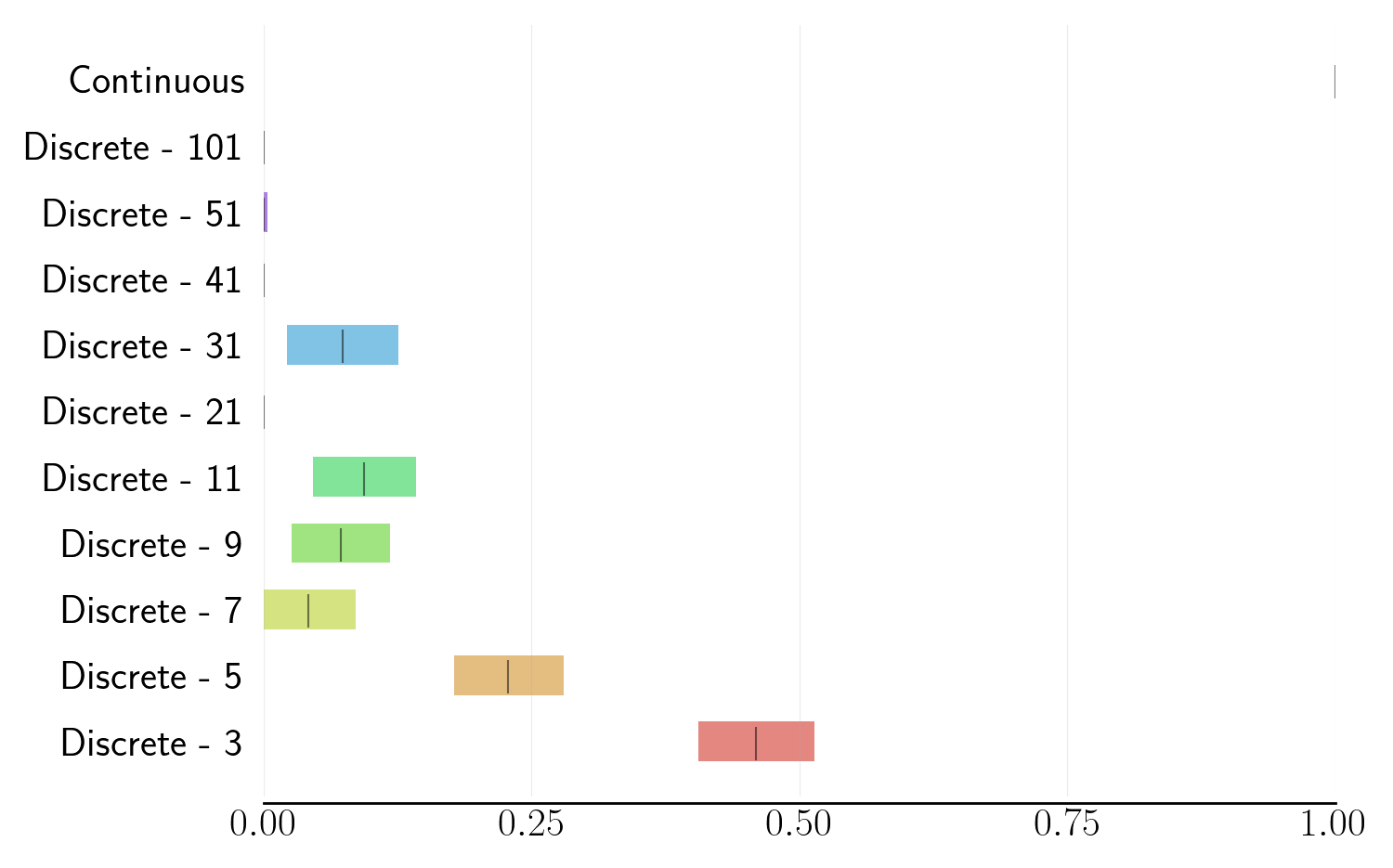}}
\subfigure[Episode length (steps).]{\includegraphics[width=0.45\linewidth]{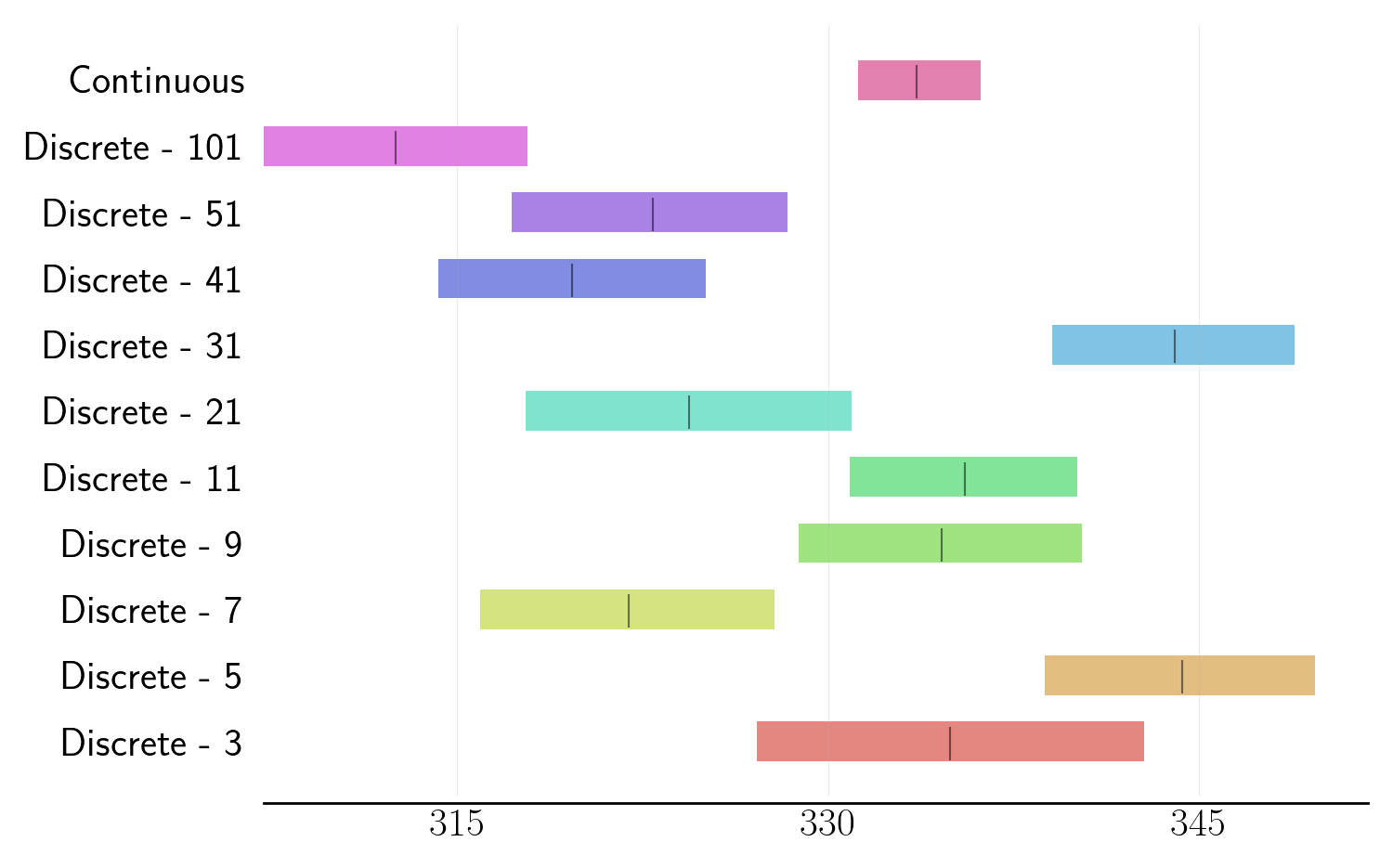}}
\caption{Comparison of final policies trained in the inspection environment with $\controlmax=0.1$N. Each marker represents the IQM of 100 trials, and the shaded region is the $95\%$ confidence interval about the IQM.}
\label{fig:app_inspection-0.1-int-est-extra}
\end{figure*}

In \figref{fig:app_inspection-1.0-int-est-extra} and \figref{fig:app_inspection-0.1-int-est-extra} we show the final policy performance with respect to the number of inspected points, success rate, and episode length for agents trained in the inspection environment with $\controlmax = 1.0$N and $\controlmax = 0.1$N respectively. Comparisons of the $\deltav$ use and reward are shown in \figref{fig:delta-v-int-est} and \figref{fig:insp-reward-int-est} respectively.

\begin{figure*}[ht]
\centering
\subfigure[Total reward.]{\includegraphics[width=0.45\linewidth]{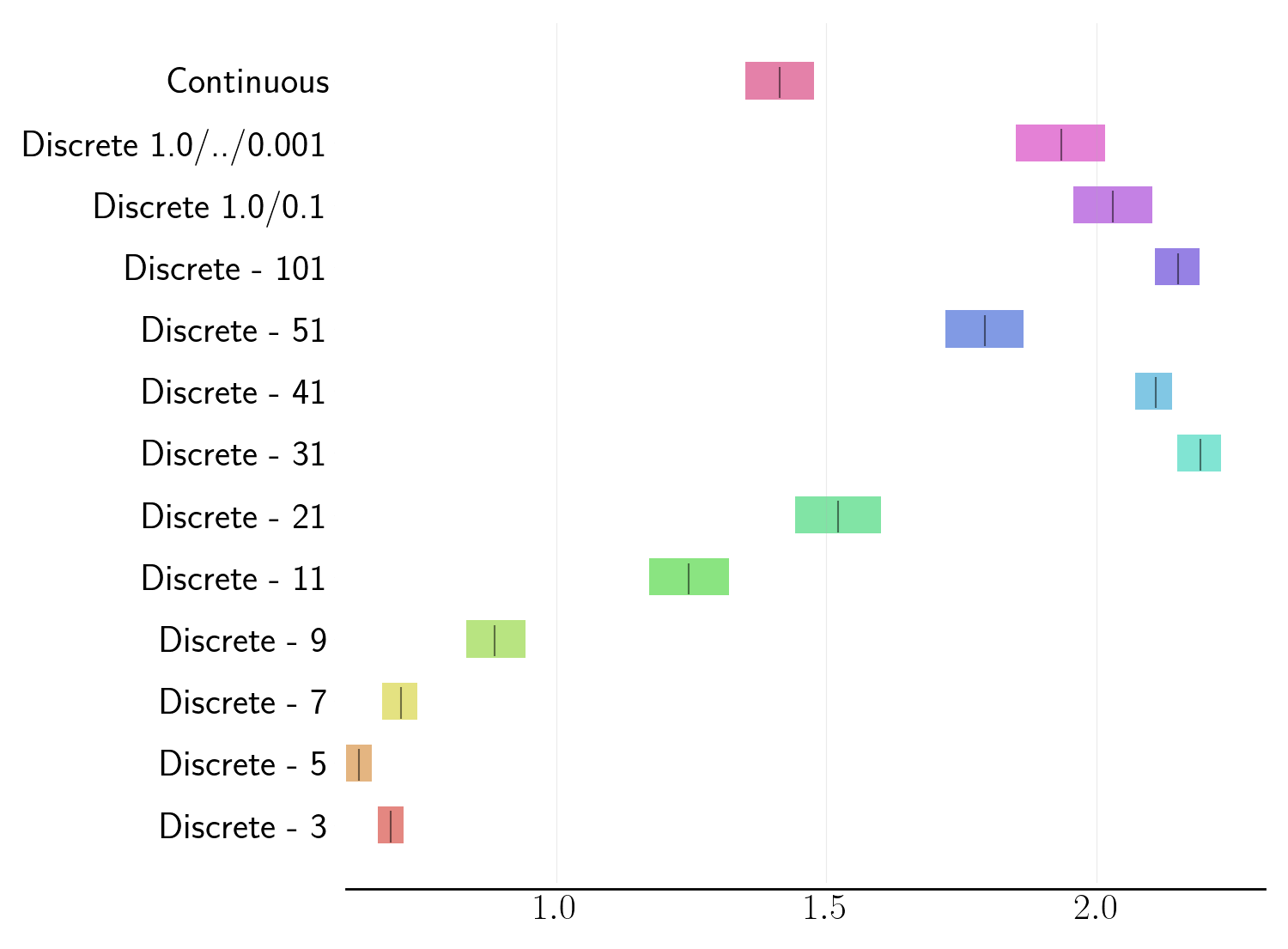}}
\subfigure[Success rate.]{\includegraphics[width=0.45\linewidth]{figures/docking/1_0/Success_int_est.png}}
\subfigure[$\deltav$ (m/s).]{\includegraphics[width=0.45\linewidth]{figures/docking/1_0/DeltaV_int_est.png}}
\subfigure[Constraint violation (\%).]{\includegraphics[width=0.45\linewidth]{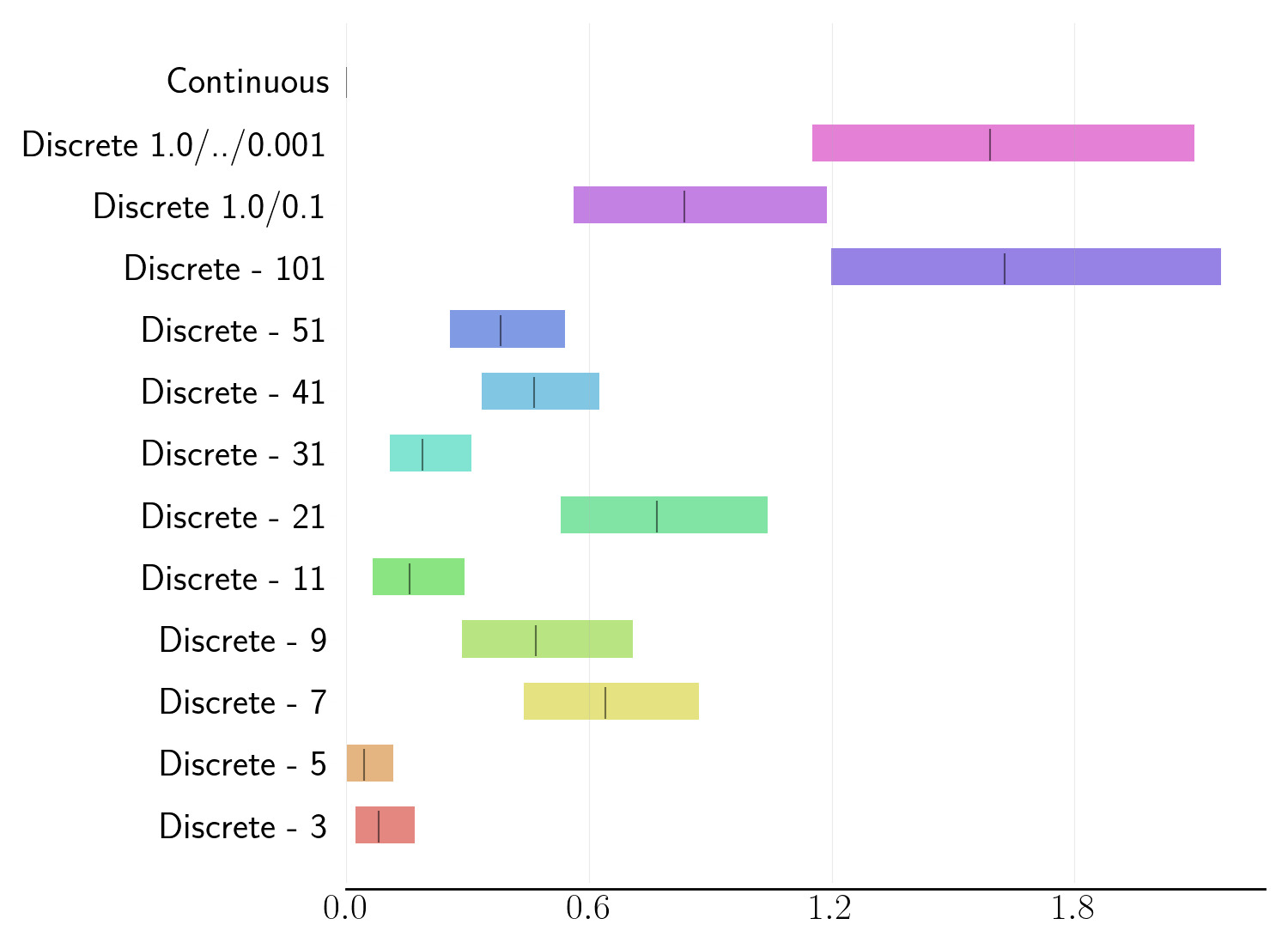}}
\subfigure[Final speed (m/s).]{\includegraphics[width=0.45\linewidth]{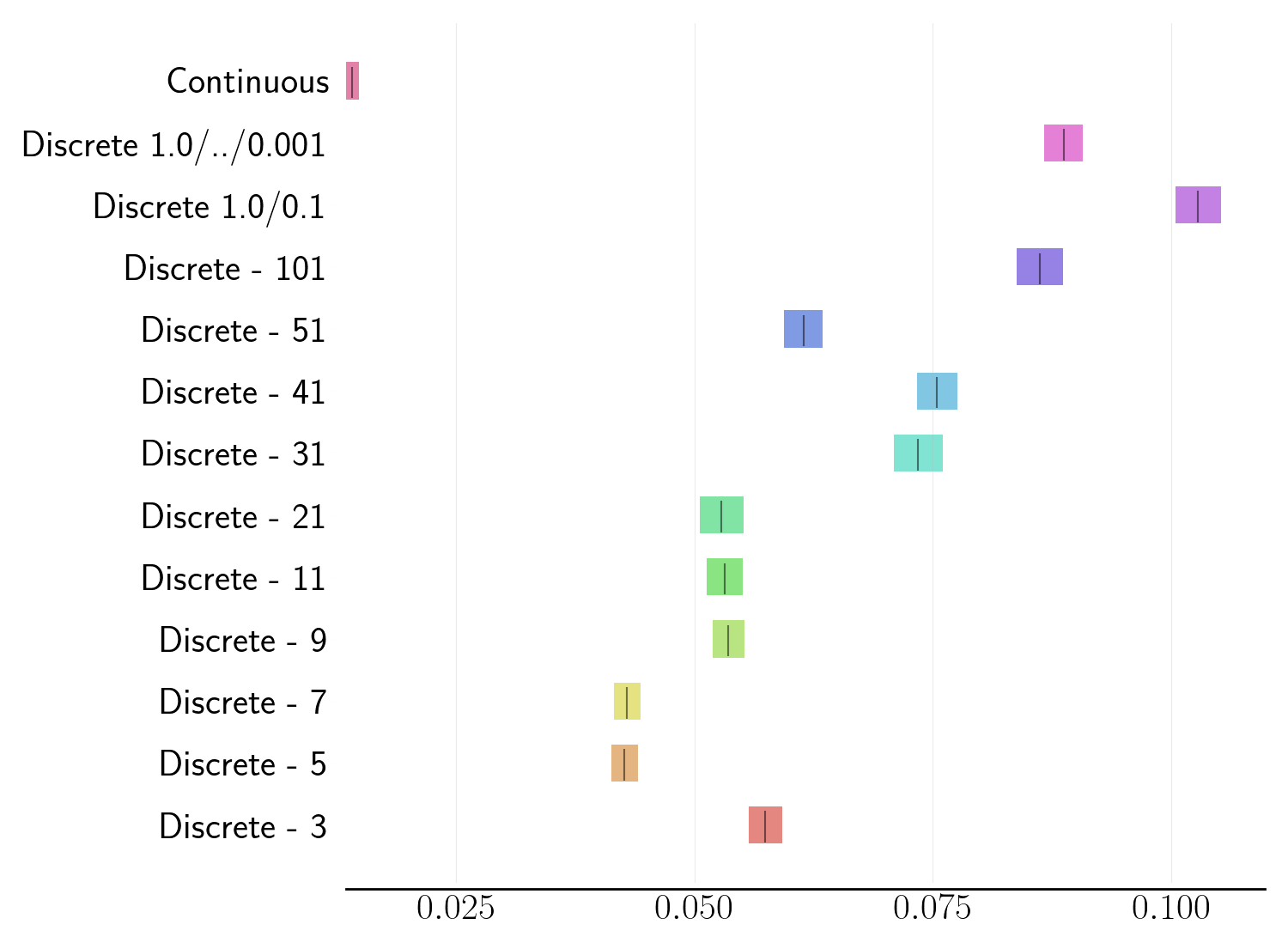}}
\subfigure[Episode length (steps).]{\includegraphics[width=0.45\linewidth]{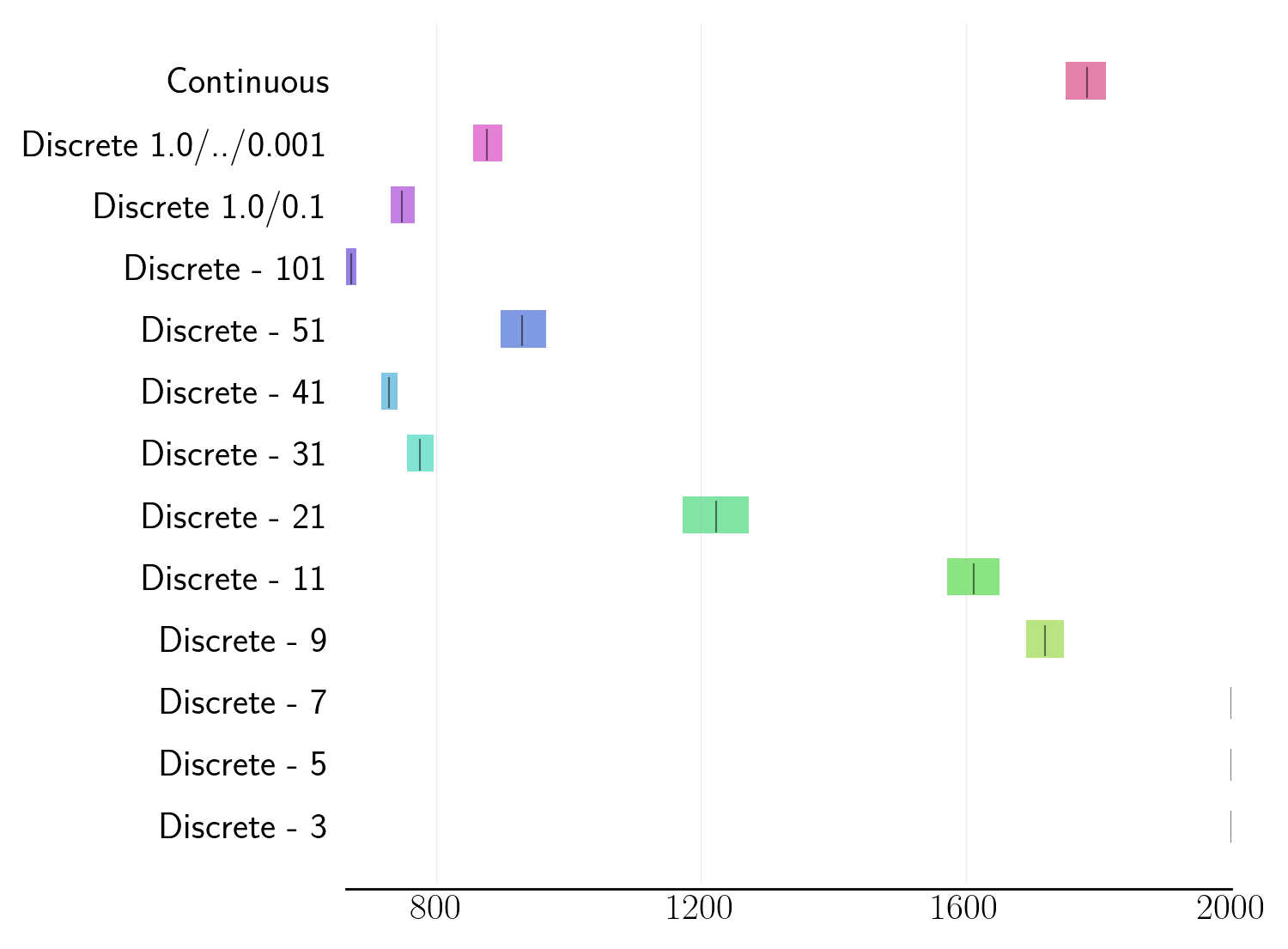}}
\caption{Comparison of final policies trained in the docking environment with $\controlmax=1.0$N. Each marker represents the IQM of 100 trials, and the shaded region is the $95\%$ confidence interval about the IQM.}
\label{fig:app_docking-1.0-int-est-extra}
\end{figure*}

\begin{figure*}[ht]
\centering
\subfigure[Total reward.]{\includegraphics[width=0.45\linewidth]{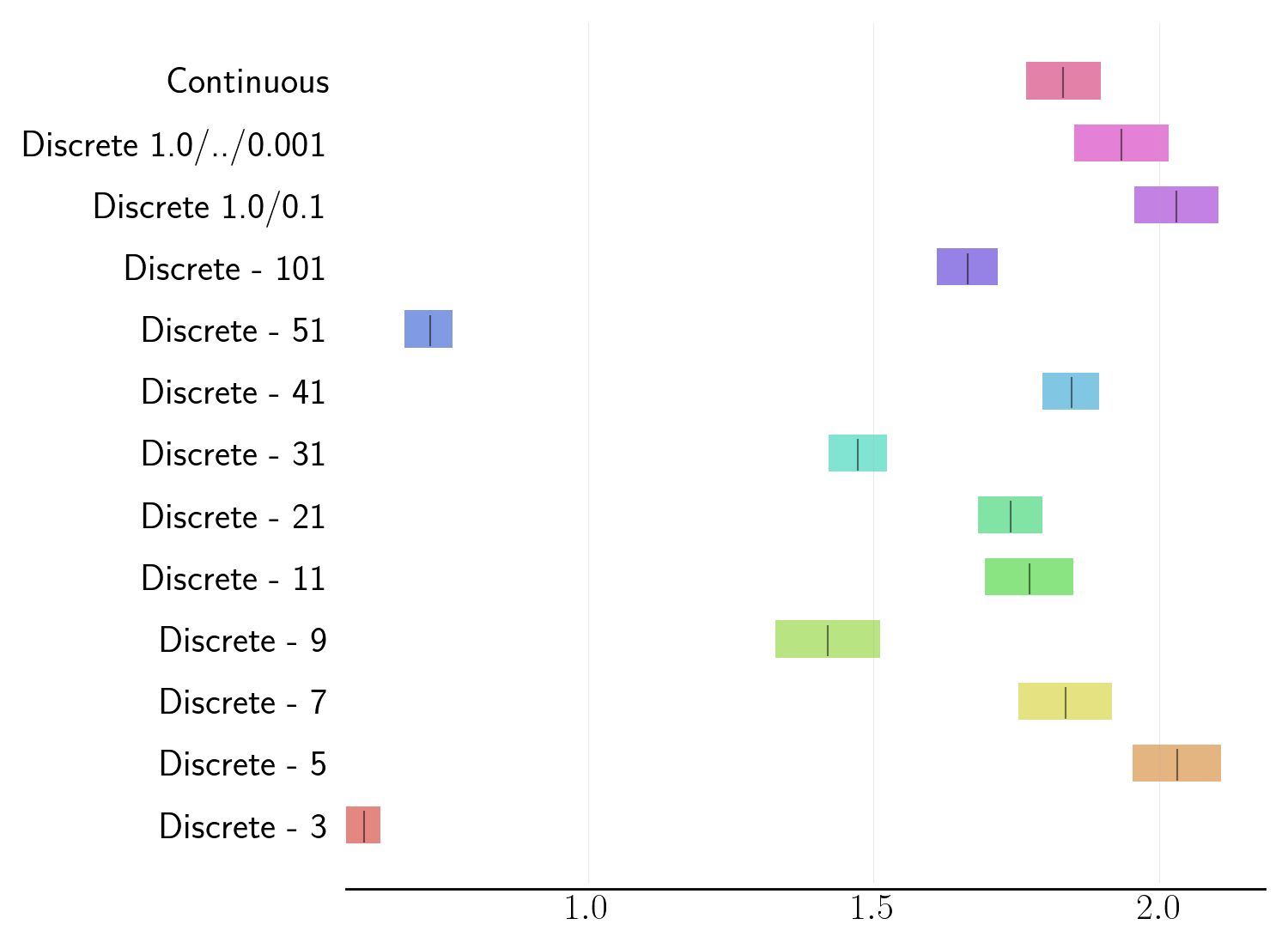}}
\subfigure[Success rate.]{\includegraphics[width=0.45\linewidth]{figures/docking/0_1/Success_int_est.png}}
\subfigure[$\deltav$ (m/s).]{\includegraphics[width=0.45\linewidth]{figures/docking/0_1/DeltaV_int_est.png}}
\subfigure[Constraint violation (\%).]{\includegraphics[width=0.45\linewidth]{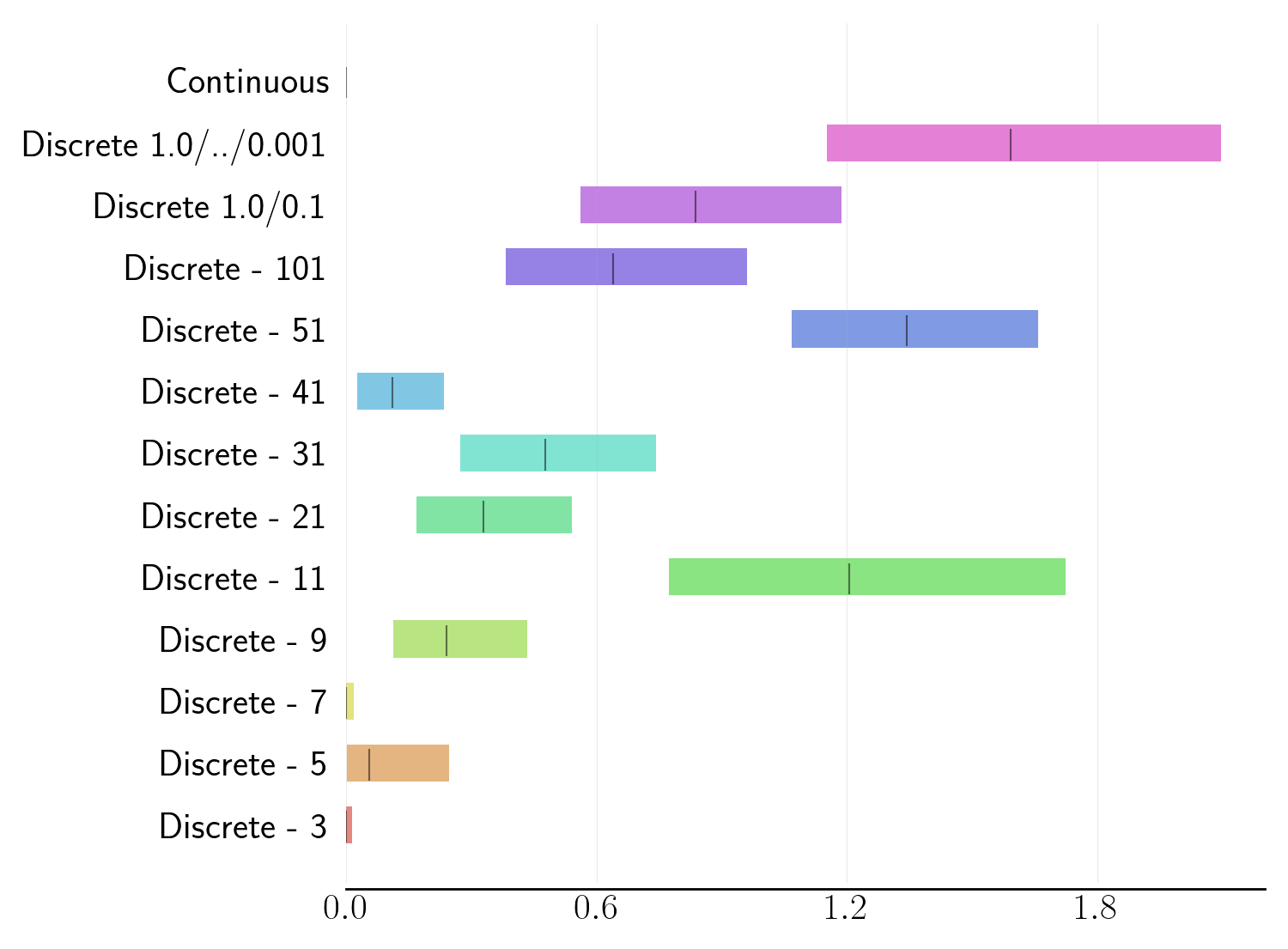}}
\subfigure[Final speed (m/s).]{\includegraphics[width=0.45\linewidth]{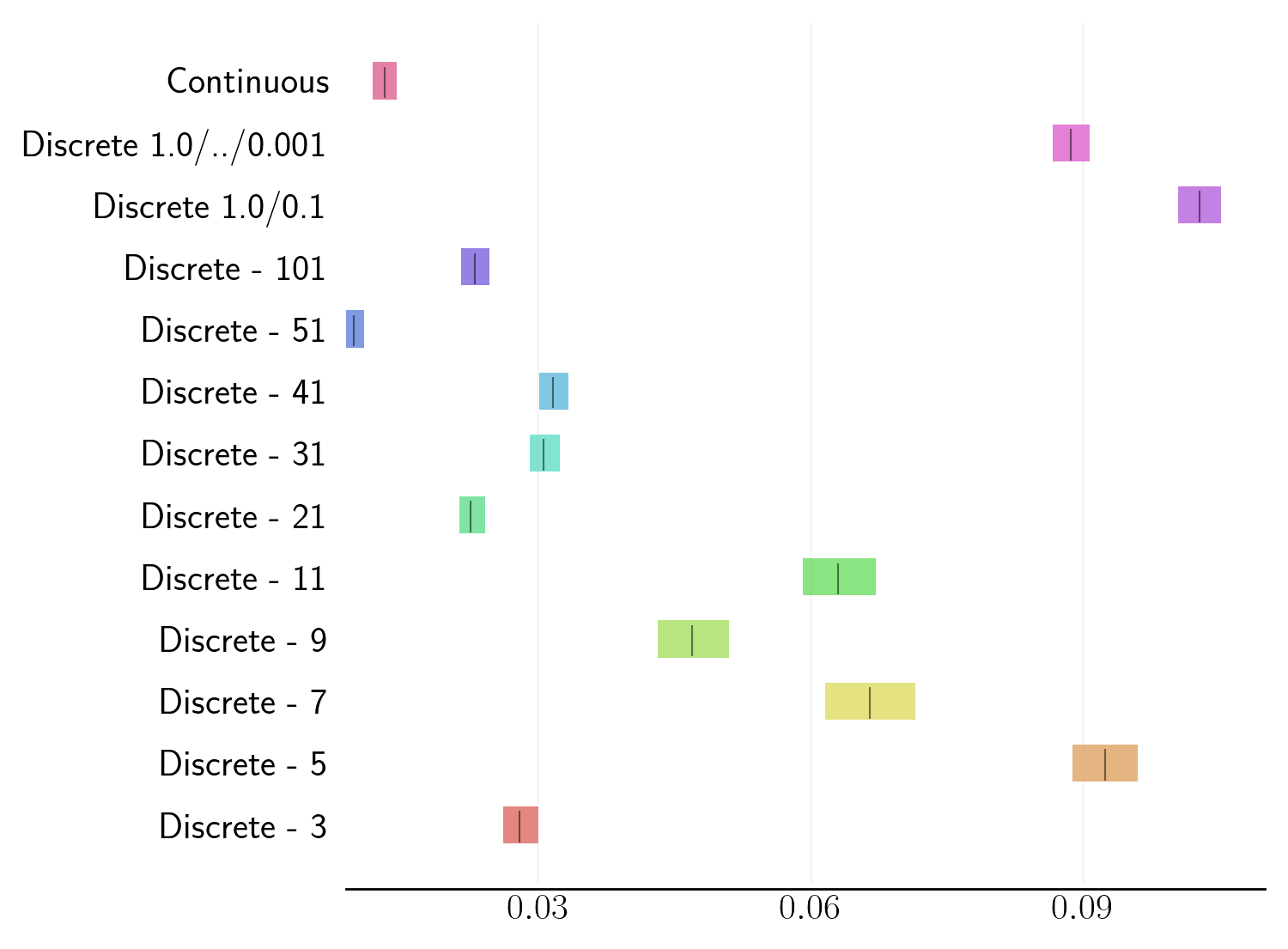}}
\subfigure[Episode length (steps).]{\includegraphics[width=0.45\linewidth]{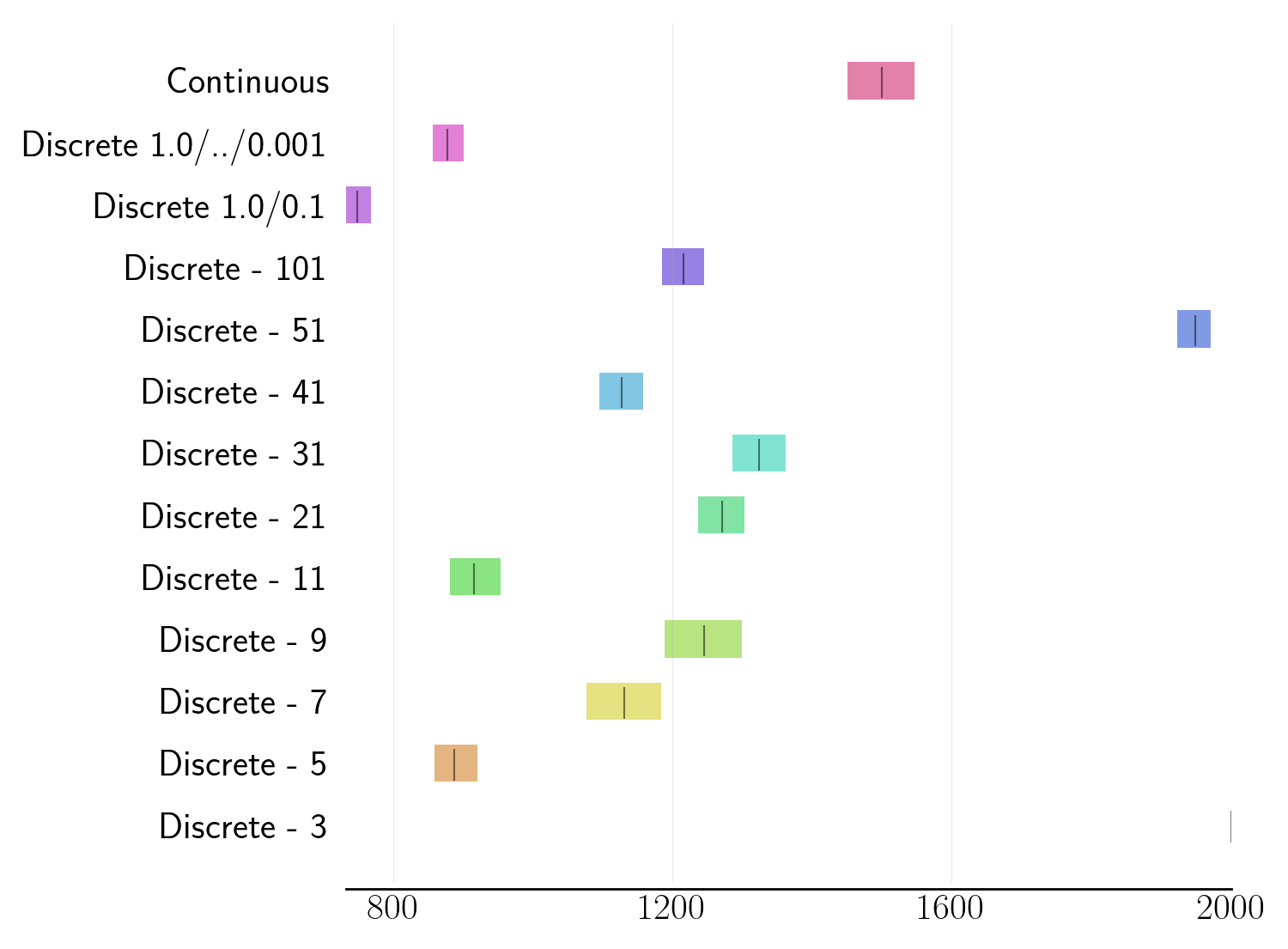}}
\caption{Comparison of final policies trained in the docking environment with $\controlmax=0.1$N. Each marker represents the IQM of 100 trials, and the shaded region is the $95\%$ confidence interval about the IQM.}
\label{fig:app_docking-0.1-int-est-extra}
\end{figure*}

In \figref{fig:app_docking-1.0-int-est-extra} and \figref{fig:app_docking-0.1-int-est-extra} we show the final policy performance with respect to the total reward, constraint violation percentage, final speed, and episode length for agents trained in the docking environment with $\controlmax = 1.0$N and $\controlmax = 0.1$N respectively. Comparisons of the $\deltav$ use and success rate are shown in \figref{fig:delta-v-int-est} and \figref{fig:dock-reward-int-est} respectively.

\end{appendices}

\end{document}